\documentclass{article}

\usepackage{microtype}
\usepackage{graphicx}
\usepackage{subcaption}
\usepackage{booktabs} %

\usepackage{hyperref}

\usepackage[accepted]{styles/arxiv_2col/arxiv_2col}

\usepackage{amsmath}
\usepackage{amssymb}
\usepackage{mathtools}
\usepackage{amsthm}

\usepackage[capitalize,noabbrev]{cleveref}

\theoremstyle{plain}
\newtheorem{theorem}{Theorem}[section]

\theoremstyle{definition}
\newtheorem{definition}[theorem]{Definition}
\newtheorem{assumption}[theorem]{Assumption}
\theoremstyle{remark}

\usepackage[textsize=tiny]{todonotes}

\usepackage{multirow}

\renewcommand{\paragraph}[1]{\textbf{#1} }

\usepackage{tikz}
\usepackage{wrapfig}
\usepackage{bm}
\usepackage{placeins}
\usepackage{tcolorbox}

\icmltitlerunning{PAINT: Parallel-in-time Neural Twins for Dynamical System Reconstruction}

\begin{document}

\twocolumn[
  \icmltitle{PAINT: Parallel-in-time Neural Twins \\for Dynamical System Reconstruction}

  \icmlsetsymbol{equal}{*}

  \begin{icmlauthorlist}
    \icmlauthor{Andreas Radler}{equal,lit}
    \icmlauthor{Vincent Seyfried}{equal,pfm,cd}
    \icmlauthor{Johannes Brandstetter}{lit,emmi}
    \icmlauthor{Thomas Lichtenegger}{pfm,cd,emmi}
  \end{icmlauthorlist}

  \icmlaffiliation{lit}{LIT AI Lab, Institute for Machine Learning, JKU Linz, Austria}
  \icmlaffiliation{pfm}{Department of Particulate Flow Modelling, JKU Linz, Austria}
  \icmlaffiliation{cd}{Christian Doppler Laboratory for Data-Assisted Simulations of Complex Flows, Linz, Austria}
  \icmlaffiliation{emmi}{Emmi AI GmbH, Linz, Austria}

  \icmlcorrespondingauthor{Andreas Radler}{radler@ml.jku.at}
  \icmlcorrespondingauthor{Johannes Brandstetter}{brandstetter@ml.jku.at}

  \icmlkeywords{Machine Learning, ICML}

  \vskip 0.3in
]

\printAffiliationsAndNotice{}  %

\begin{abstract}

Neural surrogates have shown great potential in simulating dynamical systems, while offering real-time capabilities. 
We envision \emph{Neural Twins} as a progression of neural surrogates, aiming to create digital replicas of real systems. 
A neural twin consumes measurements at test time to update its state, thereby enabling context-specific decision-making.
We argue, that a critical property of neural twins is their ability to remain \emph{on-trajectory}, i.e., to stay close to the true system state over time.
We introduce \textbf{Pa}rallel-in-t\textbf{i}me \textbf{N}eural \textbf{T}wins (\textbf{PAINT}), an architecture-agnostic family of methods for modeling dynamical systems from measurements.
PAINT trains a generative neural network to model the distribution of states in parallel over time.
At test time, states are predicted from measurements in a sliding window fashion.
Our theoretical analysis shows that PAINT is on-trajectory, whereas autoregressive models generally are not.
Empirically, we evaluate our method on a challenging two-dimensional turbulent fluid dynamics problem. 
The results demonstrate that PAINT stays on-trajectory and predicts system states from sparse measurements with high fidelity. 
These findings underscore PAINT's potential for developing neural twins 
that stay on-trajectory, 
enabling more accurate state estimation and decision-making.
\footnote{Code is available at \url{https://github.com/ParticulateFlow/PAINT-Parallel_in_time_Neural_Twins}}

\end{abstract}

\begin{figure}[h!]
    \centering
    \resizebox{8.5cm}{!}{
    \begin{tikzpicture}
    \node[anchor=south west,inner sep=0] (image) at (0,0) {
        \includegraphics[width=0.7\textwidth]{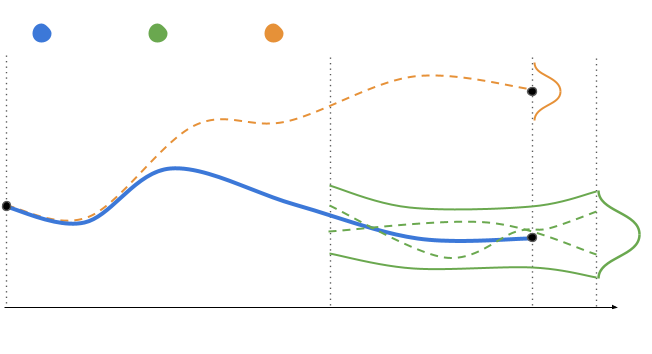}
    };

    \node[text=black, font=\normalsize, anchor=west] at (1.1,5.92) {True};
    \node[text=black, font=\normalsize, anchor=west] at (3.2,5.92) {PAINT};
    \node[text=black, font=\normalsize, anchor=west] at (5.4,5.92) {Autoregressive};
    
    \node[text=black, font=\normalsize, anchor=west] at (-0.075,0.5) {$0$};
    \node[text=black, font=\normalsize, anchor=west] at (5.65,0.5) {$t-h$};
    \node[text=black, font=\normalsize, anchor=west] at (9.65,0.5) {$t$};
    \node[text=black, font=\normalsize, anchor=west] at (10.5,0.5) {$t+n$};
\end{tikzpicture}
}
\caption{
Conceptual depiction of trajectories from the ground truth (blue), our parallel-in-time method PAINT (green) and a naive autoregressive model (orange).
Dashed lines represent sampled trajectories from the probability distribution (solid lines) given by the respective model. 
\\
The models ingest real world measurements (not depicted) at test time to reconstruct the true trajectory.
Autoregressive models predict the probability of the next state given an autoregressive state additional to real world measurements.
They typically depend on an initial state and are prone to drift off due to
\emph{over-reliance on their autoregressive state} (see Section \ref{sec:overreliance}).
In contrast, parallel-in-time models are ``on-trajectory'' and do not depend on an initial state.
They take a window of measurements and predict a distribution of subtrajectories which can be sampled and used for state and uncertainty estimation. 
}
\label{fig:drift}
\end{figure}

\section{Introduction}
\label{sec:introduction}

Modern artificial intelligence is increasingly integrated into the modeling, monitoring, and control of complex dynamical systems.  
A key development in this domain are digital twins \citep{grieves2014digital}: virtual models that run alongside a physical system, using live sensor data to update its predictions.
Unlike standard simulations, a digital twin continuously adjusts its internal state to match the real system’s behavior, thereby enabling better monitoring and automated decision-making.

Neural surrogates are neural network based predictive models which emulate expensive, 
high-fidelity simulations and offer fast predictions while maintaining reasonable accuracy.
Recent works have already demonstrated remarkable results on computational fluid dynamics (CFD)
\citep{li2022oformer,alkin2024universal,wu2024transolver,alkin2025ab_upt}, as well as on particle based systems as surrogate models for discrete element methods \citep{alkin2024neuraldem}.

However, transforming neural surrogates into neural twins is a challenging endeavor.
The current generation of neural surrogates can neither ingest nor adapt to real-time measurements during inference. 
Therefore, we asked the question: \emph{How to design neural surrogates that adapt at test time using real-time measurements?}

We do not claim to solve this problem entirely. 
However, we present insights and a new approach that advances the discussion. 
We introduce 
\textbf{Pa}rallel-in-t\textbf{i}me \textbf{N}eural \textbf{T}wins (\textbf{PAINT}), 
a family of methods for reconstructing dynamical systems from measurements. 
At the core of PAINT is training a generative neural network to model the joint conditional probability of system states given the measurements. 
At inference, it reconstructs a sequence of states from measurements using a sliding-window approach.
This makes our method interpretable and avoids any dependence on an initial condition, which constitutes an easily overlooked component in autoregressive models.

Mathematically, we show that PAINT enables accurate state estimation from measurements due to being \emph{on-trajectory}, i.e., staying close to the true trajectory even in presence of prediction errors.
We also show that autoregressive models may possess this property under certain conditions, but not in general.
Our analysis also sheds light on a phenomenon we call 
\emph{over-reliance on the autoregressive state}.
It describes the problem of an autoregressive model over-relying on an off-trajectory state, whereas an ideal model would recognize such drift and reduce the state's influence on the prediction.

Empirically, we implement FlowPAINT, an instance of PAINT based on Flow Matching. 
FlowPAINT is evaluated on a dataset of state-of-the-art large eddy simulations with varying Reynolds numbers in 2D.
This dataset is especially challenging due to the chaotic dynamics of the turbulent fluid.
The results confirm the generalization capabilities of our method.

We show -- theoretically and empirically -- how to design neural twins which stay on-trajectory for arbitrary many timesteps. 
Our main contributions are summarized as follows:
\begin{itemize}
    \item To the best of our knowledge, we are the first to train an on-trajectory neural twin on a high-dimensional fluid-dynamics problem.
    Our method is interpretable, has an inherent measure of uncertainty and makes no prior assumptions about the initial state.%
    \item Mathematically, we show that our method is on-trajectory, i.e., stays close to the true trajectory for arbitrarily long rollouts. Our analysis also sheds light on pitfalls when using autoregressive models with sparse measurements.
    \item We empirically demonstrate the effectiveness of our method on a challenging 2D turbulent fluid dynamics problem.
\end{itemize}

\section{Background and related work}

\paragraph{Notation.}
Vectors and matrices are denoted in bold, e.g., $\bm{x} \in \mathbb{R}^n$ and $\bm{A} \in \mathbb{R}^{m \times n}$.
General sequences such as $[k, \dots, t]$ are denoted in brackets as $[k,t]$.
We may also use brackets as subscripts over sequences, e.g.,  $\bm{x}_{[k,t]} = [\bm{x}_k, \dots, \bm{x}_t]$.
True or ground-truth probability distributions are denoted $p(\cdot)$, while the probability of a random variable $\bm{X}$ taking a specific value $\bm{x}$ is written as $p(\bm{X}=\bm{x})$ or, in shorthand, as $p(\bm{x})$.
Functions or distributions parameterized by neural networks use subscripts $\theta$ for their parameters, e.g., $f_\theta$ or $p_\theta(\cdot)$.

\subsection{Scalable probabilitic neural surrogates}

\paragraph{Scalable neural surrogates.}
A surrogate model, or simply a surrogate \citep{forrester2008engineering}, is a simplified computational model designed to approximate a more intricate, computationally intensive system, such as a high-fidelity simulation or a physical experiment.
Modern neural surrogates based on deep neural networks rely heavily on the transformer architecture \citep{vaswani2017attention}. They have been successfully scaled from a few thousand \citep{li2022oformer,wu2024transolver,alkin2024upt} to millions of input dimensions \citep{alkin2025ab_upt}.
Often these architectures are combined with convolutional layers \citep{lecun1989_cnn,kipf2016graphCNN} as learned encoders and decoders \citep{kohl2023benchmarkingARDiff,alkin2025ab_upt}.

\paragraph{Generative modeling of PDEs.}
Over the past years, a powerful group of related generative modeling frameworks were introduced around the works of Diffusion \citep{sohl2015diffusion,ho2020ddpm}, Flow Matching \citep{liu2022flow,lipman2022flow} and stochastic interpolants \citep{albergo2022building,albergo2023stochastic,gao2025diffusionmeetsflow}.
These frameworks have set new state-of-the-art results when modeling PDEs \citep{sestak2025lamslide,kohl2023benchmarkingARDiff}.
Additionally diffusion models seem to possess beneficial properties for modeling high frequencies \citep{lippe2023pde_refiner}, leading to more accurate and physically plausible solutions.

\subsection{Reconstruction and control of dynamical systems}

\paragraph{Bayesian inference.}
\label{sec:bayesian_inference}
Updating model states by incorporating measurement data is known as bayesian inference \citep{sarkka2023bayesian_smoothing}. 
Bayesian inference methods can be broadly categorized into 
(1)~Filtering: $\; p(\bm{x}_t \mid \bm{m}_{[0,t]})$, (2)~Prediction: $\; p(\bm{x}_{t+n} \mid \bm{m}_{[0,t]})$, and (3)~Smoothing: $\; p(\bm{x}_t \mid \bm{m}_{[0,T]})$, where $\bm{x} \in \mathbb{R}^{d_x}, \bm{m} \in \mathbb{R}^{d_m}, t \in [0, T]$ and $n \in \mathbb{N}$.
Classical approaches to bayesian inference take simplifying assumptions and usually arrive at analytical solutions. 
These approaches date back to the seminal works on Wiener and Kalman filters \citep{wiener1949filter,kalman1960}.
However, neither classical approaches nor hybrid approaches using neural networks
\citep{krishnan2015deep_kalman_filters,karl2016deep_variatonal_bayes,review_Bayes_filters} have been successfully trained on tens of thousands or more input dimensions. 
They are therefore impractical for our use case.

\paragraph{Reconstructing flows from sparse measurements.}
Prior art has explored flow reconstruction from sparse measurements. 
These measurements are usually assumed to come from Particle Tracking Velocimetry (PTV), probes in the walls or at informative locations.
\\
PAINT differentiates and/or generalizes most prior works by 
(1) learning a distribution of states instead of point estimates \citep{dang2024flronet,chaurasia2024ptv_pinn,hosseini2024flow_pinn,toscano2024KAN_pinn_reconstruct}
or (2) being scalable and end-to-end trainable \citep{sun2020reconstruction_bayesian}.
or (3) using temporal additionally to spatial information \citep{guemes2022super_res_gans,cuellar2024gan_3d_reconstruction,oommen2025flows_w_genai,kim2021unsupervised_flow,hemant2025flow_matching_near_wall}.
\\
Some works \citep{li2024learning_dps,du2024confild,amoros2026_dps_sparse_thuerey} used diffusion posterior sampling \citep{jalal2021DPS,chung2022dps}
to reconstruct flow fields from measurements based on a prior diffusion model. 
However, the quality of the reconstruction still falls short of conditional diffusion models \citep{gupta2024_dps_intractable}.

\paragraph{Control of dynamical systems.}
Instead of reconstructing the dynamical system from measurements, prior work controlled the dynamical system directly from the measurements using reinforcement learning \citep{ozan2025assimilated_MBRL,lagemann2025hydrogym}.
While these are promising directions, the lack of interpretability is a major obstacle for a broader adoption and a limitation regarding applicability.

\subsection{Computational Fluid Dynamics}

The dynamical system we will use in our experiments (see Section \ref{sec:dataset}) is a 2D fluid dynamical system in the turbulent regime.
Therefore, we give a brief introduction to computational fluid dynamics.

\paragraph{Incompressible Navier-Stokes.}
The fluid mechanics problem under the continuum assumption is governed by the Navier–Stokes equations in their incompressible form as 
\begin{equation}
\begin{aligned}
& \frac{\partial \bm{u}}{\partial t} + \bm{u} \cdot \nabla \bm{u} = - \nabla p + \mu \Delta \bm{u} + \bm{f_e}, & \nabla \cdot \bm{u} = 0
\end{aligned}
\end{equation}
where $\bm{u}$ denotes the velocity, $p$ denotes the pressure, $\mu$ denotes the viscosity and $\bm{f_e}$ denotes any kind of external body forces. 
These equations are complemented by appropriate boundary and initial conditions.
This constitutes a system of equations of highly complex, second-order nonlinear PDEs with no known general analytical solution.
Therefore, they are typically solved numerically.

\paragraph{Precision–computation tradeoff.}  
Numerical solutions of the Navier–Stokes equations involve a tradeoff between accuracy and computational cost. Direct numerical simulation (DNS) resolves the full flow field but is prohibitively expensive for most practical flows. Large-eddy simulation (LES) resolves only the largest, energy-containing scales, while Reynolds-averaged Navier–Stokes (RANS) approaches provide only mean flow fields under strong modeling assumptions \citep{davidson2015turbulence, frohlich2006large}. 
LES represents a compromise between fidelity and computational cost within this spectrum of methods and was therefore used for the experiments in this paper.

\section{Method}

\definecolor{frozen_color}{RGB}{33,102,172}
\definecolor{testtime_color}{RGB}{178,24,43}
\begin{figure}[t!]
    \centering
    \resizebox{0.5\textwidth}{!}{
        \begin{tikzpicture}[
            node distance=5cm, 
            auto
            ]
            \node[circle, draw, minimum size=2.5cm, align=center, line width=1.1pt] (R) {Dynamical\\System};
            \node[circle, draw=frozen_color, text=frozen_color, minimum size=2.5cm, below left of=R, align=center, line width=1.1pt] (C) {Simulation\\dataset};
            \node[circle, draw, minimum size=2.5cm, below right of=R, line width=1.1pt] (N) {Neural Twin};
            \draw[->, >=latex, color=frozen_color, line width=1.1pt] (R) -- node[above left] {(1) Inform} (C);
            \draw[->, >=latex, color=frozen_color, line width=1.1pt] (C) -- node[below] {(2) Train} (N);
            \draw[<-, >=latex, color=testtime_color, line width=1.1pt] (N) to[out=135, in=-45] node[above right] {(3) Measurements} (R); %
            \draw[->, >=latex, dashed, color=testtime_color, line width=1.1pt, >=latex, scale=2] (N) to[out=45, in=45] node[right=2pt, above right=2pt] {(4) Control} (R);
        \end{tikzpicture}
    }
    \caption{
    Interaction between a real-world dynamical system, the simulations dataset and a neural twin.
    The training phase is depicted in blue, the inference (test) phase in red and nodes involved in both phases in black. Dashed lines indicate optional steps.
    The workflow is structured in three phases:
    \textbf{(1)} The dynamical systems informs the design of the simulations and a diverse training dataset is generated.
    \textbf{(2)} The generated dataset is used for training the neural twin.
    \textbf{(3)} During inference, the neural twin integrates real-time measurement data to remain on-trajectory.
    \textbf{(4)} The state estimation of the neural twin can optionally be used for decision-making.}
    \label{fig:twin_triangle}
\end{figure}

\subsection{A neural twin framework for dynamical systems}

We envision neural twins in a four-step process, depicted in Figure \ref{fig:twin_triangle}.
First, design knowledge of the dynamical system is used to create the simulations. This ensures that the simulations reflect physical constraints and key behaviors. 
Second, a diverse training dataset is generated by exploring a broad spectrum of trajectories. 
This dataset is used for training a neural twin.
Third, at test time the neural twin dynamically incorporates real-time measurement data, allowing it to stay aligned with actual system trajectories and support real-time decision-making. 
Fourth, via the state estimation of the neural twin, one can optionally control the dynamical system, e.g. via rule-based methods.
The approach combines domain expertise with adaptive learning, enabling both accuracy and agility in complex environments.

\subsection{Parallel-in-time Neural Twins}
\label{sec:PAINT}

\paragraph{Problem description.}
Consider a transient dynamical system where the state space is a compact subset $\mathcal{X} \subset \mathbb{R}^{d_x}, d_x \in \mathbb{N}$ and $\bm{x}_t \in \mathcal{X}$ denotes the system state at time $t$. 
We assume a $d_m \in \mathbb{N}$ dimensional compact measurement space $\mathcal{M} \in \mathcal{R}^{d_m}$.
At each timestep $t \in \mathbb{N}$ the system provides a measurement $\bm{m}_t$ from the emission model $p_e$, where $\bm{m}_t \sim p_e(\bm{m}_t \mid \bm{x}_t)$.

\paragraph{PAINT.}
The goal of our generative model $ p_\theta $ is to sample trajectories from $p(\bm{x}_{[0,t]}) \mid \bm{m}_{[0,t]})$.
At the core of our method is \emph{joint modeling} of this distribution.
In contrast, \emph{autoregressive models} factorize this distribution which may lead to drifting (see Section \ref{sec:theory}).
\\
As the number of timesteps may become very large, PAINT models the joint probability over a window (see Figure \ref{fig:drift}) of width $h$ and may optionally predict $n$ timesteps into the future:
\begin{align}
    \label{eq:paint}
    p_\theta(\bm{x}_{[t-h,t+n]} \mid \bm{m}_{[t-h,t]}).
\end{align}
In this setup, future measurements can inform past states to obtain temporal consistency.
Furthermore, there is no reliance on the existence or an assumption about an initial condition. 
\\
PAINT is agnostic to the generative modeling framework and the neural network architecture.
Common building blocks may be transformers, MLPs or CNNs \citep{lecun1989_cnn}.

\subsection{Theoretical Considerations}
\label{sec:theory}

\paragraph{Autoregressive models} factorize the joint distribution as a product of conditionals.
Assuming the system is Markovian and conditioned on a sequence of variables $\bm{m}_{[0,t]}$, we obtain:
\begin{align}
    p(\bm{x}_{[0,t]} \mid \bm{m}_{[0,t]}) = p(\bm{x}_0 \mid \bm{m}_{[0,t]}) \cdot \prod_{i=1}^t p(\bm{x}_i \mid \bm{x}_{i-1}, \bm{m}_{[0,t]}),
\end{align}
where the autoregressive model learns the conditional probability $p_\theta(\bm{x}_t \mid \bm{x}_{t-1}, \bm{m}_{[0,t]})$.
However, in practice, due to finite compute but potentially arbitrarily long sequences, a shorter measurement window of size $h \in \mathbb{N}$ is used as follows
\begin{align}
    \label{eq:AR_model}
    p_\theta(\bm{x}_t \mid \bm{x}_{[t-k, t-1]}, \bm{m}_{[t-h,t]})
\end{align}

\paragraph{Autoregressive model drift.}
Our analysis is leaned on the analysis of error growth of dynamical systems \citep{orrell2005estimating_error_growth} and key observations made in \citep{hess2023generalized_teacher_forcing}.
We assume a fully differentiable neural network,
which learns the probability distribution $p_\theta(\bm{x}_{t}||\bm{x}_{t-1}, \bm{m}_{[0,t]})$.
To sample from this probability distribution $\bm{x}_{t} \sim p_\theta(\bm{x}_{t}||\bm{x}_{t-1}, \bm{m}_{[0,t]})$  we rewrite the neural network to a deterministic function with a stochastic input $\eta_t$: 
\begin{align}
    \bm{x}_{t} &= f_\theta(\bm{x}_{t-1}, \bm{m}_{[0,t]}, \eta_t)
    \quad\quad \eta_t \sim p(\eta)    
\end{align}
The Jacobian is then defined as $
\bm{J}_t \coloneqq \partial f_\theta/\partial \bm{x}_t
$.
In contrast to \citet{orrell2005estimating_error_growth}, this is a discrete random ordinary differential equation (RODE), i.e., an ODE with a random input at each timestep.
\\
To show the model drift from small perturbations, we fix a path of the RODE by fixing $\eta_{[0,t]}$.
For simplicity, we then write $F_\theta^t(\bm{x}_t) = f_\theta(\bm{x}_{t-1}, \bm{m}_{[0,t]}, \eta_t)$ where we absorb the measurement $\bm{m}_{[0,t]}$ and the stochastic input $\eta_t$ into the function, such that $\bm{x}_{t} = F_\theta^t(\bm{x}_{t-1})$.
Hence for the generation of a state $\bm{x}_t$ from a previous state $\bm{x}_k$ it follows from function composition:
\begin{align}
        \bm{x}_{t} 
        &= F_\theta^{t} \circ F_\theta^{t-1} \circ ... \circ F_\theta^{k+1}(\bm{x}_{k}) 
        = F_\theta^{[k+1,t]}(\bm{x}_{k})
    \end{align}
The first-order Taylor approximation of a small input perturbation at timestep $k<t$ is:
\begin{align}
    F_\theta^{[k+1,t]}(\bm{x}_{k} + \epsilon)
    &= F_\theta^{[k+1,t]}(\bm{x}_{k}) + 
    \epsilon \prod_{i=k+1}^t J_i(\bm{x}_{i-1}) + o(\epsilon^2)
\end{align}
This means, a deviation at timestep $k$ induces a deviation multiplied by the \emph{Jacobian product series} at timestep $t$.
This has also been experimentally explored by \citet{hess2023generalized_teacher_forcing}, who include visualizations of the Jacobian product series for chaotic dynamical systems.

\paragraph{On-trajectory.}
In the following we will formalize the notion of a model being ``on-trajectory''.
For this, we rely on Assumption~\ref{ass:measurement_decay}, which states that the predictive relevance of past measurements decays within a finite window $h$. 
\begin{assumption}\label{ass:measurement_decay}
\textbf{Temporal decay of measurement information}: 
For the given dynamical system, there is a finite window size $h \in \mathbb{N}$, such that almost all information from past measurements comes from measurements within this window.
Formally, for all $\epsilon > 0$ and all $t \in \mathbb{N}$ there exists a $h \ll t$ such that
$||p(\bm{x}_t \mid \bm{m}_{[t-h,t]}) - p(\bm{x}_t \mid \bm{m}_{[0,t]})|| < \epsilon$.
\end{assumption}
Next, we proceed with our definition of on-trajectory. 
\begin{definition}\label{def:on_traj}
\textbf{On-trajectory}: 
Assuming unbounded model size, compute and data, a model is \emph{on-trajectory} if the modeled distribution is arbitrarily close to the true distribution.
Formally, for all $\epsilon > 0$ and all $t \in \mathbb{N}$, $\bm{x} \in \bm{\mathcal{X}}$ there is model, data and compute such that:
$||p_\theta(\bm{X}_t = \bm{x} \mid \bm{m}_{[0,t]}) - p(\bm{X}_t = \bm{x} \mid \bm{m}_{[0,t]})|| < \epsilon$. 
\end{definition}

\paragraph{Parallel-in-time models are on-trajectory under Assumption 
\ref{ass:measurement_decay}
.}
We derive this in Appendix~\ref{app:proof_paint_on_traj}.
Intuitively, for each $\epsilon$ we can choose the window $h$ large enough to guarantee arbitrary closeness.
In contrast, autoregressive models are in general not on-trajectory.
As a constructed counterexample, consider an underlying chaotic system where the model learns to ignore the measurements
(for further discussion, see \citet{hess2023generalized_teacher_forcing} and Appendix \ref{app:proof_AR_off_traj}).

\subsection{Over-reliance on the autoregressive state}
\label{sec:overreliance}

The analysis in Section \ref{sec:theory} highlights a fundamental problem we call over-reliance on the autoregressive state.
Intuitively, it states that for a naive autoregressive model the autoregressive state plays an ambiguous role:
If the state is on-trajectory, it helps the model as it provides useful information for the next state.
If the state is off-trajectory, the model may still make use of it for prediction and drift off further. 

\paragraph{Measurement informativeness and uncertainty modeling.}
In principle, the only way for the model to correct an off-trajectory autoregressive state is with informative measurements 
(this excludes edge cases, e.g., where the model converges to a global steady state).
Ideally, a model would detect when the autoregressive state drifts off and decrease its influence on the prediction.
This can theoretically be achieved by a model which models the state uncertainty explicitly. However reliable uncertainty modeling for high-dimensional state spaces and long trajectories is a challenging open research problem.

\paragraph{Training paradigm.}
Over-reliance on autoregressive state is also influenced by the training strategy. 
Autoregressive models are usually trained with teacher-forcing, where the ground-truth previous state $\bm{x}_t$ is used to predict the next state $\bm{x}_{t+1}$.
As these states are highly correlated, the model might learn to over-rely on $\bm{x}_t$ and under-rely on $\bm{m}_{t+1}$.
This could potentially be mitigated by training techniques such as generalized teacher forcing \citep{hess2023generalized_teacher_forcing} or unrolled training \citep{kohl2023benchmarkingARDiff}.

\paragraph{Connection to Jacobian.}
How much information is taken from the autoregressive state is indicated by $\bm{J}_t(\bm{x}_t)$.
To avoid autoregressive error accumulation, the Jacobian product series should ideally have a small norm.
Notice that parallel-in-time models predict $\bm{x}_{[t-h,t]} = f_\theta(\bm{m}_{[k,t]}, \eta)$ without any recurrence.
Consequently the Jacobians are all zero-matrices.

\section{Experiments}
\label{sec:experiments}

\subsection{Dataset and evaluation}
\label{sec:dataset}

\paragraph{Turbulent single jet dataset.}
We generate a CFD dataset using the incompressible, pressure-based solver pimpleFOAM in OpenFOAM~8 \citep{opencfd2009open}, with subgrid scales modeled by the Smagorinsky LES model \citep{pope2001turbulent, frohlich2006large}. Simulations are performed on a structured mesh, and for training purposes, a subregion is extracted and interpolated onto a regular 128 x 128 grid. %
Since the state-space represents 2D velocities, we will write $\bm{u}_t$ instead of $\bm{x}_t$ with $\bm{u}_t \in \mathbb{R}^{d_1 \times d_2}$.
Appendix \ref{app:dataset} provides the train/val/test splits across Reynolds numbers as well as further details about the dataset.

\paragraph{Physical coherence.}
Due to the chaotic and rapidly diverging nature of turbulent flows, comparing a single predicted frame or a short sequence to the ground truth is not meaningful. 
Research in turbulence instead focuses on the flow's inherently reproducible statistical properties \citep{pope2001turbulent, davidson2015turbulence}. 
Given a possibly infinitely long trajectory $\bm{u}_{[0,t]}$, we follow a common notation in fluid dynamics and decompose $\bm{u}_{[0,t]}$ into a \emph{time-averaged mean} component $\overline{\bm{u}} \coloneqq \lim_{t \rightarrow \infty} \frac1t \sum_{i=1}^t \bm{u}_i$ and a turbulent fluctuation component $\bm{u}'_{[0,t]}$, such that $\bm{u}_{[0,t]} = \overline{\bm{u}} + \bm{u}'_{[0,t]}$.
The \emph{variance over time} is denoted as $\overline{\bm{u}'^2} \coloneqq \lim_{t \rightarrow \infty} \frac1t \sum_{i=1}^t (\bm{u}_i - \overline{\bm{u}})^2$.
Further, we define \(E(k)\) as the \emph{mean kinetic energy spectrum}.
While these metrics are defined over an infinite-time, in practice they are computed over a large, finite horizon.

\begin{wrapfigure}[8]{r}{0.3\textwidth} %
    \vspace{-12pt}
    \centering
    \begin{subfigure}[t]{0.4\linewidth} %
        \includegraphics[width=\linewidth]{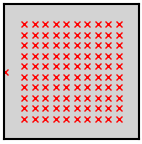}
        \caption{Grid.}
        \label{fig:grid}
    \end{subfigure}
    \hspace{0.1cm}
    \begin{subfigure}[t]{0.4\linewidth}
        \includegraphics[width=\linewidth]{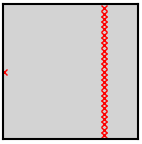}
        \caption{Vertical.}
        \label{fig:vertical}
    \end{subfigure}
    \caption{Probe positions.}
    \label{fig:probes}
\end{wrapfigure}

\paragraph{Probe constellations.}
The model is trained using 25 random probe points that are resampled for each data point and every iteration.
At inference, we investigate the two fixed probe point constellations depicted in Figure \ref{fig:probes}. 
\emph{Grid}: A 10×10 grid of 100 probe points evenly spaced across the domain (see Figure \ref{fig:grid}).
\emph{Vertical}: A linear arrangement of 25 probe points positioned at the 3/4 axial location of the domain (see Figures \ref{fig:vertical}).
In both configurations, an additional probe point is placed at the center of the inlet of the jet to capture the inlet velocity.

\begin{figure*}[ht!]
    \centering
    \includegraphics[width=0.85\linewidth]{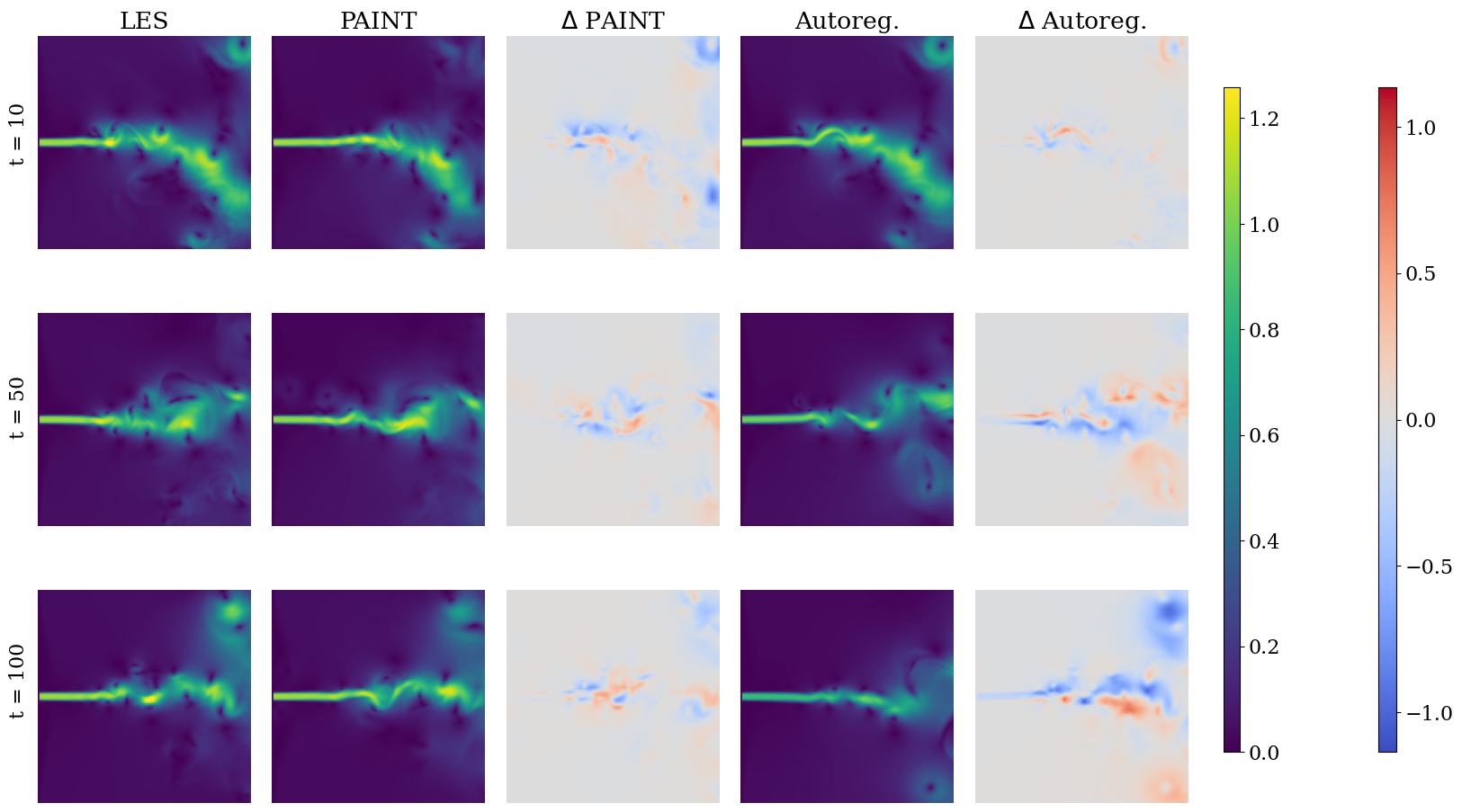}
    \caption{Ground truth vs. predicted states at different timesteps for 
    $Re = 2100$
    using the \emph{vertical} probe constellation.
    The prediction labeled \emph{PAINT} corresponds to FlowPAINT 16/8. Unlike the autoregressive U-Net, PAINT does not accumulate errors over time, whereas the U-Net exhibits a pronounced temporal drift. All values are normalized by the mean inlet velocity.}
    \label{fig:snapshots_comparison_paint_autoreg}
\end{figure*}

\subsection{Models}
\label{subsec:models}

\paragraph{PAINT via data-coupled Flow Matching.}
PAINT is agnostic to the generative modeling paradigm, as long as it allows for conditional generation.
We therefore implement FlowPAINT as a concrete instantiation of PAINT using a common Flow Matching implementation \citep{lipman2024flowmatchingguidecode}.
Additionally, the unmasking of non-measurement points in the domain is modeled as data-coupled distribution matching \citep{albergo2023datacoupled}.
Notably, the masking ratio of our method is very high (25 of 16K pixels are unmasked).
In early experiments we observed that the gradient signal was too weak for the vicinity of the probes.
Therefore, we introduce spatial weighting to the Flow Matching loss, which puts higher weight on pixels in the neighborhood of the probe points. 
\\
PAINT is also agnostic to the neural network architecture, we therefore use a Diffusion Transformer (DiT) \citep{peebles2023DiT} with convolutional layers for patching and unpatching.
Further experimental details can be found in Appendix \ref{app:architecture}.

\paragraph{Autoregressive Baselines.}
PAINT is fundamentally different to autoregressive models, because it predicts a sequence of consecutive timesteps at once.
In contrast, an autoregressive model predicts the sequence by regressing on its own previous output.
Therefore the comparison to an autoregressive model, is inherently difficult.
In this work we tested 2 baselines.
First, we compare PAINT to the same DiT architecture.
However, this setting did not lead to stable rollouts (for further details see Appendix~\ref{app:architecture}).
\\
To provide a more insightful baseline, we compare PAINT to an established autoregressive UNet baseline from \citet{kohl2023benchmarkingARDiff}.
The UNet uses convolutional and linear attention \citep{shen2021_linear_attn} blocks.
We follow their proposed paradigm and condition on two past autoregressive states and on the probe values by concatenating the mask and probe values to the channels.
The model sizes were selected to closely match the number of parameters, with FlowPAINT comprising 19.8M parameters and the autoregressive UNet 20.6M. 
To promote a fair comparison, the optimization hyperparameters and batch sizes were carefully chosen. 
These and other architectural decisions are explained in further detail in Appendix~\ref{app:architecture}.

\paragraph{Probe conditioning and future prediction.}
Across the experimented models, different measurement histories $h$ and future predictions $n$ are possible (also see Equations \ref{eq:paint} and \ref{eq:AR_model}).
We add the postfix $h+1/n$ to distinguish between these hyperparameters.
E.g. ``FlowPAINT 16/8'' learns the probability $p(\bm{x}_{[t-15,t+8]} \mid \bm{m}_{[t-15,t]})$ of 16 measurement timesteps and predicting 8 timesteps steps into the future.
Similarly for autoregressive models, ``Autoregressive 16/0'' denotes the UNet architecture from \citet{kohl2023benchmarkingARDiff} which learns $p(\bm{x}_{t} \mid \bm{x}_{[t-2,t-1]}, \bm{m}_{[t-16,t]})$.
Our main results are reported in the 16/0 setting for the best comparability.
More settings are reported in the ablations setting in Appendix \ref{app:results} and Table \ref{tab:results_appendix}.

\subsection{Results}

\begin{table*}[ht!]
\centering
\caption{Results for physical coherence of FlowPAINT (ours) and the autoregressive UNet baseline (AR UNet).
The MAE, MSE and RMSE are taken between the ground truth and the predicted quantity.
$\overline{\mathbf{u}}$ and $\overline{\mathbf{u}'^2}$ denote the mean and variance over time.
$E(k)$ is the predicted kinetic energy spectra.
$\text{MSE}(u_{[0,t]})$ is the mean squared error between the predicted and the ground-truth trajectory as a whole.
All values are taken along 900 timesteps in the test trajectories. 
The values after $\pm$ are standard deviations from averaging over 10 seeds.
Generally, the higher Reynolds number seems to be more difficult for both models.} 
\label{tab:results}

\begin{tabular}{llcl|cccc}
\textbf{Model} & \textbf{Cond.} & \textbf{$Re$} & \textbf{Probes} & ${\text{MAE}}(\overline{\mathbf{u}}) \downarrow$  & ${\text{MAE}}(\overline{\mathbf{u}'^2}) \downarrow$  & $\text{MSE}([\mathbf{\mathbf{u_{[0,t]}}}]) \downarrow$ & $\text{RMSE}(E(k)) \downarrow$ \\
 \hline

& \multicolumn{1}{l}{} &   &  & {\fontsize{8}{11}\selectfont $ \times 10^{-3} $} & {\fontsize{8}{11}\selectfont $ \times 10^{-5} $} & {\fontsize{8}{11}\selectfont $ \times 10^{-5} $} & {\fontsize{8}{11}\selectfont $ \times 10^{-6} $} \\
 \hline
\multirow{2}{*}{FlowPAINT} & 16/8  &  \multirow{3}{*}{2100} & \multirow{3}{*}{grid} & $1.2 \pm 1.6$ & $9.9 \pm 21.7$ & $8.1 \pm 3.6$ & $6.9$ \\
 & 16/0 &  & & $1.3 \pm 2.6$ & $8.7 \pm 15.1$ & $10.8 \pm 3.9$ & $6.3$ \\
\multirow{1}{*}{Autoreg.} & 16/0  &  &  & $8.2 \pm 13.7$ & $93.0 \pm 103.2$ & $155.0 \pm 50.1$ & $66.3$ \\
 \hline
\multirow{2}{*}{FlowPAINT} & 16/8  & \multirow{3}{*}{2100} & \multirow{3}{*}{vertical} & $2.2 \pm 3.1$ & $30.2 \pm 38.4$ & $49.0 \pm 21.0$ & $20.1$ \\
& 16/0  & & & $3.0 \pm 3.2$ & $30.0 \pm 41.6$ & $68.2 \pm 25.3$ & $20.5$ \\
\multirow{1}{*}{Autoreg.} & 16/0  &  & & $7.4 \pm 12.5$ & $87.6 \pm 96.1$ & $156.2 \pm 50.9$ & $62.4$ \\
 \hline
\multirow{2}{*}{FlowPAINT} & 16/8  & \multirow{3}{*}{1100} & \multirow{3}{*}{grid} & $0.9 \pm 1.3$ & $2.2 \pm 5.5$ & $2.0 \pm 1.0$ & $2.0$ \\
& 16/0  & & & $1.2 \pm 2.8$ & $6.2 \pm 10.7$ & $4.6 \pm 1.6$ & $4.6$ \\
\multirow{1}{*}{Autoreg.} & 16/0  &  & & $3.7 \pm 6.3$ & $16.0 \pm 24.1$ & $29.9 \pm 8.8$ & $12.3$ \\
 \hline
\multirow{2}{*}{FlowPAINT} & 16/8  & \multirow{3}{*}{1100} & \multirow{3}{*}{vertical} & $1.1 \pm 1.6$ & $6.3 \pm 11.1$ & $8.9 \pm 3.4$ & $3.7$ \\
  & 16/0   & & & $3.0 \pm 5.8$ & $8.2 \pm 15.8$ & $24.2 \pm 6.5$ & $9.9$ \\
\multirow{1}{*}{Autoreg.} & 16/0  &  & & $3.7 \pm 6.1$ & $15.9 \pm 23.9$ & $29.5 \pm 8.8$ & $12.3$ \\
\end{tabular}
\end{table*}

\begin{figure}[t]
    \centering
    \begin{subfigure}{\linewidth}
        \centering
        \includegraphics[width=1.\linewidth]{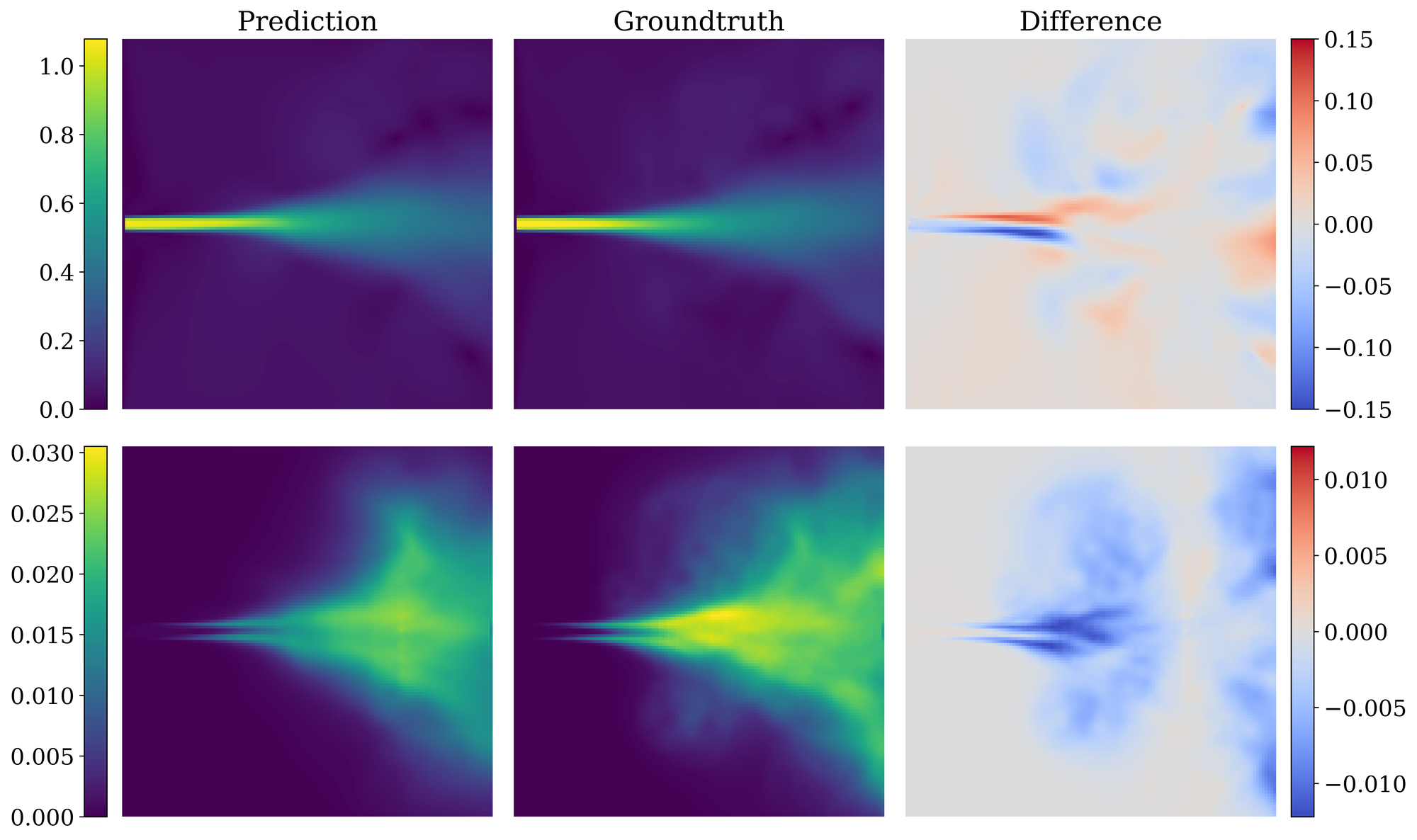}
    \end{subfigure}
    \caption{Mean velocity $\overline{u}$ (top row) and variance $\overline{u'^2}$ (bottom row) of the reconstructed $Re=2100$ test trajectory using FlowPAINT 16/8 and the \emph{vertical} probe constellation compared with the ground truth. All values are normalized by the mean inlet velocity.
    }
    \label{fig:Re2100_vertical}
\end{figure}

The quantitative results are reported in Table \ref{tab:results}. FlowPAINT outperforms the autoregressive UNet across all metrics, probe constellations and Reynolds numbers in the test set.
Further results and ablations can be found in Appendix \ref{app:results} and Table~\ref{tab:results_appendix}.

\paragraph{Recovery of the physical flow characteristics.} 
We observe the models ability to reconstruct physical plausible states from different constellations of probe points (see Figures \ref{fig:Re2100_vertical} and \ref{fig:Re2100vertical_physics}). 
The model's ability to reproduce the correct physical statistics seems to depend directly on the number and position of the sample points (as can be seen when comparing Figure \ref{fig:Re2100_vertical} with Figures \ref{fig:Re2100grid_physicsA}, \ref{fig:Re1100grid_physicsA} and \ref{fig:Re1100vertical_physicsA}).

\begin{figure*}[h]
    \centering
    \begin{subfigure}{0.4\linewidth}
        \centering
        \includegraphics[width=\linewidth]{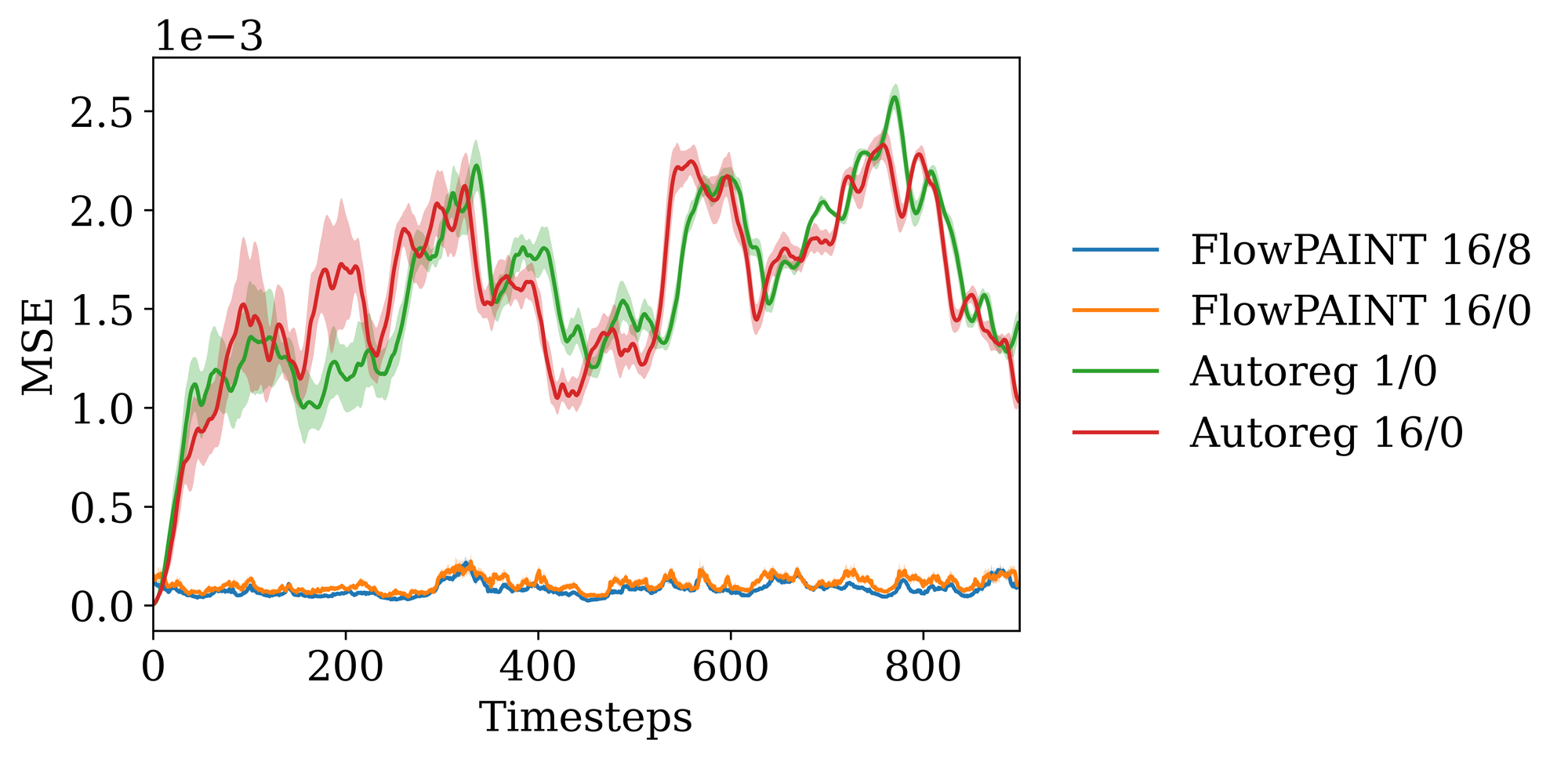}
        \caption{\emph{Grid} probe constellation.}
    \end{subfigure}
    \hspace{0.5cm}
    \begin{subfigure}{0.4\linewidth}
        \centering
        \includegraphics[width=\linewidth]{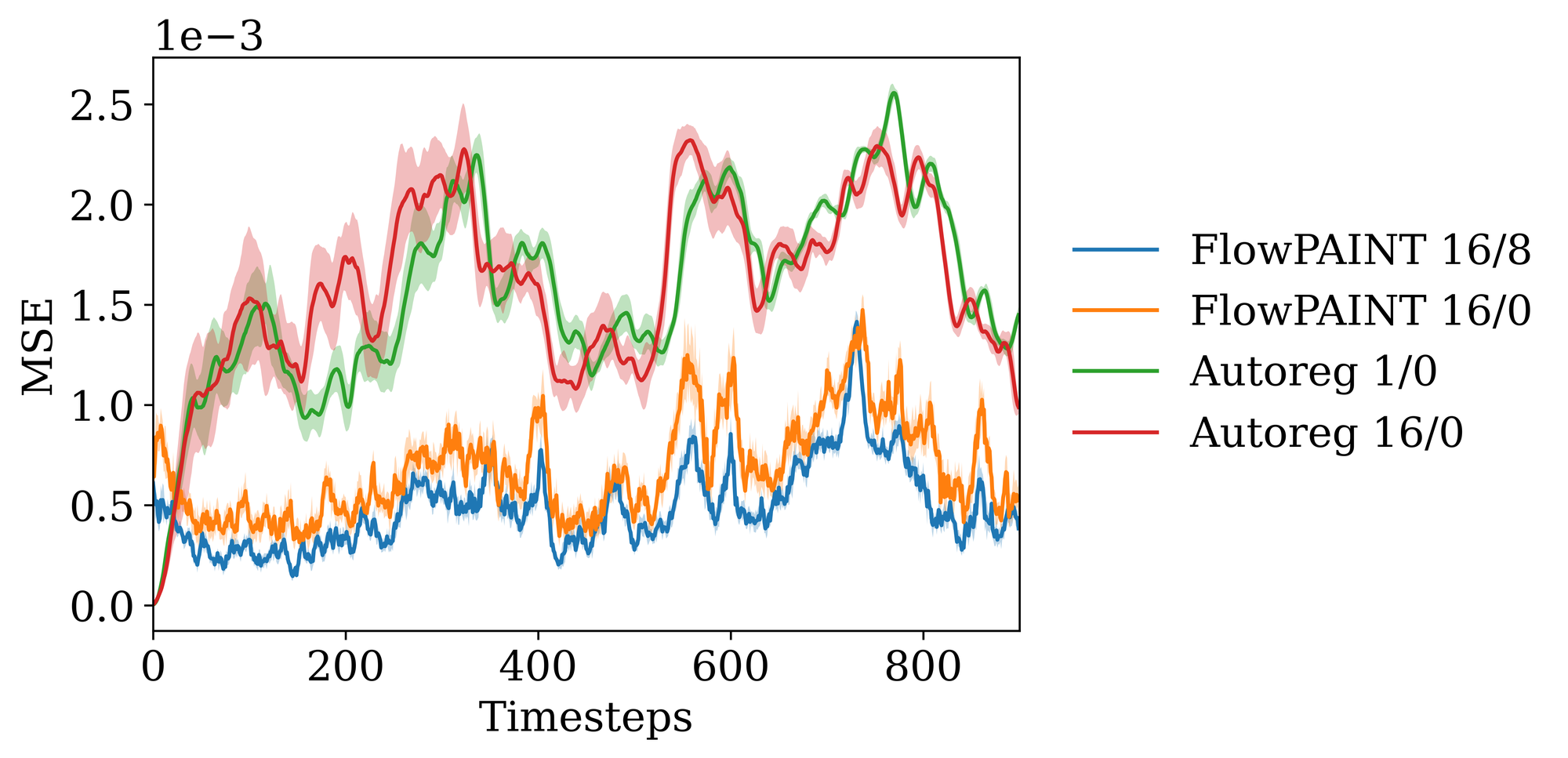}
        \caption{\emph{Vertical} probe constellation.}
    \end{subfigure}

    \caption{MSE over time of ground-truth vs. predicted trajectories for the $Re=2100$ test trajectory and different probe point constellations, comparing the autoregressive baselines with FlowPAINT. Notably, the autoregressive approach yields nearly identical error profiles for both probe constellations, failing to leverage the additional information provided by the more dense probe placements. The autoregressive model indicates over-reliance on the autoregressive state.}
    \label{fig:Re2100_MSE_over_time}
\end{figure*}

\paragraph{FlowPAINT models maintain stable error, while autoregressive UNet drifts.} 
The autoregressive approach exhibits a drift in MSE when rolled out (compare Figures \ref{fig:snapshots_comparison_paint_autoreg}, \ref{fig:Re2100_MSE_over_time} and
\ref{fig:Re1100_MSE_over_time}). 
Notably, Figure \ref{fig:snapshots_comparison_paint_autoreg} depicts a degradation of the autoregressive model's state over time, which has been analyzed in prior work \citep{kohl2023benchmarkingARDiff}.
Importantly, the autoregressive model benefits from a known, correct initial state, which is often unknown in practice.

\begin{figure}[h!]
    \centering
    \includegraphics[width=0.9\linewidth]{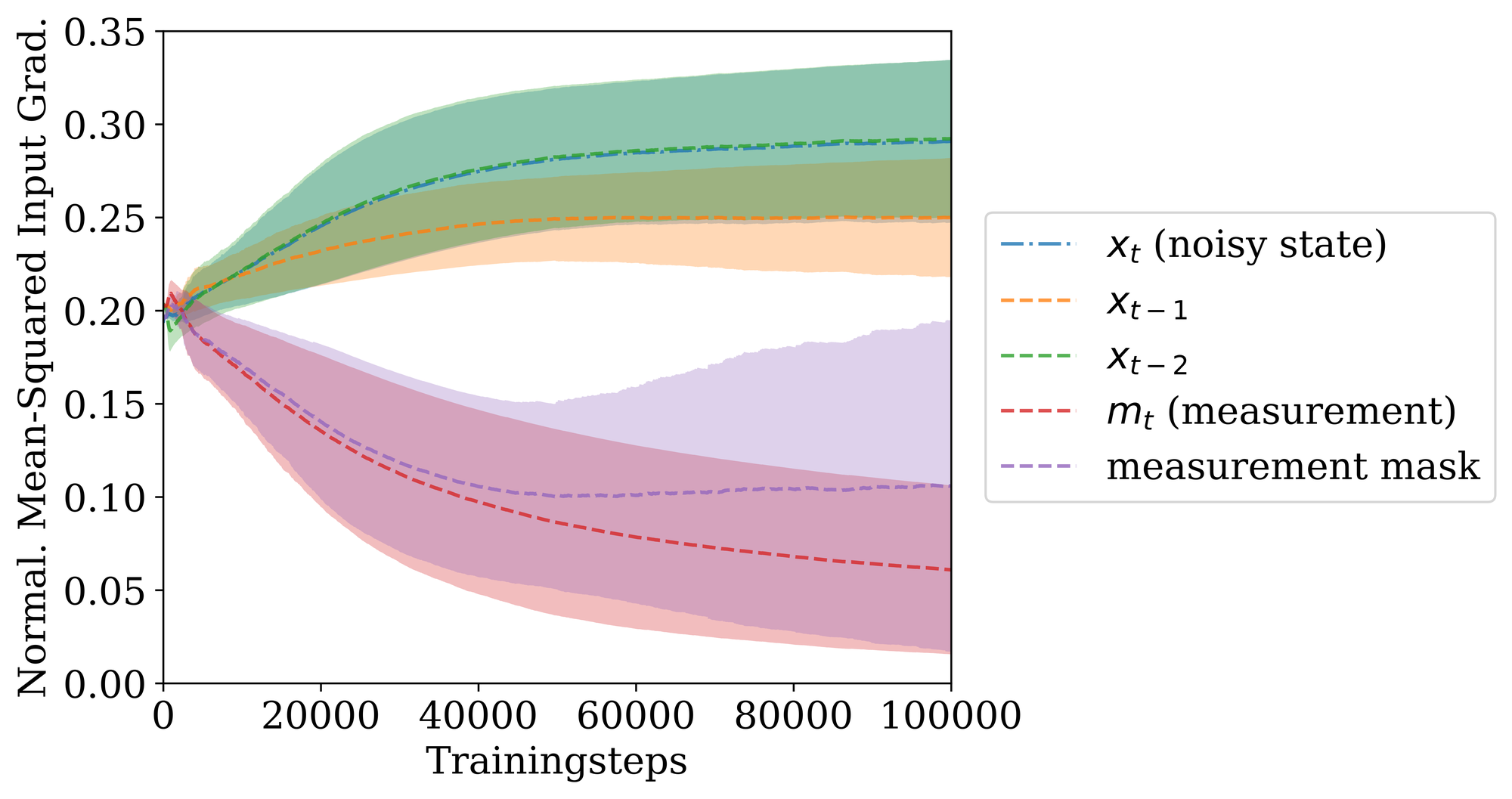}
    \caption{
    Relative input sensitivity of the loss with respect to different input channel groups, measured using mean squared gradient norms with respect to the model input. 
    Gradients are aggregated over spatial and batch dimensions and grouped by input type, as the conditioning for the autoregressive model (Autoreg. 1/0) is provided via the channel dimension. The gradients are normalized to obtain relative contributions for each conditioning signal, and the running mean (dashed line) and standard deviation (shaded area) of these quantities are visualized.
    Over the course of the training, sensitivity with respect to the measurement inputs decreases, while sensitivity to the autoregressive states increases. This could suggest, that the model increasingly relies on the more informative autoregressive context relative to the measurements.
    }
    \label{fig:tracked_gradients}
\end{figure}

\paragraph{Over-reliance on autoregressive state.}
By analyzing the gradients of the loss with respect to the model input, one can estimate how predictions depend on different conditioning signals. For the autoregressive baseline (Autoreg. 1/0), the results indicate that the model primarily relies on the autoregressive states while largely ignoring the less-informative measurements (see Figure \ref{fig:tracked_gradients}). This provides strong indication of an over-reliance on the autoregressive components. 

A related conclusion can be drawn by comparing the MSE of the two autoregressive baselines (Autoreg. 1/0 and 16/0) over rollout (see Figure \ref{fig:Re2100_MSE_over_time}), where both models exhibit very similar error trajectories. This suggests that the additional information provided by the longer measurement history in Autoreg 16/0 is not effectively used, indicating that the model overly relies on its autoregressive state.

\paragraph{Measure of uncertainty.}
One can obtain a measure for uncertainty by sampling several times and computing the variance. 
In Figure \ref{fig:std_of_prediction} it is visible that the model is most certain around the probe positions and most uncertain to the left and the right.

\paragraph{Stability through future predictions.}
Comparing FlowPAINT 16/0 with the FlowPAINT 16/8 variant highlights the benefit of incorporating future-state predictions.
This holds even if we use the same number of predicted frames per batch (see Appendix~\ref{app:results} for more details).

\paragraph{Varying the number of measurement points.}
We depict the MSE with varying probe measurements points in the grid setting in Figure  
\ref{fig:varied_probes}.
Although FlowPAINT was trained with 25 randomly resampled probe points for each datapoint, the best performance is obtained with around 100 probe points.
The autoregressive UNet's MSE stays approximately the same, even if the number of measurement points are increased.

\subsection{Limitations}

\paragraph{Sampling a sequence vs. a single state.}
PAINT may be used for sampling sequences or individual timesteps. 
Sampling a connected, temporally smooth sequence, $\bm{x}_{[t-h,t]} \sim p(\bm{x}_{[t-h,t])} \mid \bm{m}_{[t-h,t]})$ aids result interpretation (see Figure \ref{fig:sequence_Re2100}). 
Sampling individual timesteps $\bm{x}_t \sim p(\bm{x}_t \mid \bm{m}_{[t-h,t]})$, may be used for real-time state estimation in a sliding window context, albeit without the same temporal smoothness.

\paragraph{Discontinuities.} A fundamental limitation of parallel-in-time models is the inherently discontinuous temporal nature of the predicted trajectory.
This lack of temporal smoothness makes generated trajectories less interpretable and complicates the analysis of temporal properties. 
However, this limitation is acceptable in scenarios where the primary objective is to obtain an estimate of the system's current or near-future state.
In our experiments, we also observe that for reasonably small time steps, the induced discontinuities relative to an autoregressive baseline are modest (see Figure \ref{fig:discontinuities} in the Appendix \ref{app:results}).

\paragraph{Compute at training and inference.}
FlowPAINT needs substantially more training resources and is slower at inference than the autoregressive baseline.
It was trained on 12 A100 (64 GB VRAM) for 30 hours, whereas the UNet was trained on a single H100 (80GB VRAM) for 24 hours.
Notably, the UNet takes about $66$ milliseconds and FlowPAINT around $6.6$ seconds for 20 denoising steps.
Further details can be found in Appendix \ref{app:architecture}.

\section{Conclusion}

In this work, we introduced Parallel-in-time Neural Twins (PAINT), a novel data-driven framework designed to bridge the gap between static neural surrogates and adaptive digital replicas for real-time reconstruction of dynamical systems. PAINT employs a generative neural network to model the joint conditional probability of system states, enabling it to provably stay on-trajectory without reliance on initial conditions. 
One of our key findings is the problem of over-reliance on autoregressive state in standard autoregressive models.
Our insights on this issue are supported by both theoretical analysis and experimental results.

The practical effectiveness of our approach was demonstrated through the application of FlowPAINT to 2D turbulent fluid dynamics, underscoring the potential of PAINT for complex, real-world systems. 
Taken together, this paper marks a significant advancement in the design of interpretable and widely applicable neural twins for intricate dynamical systems, paving the way for more accurate and adaptable modeling techniques.

\paragraph{Limitations and future work}
A major limitation of our method are the higher computational costs. 
Future work could explore more efficient architectures or alternative training paradigms, such as subsampling timesteps, to improve scalability.
Regarding the observed over-reliance on the autoregressive state, we provide theoretical insights and empirically validate behavior in our experiments.
However, these findings are specific to the considered data, task, and model and may not directly generalize to other settings.
Another limitation is the sliding window approach, which prevents truly continuous generations. 
This could potentially be mitigated by stitching techniques \citep{wei2019stitching}.

\FloatBarrier

\section*{Acknowledgements}

The ELLIS Unit Linz, the LIT AI Lab, the Institute for Machine Learning, are supported by the Federal State Upper Austria. We thank the projects FWF AIRI FG 9-N (10.55776/FG9), AI4GreenHeatingGrids (FFG- 899943), Stars4Waters (HORIZON-CL6-2021-CLIMATE-01-01), FWF Bilateral Artificial Intelligence (10.55776/COE12). 
We thank the European High Performance Computing initiative for providing computational resources (EHPC-DEV-2024D06-055, EHPC-DEV-2024D08-061, EHPC-DEV-2025D08-108).
We thank NXAI GmbH, Audi AG, Silicon Austria Labs (SAL), Merck Healthcare KGaA, GLS (Univ. Waterloo), T\"{U}V Holding GmbH, Software Competence Center Hagenberg GmbH, dSPACE GmbH, TRUMPF SE + Co. KG.


\bibliographystyle{styles/arxiv_2col/arxiv_2col}

\newpage
\appendix
\onecolumn

\newpage

\section{Theory}
\label{app:theory}

\subsection{Parallel-in-time models are on-trajectory}
\label{app:proof_paint_on_traj}
Here we provide a simple derivation to show that parallel-in-time models are on-trajectory under Assumption \ref{ass:measurement_decay}.
We will show that for an appropriate window size $h$, for all $t \in \mathbb{N}$ and any $\epsilon_1 > 0$:
\begin{align}
 ||p_\theta(\bm{x}_t \mid \bm{m}_{[t-h,t]}) - p(\bm{x}_t \mid \bm{m}_{[0,t]})|| < \epsilon_1.
\end{align}
First, from Assumption \ref{ass:measurement_decay} choose a window size $h$ such that:
\begin{align}
 ||p(\bm{x}_t \mid \bm{m}_{[t-h,t]}) - p(\bm{x}_t \mid \bm{m}_{[0,t]})|| < \epsilon_2 < \frac{\epsilon_1}{2}.
\end{align}
Then, from the Universal Function Approximator Theorem \citep{cybenko1989UFA}, choose an approximation of the true distribution as follows:
\begin{align}
||p_\theta(\bm{x}_t \mid \bm{m}_{[t-h,t]}) - p(\bm{x}_t \mid \bm{m}_{[t-h,t]})|| < \epsilon_3 < \frac{\epsilon_1}{2}.
\end{align}
Summing up the individual $\epsilon_1$ and $\epsilon_2$ the total approximation error is: $\epsilon_2 + \epsilon_3 < \frac{\epsilon_1}{2} + \frac{\epsilon_1}{2} < \epsilon_1$.
\\

\subsection{In general, autoregressive models are not on-trajectory}
\label{app:proof_AR_off_traj}
Here we show a more detailled counterexample to illustrate that autoregressive model of the form
\begin{align}
    p(\bm{x}_{[0,t]} \mid \bm{m}_{[0,t]}) = p(\bm{x}_0 \mid \bm{m}_{[0,t]}) \cdot \prod_{i=1}^t p_\theta(\bm{x}_i \mid \bm{x}_{i-1}, \bm{m}_{[0,t]}),
\end{align}
do not generally fulfill the on-trajectory property.
We construct this example while demonstrating the over-reliance on autoregressive state.
Being on-trajectory is defined as: for all $t \in \mathbb{N}$ and all $\epsilon > 0$:
\begin{align}
 ||p_\theta(\bm{x}_t \mid \bm{m}_{[0,t]}) - p(\bm{x}_t \mid \bm{m}_{[0,t]})|| < \epsilon.
\end{align}
From Section \ref{sec:theory} it is known that autoregressive models accumulate errors with the product of Jacobians, similar to classical ODEs \citep{orrell2005estimating_error_growth}.

\paragraph{Logistic map.}
Let $f: \mathbb{R} \mapsto \mathbb{R}$ be the logistic map, which is chaotic for $r = 3.8$ \citep{may1976logistic_map}:
\begin{align}
    f(x_{t-1}, \eta_t) &= r \cdot x_{t-1} \cdot (1 - x_{t-1})
\end{align}
Then we assume the model $f_\theta$ ignores the measurements and learns to predict $\bm{x}_t$ from $\bm{x}_{t-1}$ alone, but with a small error $\epsilon' > 0$.
\begin{align}
    f_\theta(x_{t-1},  m_{[0,t]}, \eta_t) &= r \cdot x_{t-1} \cdot (1 - x_{t-1}) + \epsilon'
\end{align}
This is an extreme case of over-reliance on autoregressive state, namely solely relying on it and not relying on the measurements at all.
For infinitesimal perturbations in \( x_{t-1} \) exponentiate over time, causing \( p_\theta(x_t \mid m_{[0,t]}) \) to diverge from the true trajectory distribution \( p(x_t \mid m_{[0,t]}) \) even for arbitrarily small \( \epsilon' \).
Thus, the autoregressive factorization is not on-trajectory because the model's sequential predictions amplify initial errors, violating the required closeness condition for all \( t > 0 \).

\section{Experimental details}
\label{app:experimental_details}

\subsection{Dataset}
\label{app:dataset}

\paragraph{Turbulent single jet dataset.}
We use an in-house dataset to ensure its quality and match the specific requirements of our study. As we have the capability to generate reliable high-fidelity data, additional external datasets were not needed. We consider the domain shown in Figure~\ref{fig:computational_domain}, which produces a two-dimensional turbulent free jet. The inlet is prescribed by a turbulent power-law velocity profile, with the system dynamics controlled by the mean inlet velocity. The Reynolds number is defined with respect to the inlet velocity as
\begin{equation}
    Re = \langle u \rangle_{\mathrm{inlet}} h_{\mathrm{inlet}} \nu^{-1}.
\end{equation}
The dataset is generated using the incompressible, pressure-based solver \texttt{pimpleFOAM} in \texttt{OpenFOAM~8} (\cite{opencfd2009open}), with subgrid scales modeled by the Smagorinsky LES model \citep{pope2001turbulent, frohlich2006large}. Simulations are performed on a structured mesh consisting of 19040 cells, and for training purposes, a subregion is extracted and interpolated onto a regular 128 × 128 grid to facilitate adaptability. The numerical simulations made use of a timestep size of $\Delta t = 6.5 \times 10^{-4}\text{s}$. For training purposes, the data are sampled every 70th timestep, resulting in an effective temporal resolution of $4.55 \times 10^{-2}\text{s}$. For each Reynolds number, the dataset contains 1000 such snapshots.

\usetikzlibrary{arrows.meta, positioning}

\begin{figure}[h]
    \centering
        \resizebox{0.8\columnwidth}{!}{
        \begin{tikzpicture}[every node/.style={font=\sffamily\Large}, >=Stealth]
            
            \draw[line width=1.5pt] (0,0) -- (0,1.8);          
            \draw[line width=1.5pt] (0,1.8) -- (-1.5,1.8);     
            \draw[line width=0.5pt] (-1.5,1.8) -- (-1.5,2.2);  
            \draw[line width=1.5pt] (-1.5,2.2) -- (0,2.2);     
            \draw[line width=1.5pt] (0,2.2) -- (0,4);          
            \draw[line width=0.5pt] (0,4) -- (4,4);            
            \draw[line width=0.5pt] (4,4) -- (4,0);            
            \draw[line width=0.5pt] (4,0) -- (0,0);

            \node[draw, dashed, line width=0.5pt, fill=gray!30, minimum width=3cm, minimum height=3cm] at (1.55,2) (inner) {};
            
            \draw[->, thick] (-1.2,2) -- (-0.2,2);  
            
            \node[right, align=center] at (-3.7,2.8) 
                    (inlet) {$u = u_\mathrm{inlet}$};
            \draw[line width=0.25pt] (-2.1,2.6) -- (-1.5,2.1);             
            
            \node[right, align=center] at (-2.6,3.6) 
                (wall) {$u = 0$};
            \draw[line width=0.25pt] (-1.6,3.4) -- (0,2.7);             
            
            \node[right, align=center] at (5,2.5)
                (outlet) {$\begin{aligned}
                    \frac{\partial u}{\partial n} &= 0 \;\mathrm{for}\; u \cdot n \geq 0, \\
                    u &= 0 \;\mathrm{for}\; u \cdot n < 0.
                \end{aligned}$};
            \draw[line width=0.25pt] (4,2.0) -- (5.0,2.3);             

            \draw[<->] (-1.8,1.8) -- (-1.8,2.2) node[midway,left] {$h_\mathrm{inlet}$};
            \draw[line width=0.1pt] (-1.8,1.8) -- (-1.5,1.8);             
            \draw[line width=0.1pt] (-1.8,2.2) -- (-1.5,2.2);

        \end{tikzpicture}
        }
    \caption{
    Computational domain of the CFD simulation, featuring a thin channel with a prescribed inlet velocity profile, transitioning into an expanded outflow region zone. The channel and outflow orifice walls enforce a no-slip boundary condition, while the top, bottom and right boundaries of the outflow region implement a conditional Neumann/Dirichlet outflow condition. The light-grey subregion denotes the part of the CFD domain considered for model training. The arrow shows the imposed inlet flow direction.
    }
    \label{fig:computational_domain}
\end{figure}

\paragraph{A note on 2D turbulence.}
In fluid mechanics, the behavior of turbulent flows differs fundamentally between three and two dimensions. In three dimensions, turbulence is characterized by the \emph{energy cascade}: the largest eddies form due to inertial instabilities but tend to exist only briefly, breaking up into progressively smaller eddies. This cascade continues until viscous forces dissipate the smallest scales.
\\
In contrast, in two dimensions, the energy cascade is reversed, transferring energy from smaller to larger eddies, which produces coherent, long-lived vortices. Intuitively, this can be explained by \emph{vortex stretching}. Consider a thin tube of vorticity: if a velocity gradient along the tube axis is present, the tube stretches, and due to conservation of angular momentum, the vorticity magnitude increases. This mechanism is absent in two dimensions because there is no third dimension along which stretching can occur. As a result, vortices cannot break up and tend to merge into larger structures if they have the same rotational sense, which explains the inverse energy cascade.
\\
Mathematically, this is seen in the vorticity transport equation derived from the curl of the Navier–Stokes equations. In 2D, the velocity and vorticity vectors are perpendicular, which makes the vortex stretching term identically zero.
\\
In reality, there is no perfectly two-dimensional turbulence; even quasi-2D flows, such as atmospheric flows, exhibit three-dimensional structures at smaller scales. Nevertheless, we are confident that the proposed method, having demonstrated robust performance in two-dimensional turbulence, will generalize effectively to three-dimensional turbulent flows. For a more detailed discussion of two-dimensional turbulence, see \cite{davidson2015turbulence}.

\paragraph{Train/val/test splits.}
We consider 18 different training trajectories with varying Reynolds numbers, as explained above. The Reynolds numbers used range from 700 to 2400 in steps of 100.
The split is performed at random, but avoiding that any validation or test data is placed in the extrapolation regime.
The exact split can be seen in Table \ref{tab:reynolds_split}.

\begin{table}[h]
\centering
\caption{Reynolds numbers by split.}
\label{tab:reynolds_split}
\begin{tabular}{l||ccc}
\textbf{Reynolds} & train & val & test \\
\hline
700  & X & & \\
800  & X & & \\
900  & X & & \\
1000 & X & & \\
1100 & & & X \\
1200 & X & & \\
1300 & X & & \\
1400 & X & & \\
1500 & X & & \\
1600 & & X & \\
1700 & X & & \\
1800 & X & & \\
1900 & X & & \\
2000 & X & & \\
2100 & & & X \\
2200 & X & & \\
2300 & X & & \\
2400 & X & & \\
\end{tabular}
\end{table}

\subsection{Implementation and architecture}
\label{app:architecture}

\paragraph{Architecture and hyperparameters.}
FlowPAINT and the autoregressive UNet were both implemented in Pytorch \citep{paszke2017pytorch}.
For FlowPAINT we trained in float16 mixed precision.
The most important training parameters are provided in Table \ref{tab:exp_details}.
Our code also uses FlashAttention \citep{dao2022flashattention,dao2023flashattention2}. 
For both models we use 20 denoising steps for generating samples.

\paragraph{Autoregressive UNet.}
For the UNet we followed \citep{kohl2023benchmarkingARDiff} and took the implementation of the public Github repository \footnote{\url{https://github.com/tum-pbs/autoreg-pde-diffusion}}.
We matched the number of parameters by increasing the default model dimension from 128 to 224.
For probe-conditioning, we followed the same recipe as for prior timestep conditioning and added mask and probe values as additional channels. We trained two UNet variants, one conditioned on probe measurements from the current timestep and another conditioned on a 16-timestep probe history, to ensure a fair comparison with FlowPAINT as well as demonstrate that even an extensive probe history cannot overcome the dominance of the autoregressive state in a given prediction.

\paragraph{Autoregressive Diffusion Transformer.}
We tested the Diffusion Transformer \citep{peebles2023DiT} architecture used for PAINT as an autoregressive model.
However, we observe that its predictive performance degrades rapidly after only a few rollout steps (see Figure \ref{fig:comp_baseline}). 
Consequently, we rely on the above-mentioned well-established UNet architecture mentioned above for the primary comparison to ensure a fair and robust baseline.

\paragraph{Optimization hyperparameters.}
We employed well-established techniques and hyperparameters for optimization, including AdamW \citep{loshchilov2017adamw}, cosine annealing of the learning rate, etc. 
These choices were grounded in their proven effectiveness and widespread adoption in the literature. 
Although we acknowledge that tuning these hyperparameters could potentially yield marginal performance improvements, we do not anticipate that such modifications would substantially impact the primary conclusions and insights drawn from our experiments. 
Given the computational constraints and the dimensionality of the hyperparameter search space, we chose to use the established settings for the sake of experimental tractability. Future work may explore the potential benefits of hyperparameter tuning in this context to validate this assumption.

\paragraph{Selection of batch size and timesteps per batch.}
As mentioned in Section~\ref{sec:experiments}, a direct comparison between PAINT and the autoregressive baseline is not entirely feasible, given PAINT's prediction of a full sequence of states.
Therefore, we carefully evaluated and chose the \emph{batch size} and the \emph{timesteps per batch} according to the following procedure
\\
For the autoregressive UNet, we opted for the maximum batch size that could be accommodated by a single GPU, which amounted to 144. 
We also tested a batch size of 288 on 2 GPUs via data-parallel training. 
However, as there was no significant change in training nor validation loss, we settled on a batch size of 144 for all experiments.
\\
In the case of FlowPAINT 16/8, we maintained the same batch size of 144. 
For FlowPAINT 16/0, we tested two variants: one with a batch size of 144, and another with an increased batch size of $144 \times 1.5 = 216$. The latter was chosen to align the number of predicted timesteps with the 16/8 setting.
Given the negligible difference observed between the two variants, we proceeded with a batch size of 144 for the remainder of the FlowPAINT experiments.

\begin{table*}[h!]
\centering
\caption{Experimental details for FlowPaint and the autoregressive baseline.}
\label{tab:exp_details}
\begin{tabular}{llcc}
\hline
 &  & FlowPaint & AR baseline \\
\hline
Data &  &  &  \\
 & Autoregressive history length & 0 & 2 \\
 & Probe history length & 16 & [16, 1] \\
 & Forward prediction & [8, 0] & 1 \\
\hline
Optimization &  &  &  \\
 & Optimizer & AdamW & AdamW \\
 & AdamW decay & {0.05} & {0.05} \\
 & Learning rate & $10^{-4}$ & $10^{-4}$ \\
 & Learning rate start & $5 \cdot 10^{-7}$ & $5 \cdot 10^{-7}$ \\
 & Learning rate warmup steps & 10000 & 10000 \\
 & Learning rate end & $10^{-5}$ & $10^{-5}$ \\
 & Train steps & 100K & 100K \\
 & Batch size & [144, 216] & 144 \\
\end{tabular}
\end{table*}

\paragraph{Compute requirements.}
FlowPaint is expensive during training, as it needs to reconstruct a video in parallel.
We had limited access to a cluster environment and trained FlowPaint on 3 nodes each consisting of 4 NVIDIA Ampere A100 GPUs with 64GB VRAM.
Training took around 30 hours for 100K iterations.
\\
The autoregressive baseline is comparably cheap. 
It was trained on either a single Nvidia RTX-PRO6000 Blackwell 96GB or a single Nvidia H100-SXM-80GB for ca. 24 hours. 
The inference times are reported in the main paper.

\begin{table*}[h]
\centering
\caption{Architectural details of FlowPaint.}
\label{tab:arch_details}
\begin{tabular}{llc}
\hline
Encoding &  &   \\
 & num\_layers of 3x3x3 conv & {3}  \\
 & patch size in pixels & 4x4  \\
\hline
Transformer &  &    \\
 & model\_dim & {192}\\
 & layers & {10}  \\
 & num\_heads & {3} \\
 & Spatial Block & Yes \\
 & Temporal Block & Yes \\
 & Temporal Convolution & No \\
\hline
Decoding &  &  \\
 & num\_layers of 3x3x3 conv & {3} 
\end{tabular}
\end{table*}

\subsection{Additional results and ablations}
\label{app:results}

\begin{figure*}[h]
    \centering
    
    \begin{subfigure}{0.9\linewidth}
        \centering
        \includegraphics[width=\linewidth]{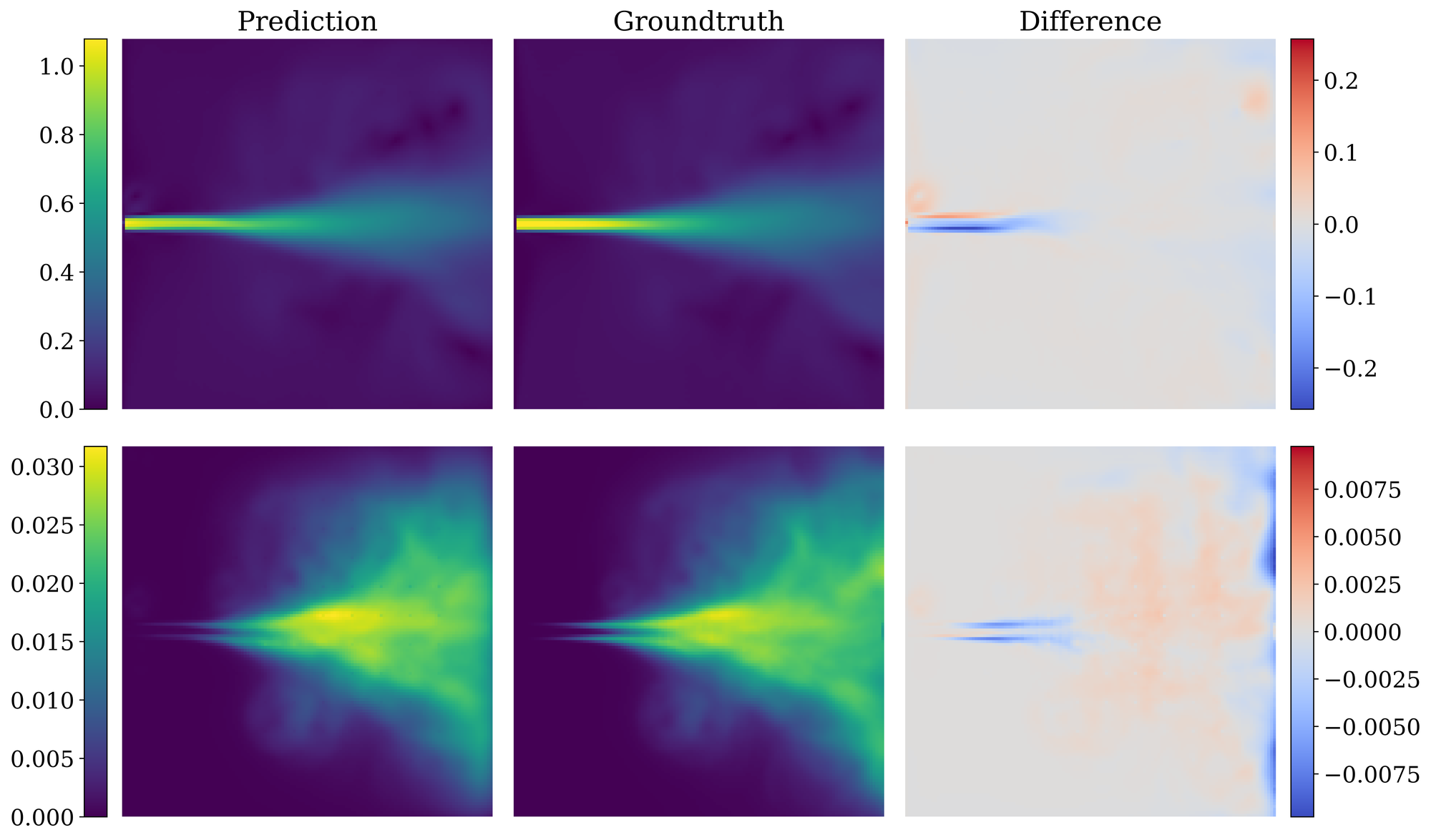}
        \caption{Mean velocity ($\overline{u}$) and variance ($\overline{u'^2}$) of the reconstructed trajectories compared with the ground truth. All values are normalized by the mean inlet velocity.}
        \label{fig:Re2100grid_SF_physicsA}
    \end{subfigure}
    
    \vspace{0.5cm}
    
    \begin{subfigure}{0.48\linewidth}
        \centering
        \includegraphics[width=\linewidth]{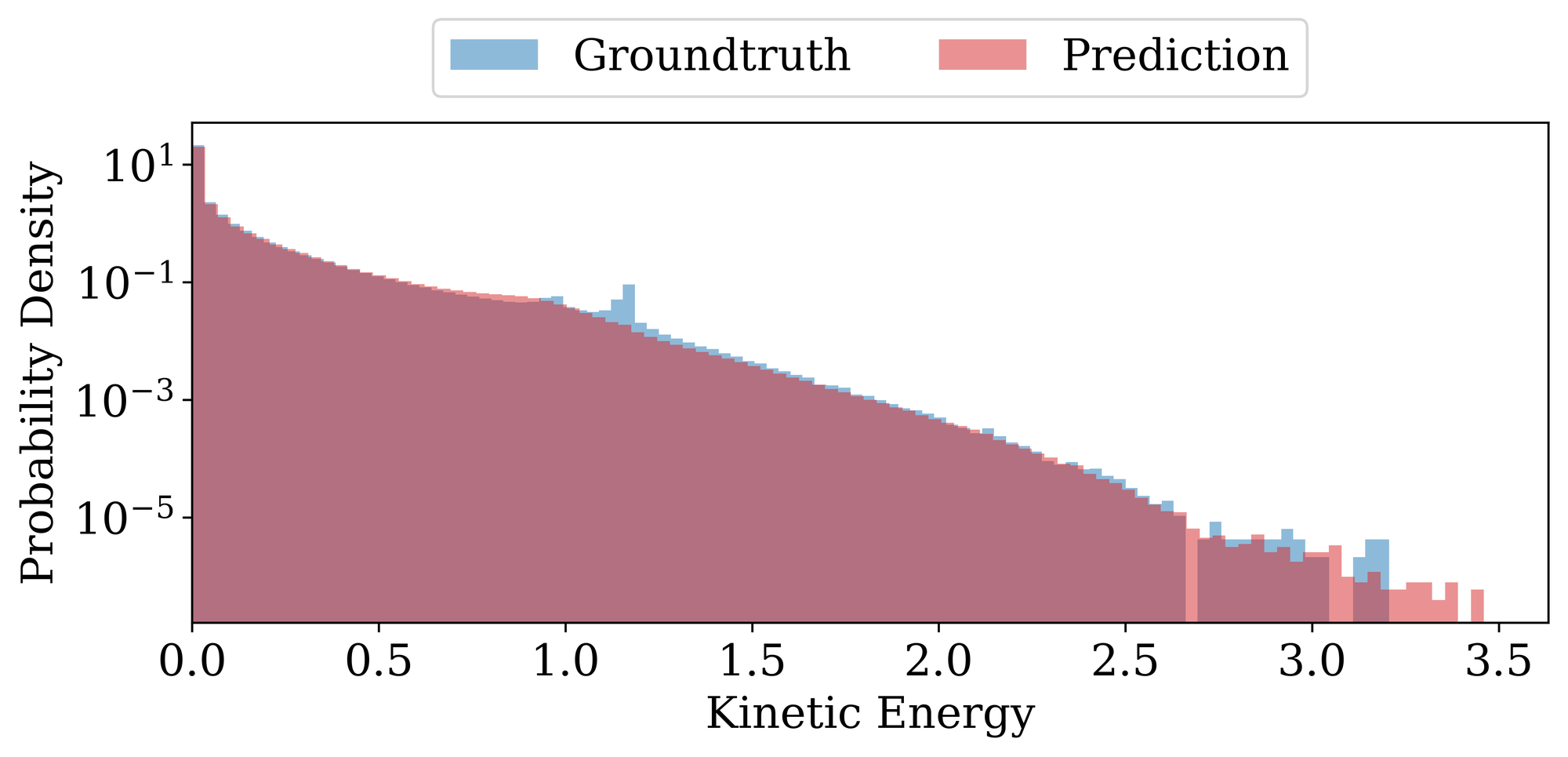}
        \caption{Histogram of time-averaged kinetic energy.}
        \label{fig:Re2100grid_SF_physicsB}
    \end{subfigure}
    \hfill
    \begin{subfigure}{0.48\linewidth}
        \centering
        \includegraphics[width=\linewidth]{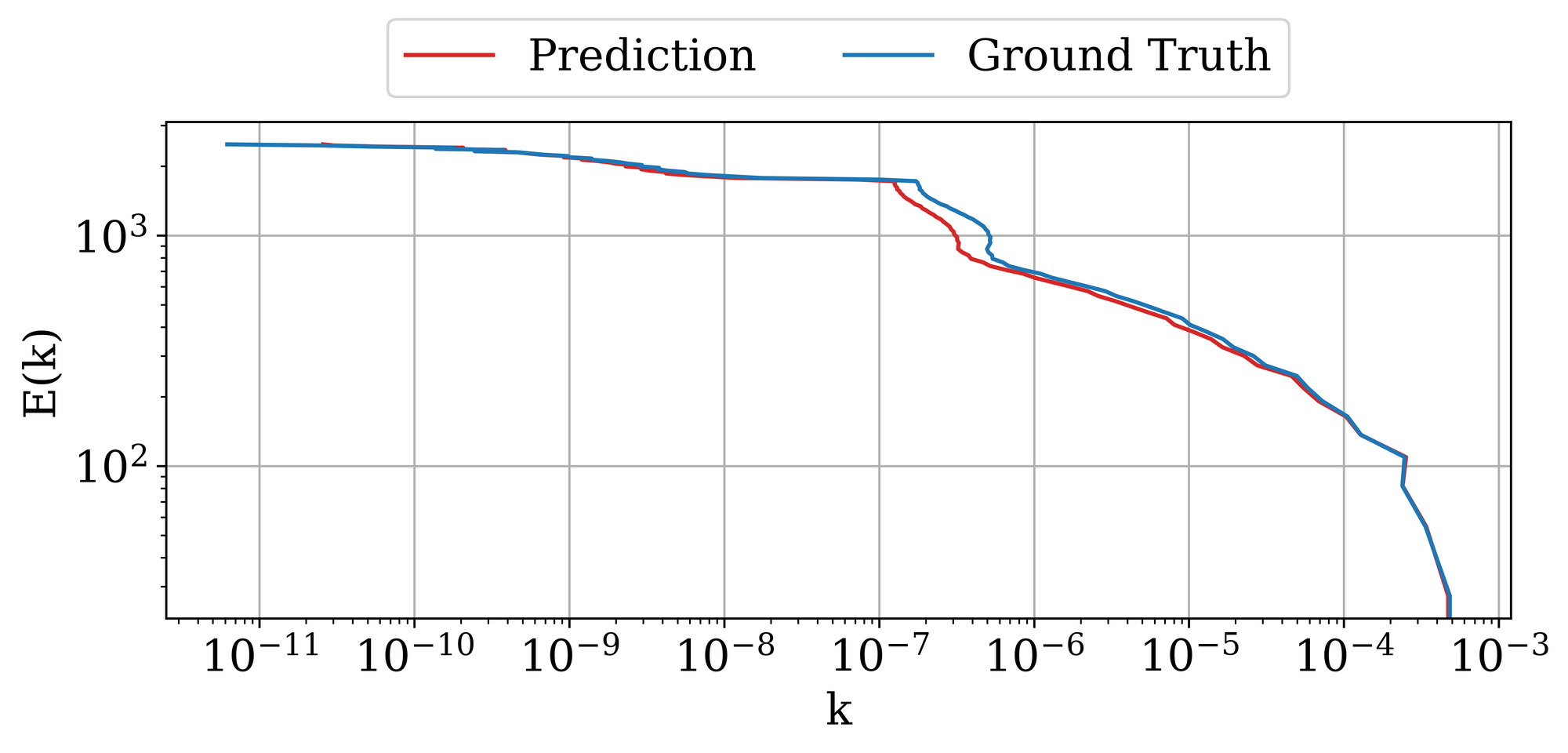}
        \caption{Time-averaged kinetic energy spectrum.}
        \label{fig:Re2100grid_SF_physicsC}
    \end{subfigure}

    \caption{Statistical analysis of reconstructed flow fields using FlowPAINT 16/0 for Reynolds number 2100 and a \emph{grid} probe point constellation.}
    \label{fig:Re2100grid_SF_physics}
\end{figure*}

\begin{figure*}[h]
    \centering
    
    \begin{subfigure}{0.9\linewidth}
        \centering
        \includegraphics[width=\linewidth]{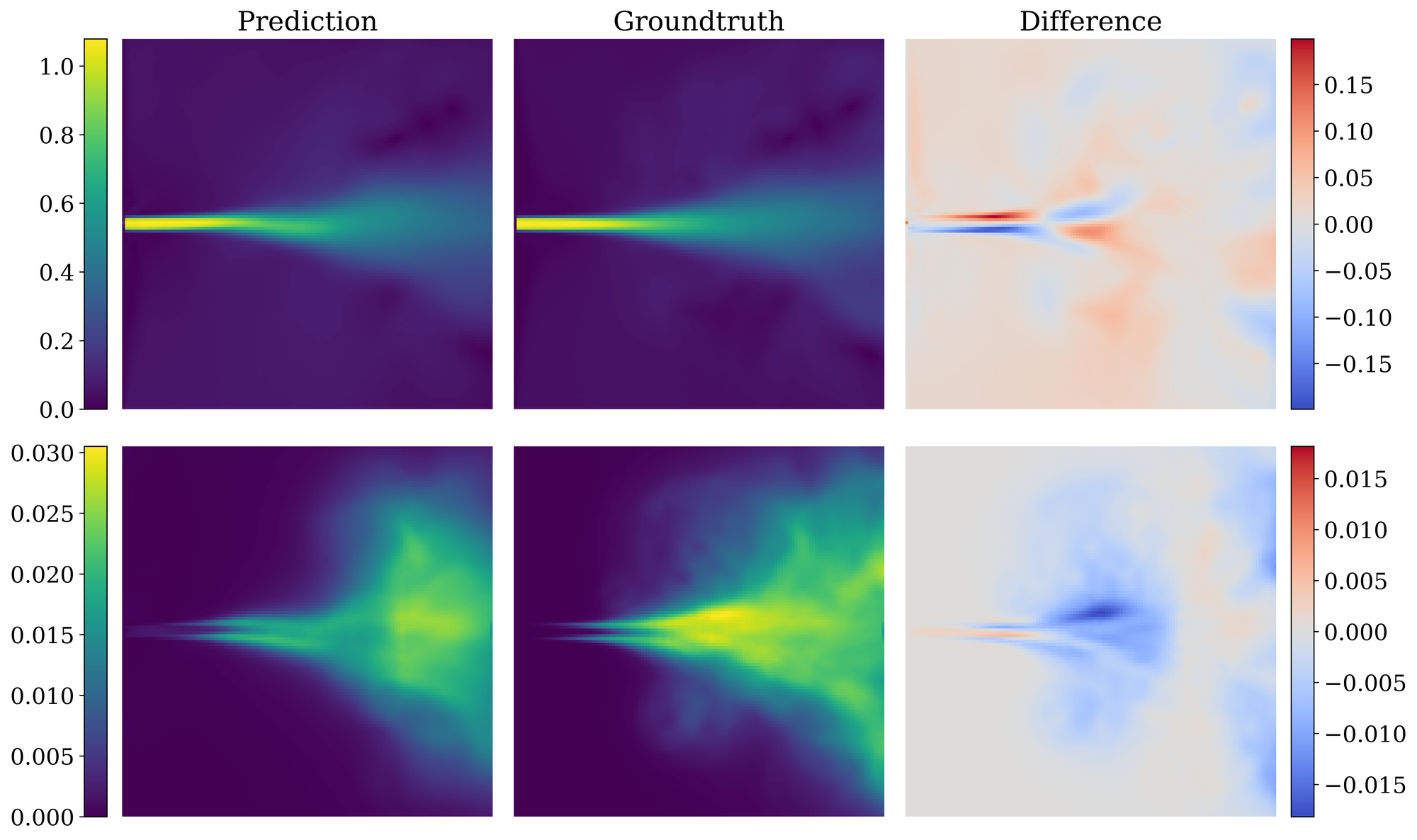}
        \caption{Mean velocity ($\overline{u}$) and variance ($\overline{u'^2}$) of the reconstructed trajectories compared with the ground truth. All values are normalized by the mean inlet velocity.}
        \label{fig:Re2100vertical_SF_physicsA}
    \end{subfigure}
    
    \vspace{0.5cm}
    
    \begin{subfigure}{0.48\linewidth}
        \centering
        \includegraphics[width=\linewidth]{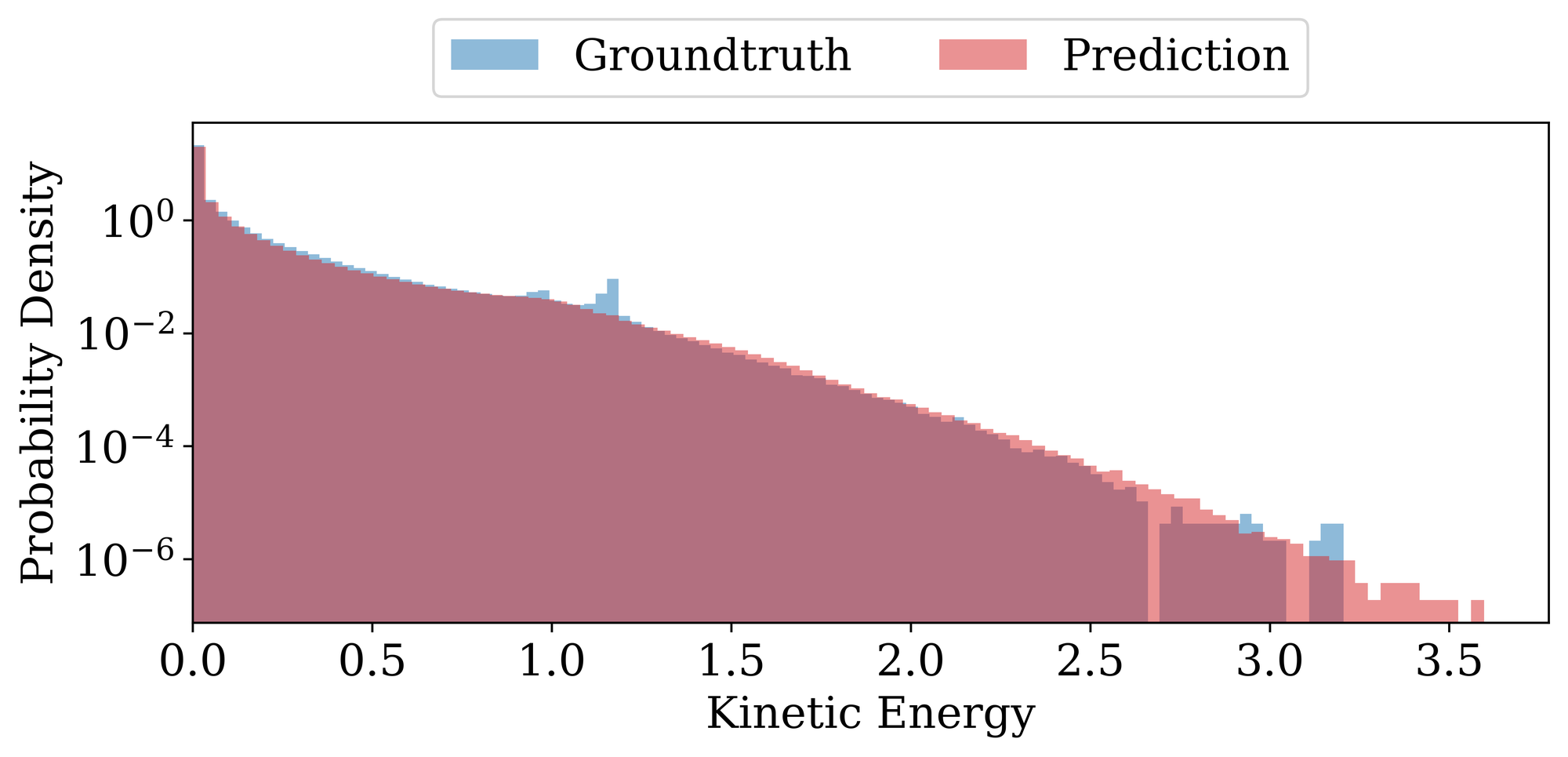}
        \caption{Histogram of time-averaged kinetic energy.}
        \label{fig:Re2100vertical_SF_physicsB}
    \end{subfigure}
    \hfill
    \begin{subfigure}{0.48\linewidth}
        \centering
        \includegraphics[width=\linewidth]{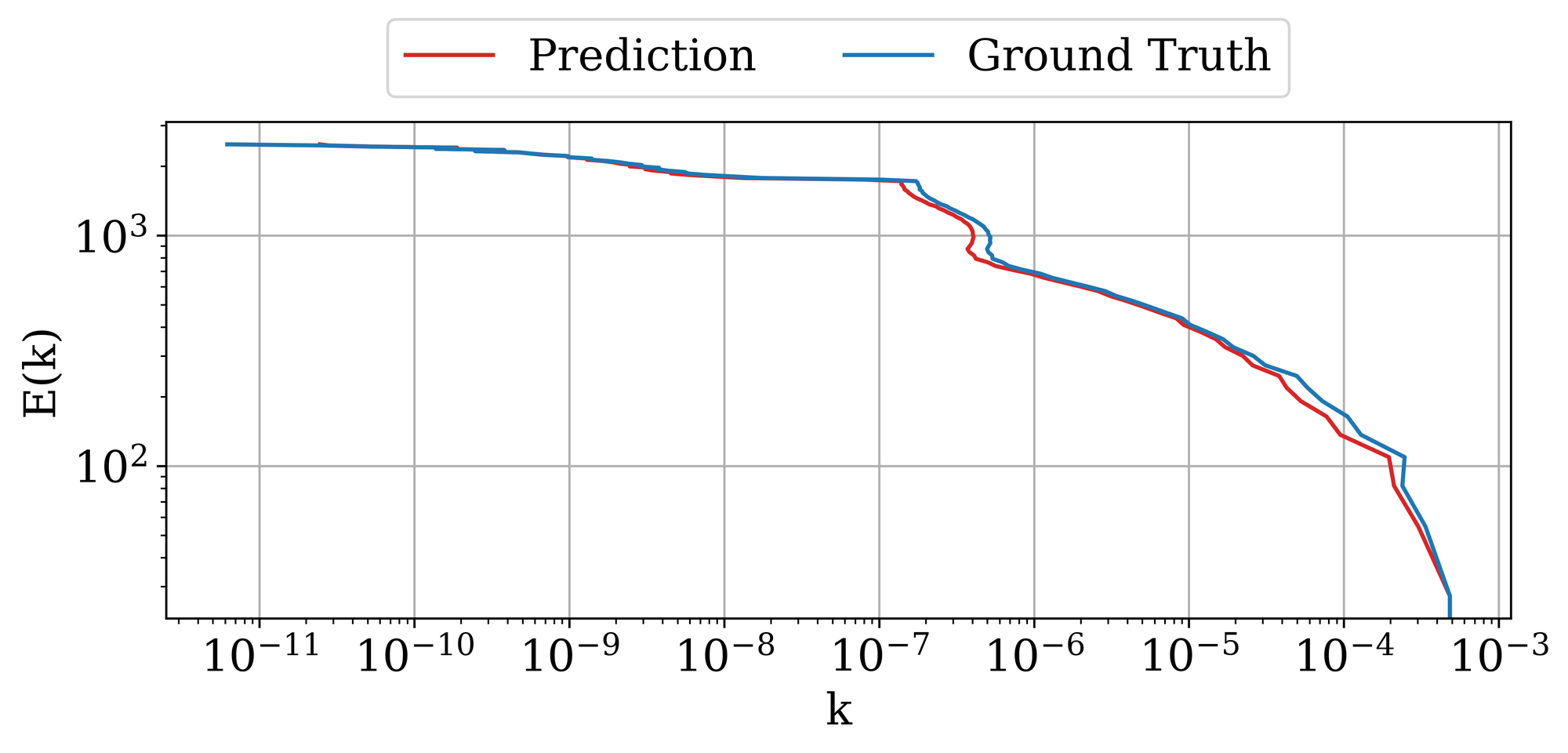}
        \caption{Time-averaged kinetic energy spectrum.}
        \label{fig:Re2100vertical_SF_physicsC}
    \end{subfigure}

    \caption{Statistical analysis of reconstructed flow fields using FlowPAINT 16/0 for Reynolds number 2100 and a \emph{vertical} probe point constellation.}
    \label{fig:Re2100vertical_SF_physics}
\end{figure*}

\begin{figure*}[h] 
    \centering
    
    \begin{subfigure}{0.9\linewidth}
        \centering
        \includegraphics[width=\linewidth]{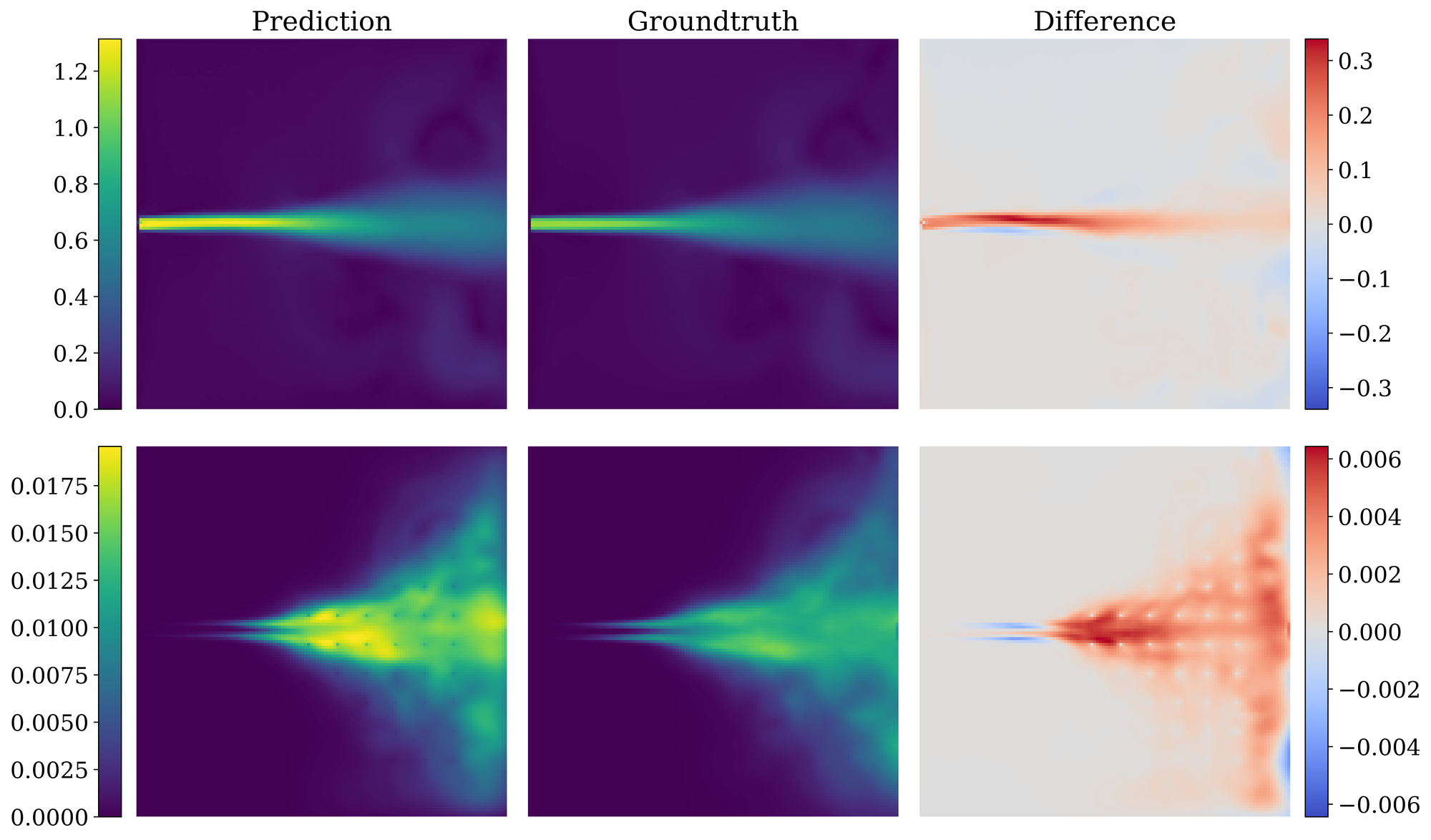}
        \caption{Mean velocity ($\overline{u}$) and variance ($\overline{u'^2}$) of the reconstructed trajectories compared with the ground truth. All values are normalized by the mean inlet velocity.}
        \label{fig:Re1100grid_SF_physicsA}
    \end{subfigure}
    
    \vspace{0.5cm}
    
    \begin{subfigure}{0.48\linewidth}
        \centering
        \includegraphics[width=\linewidth]{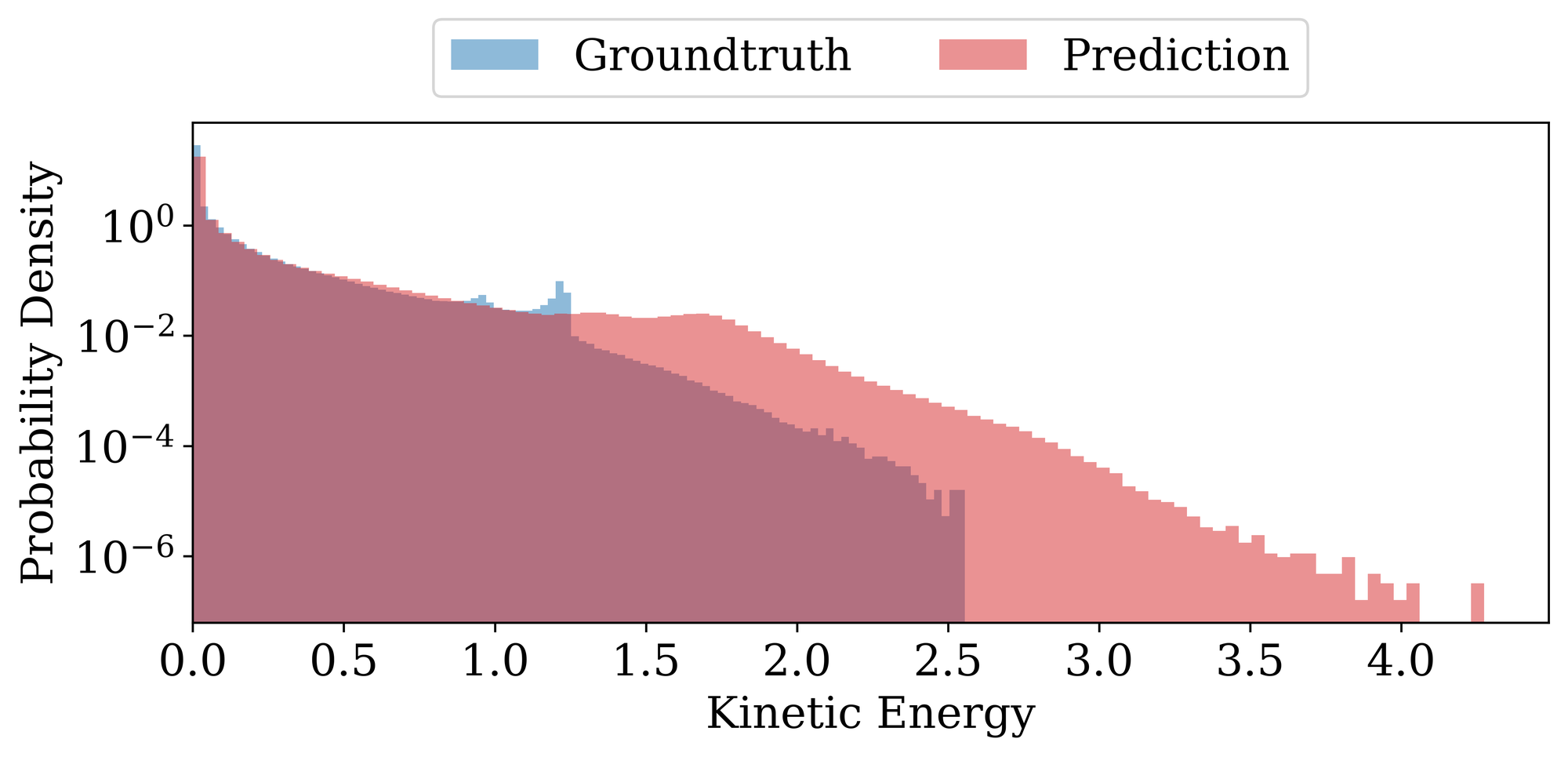}
        \caption{Histogram of time-averaged kinetic energy.}
        \label{fig:Re1100grid_SF_physicsB}
    \end{subfigure}
    \hfill
    \begin{subfigure}{0.48\linewidth}
        \centering
        \includegraphics[width=\linewidth]{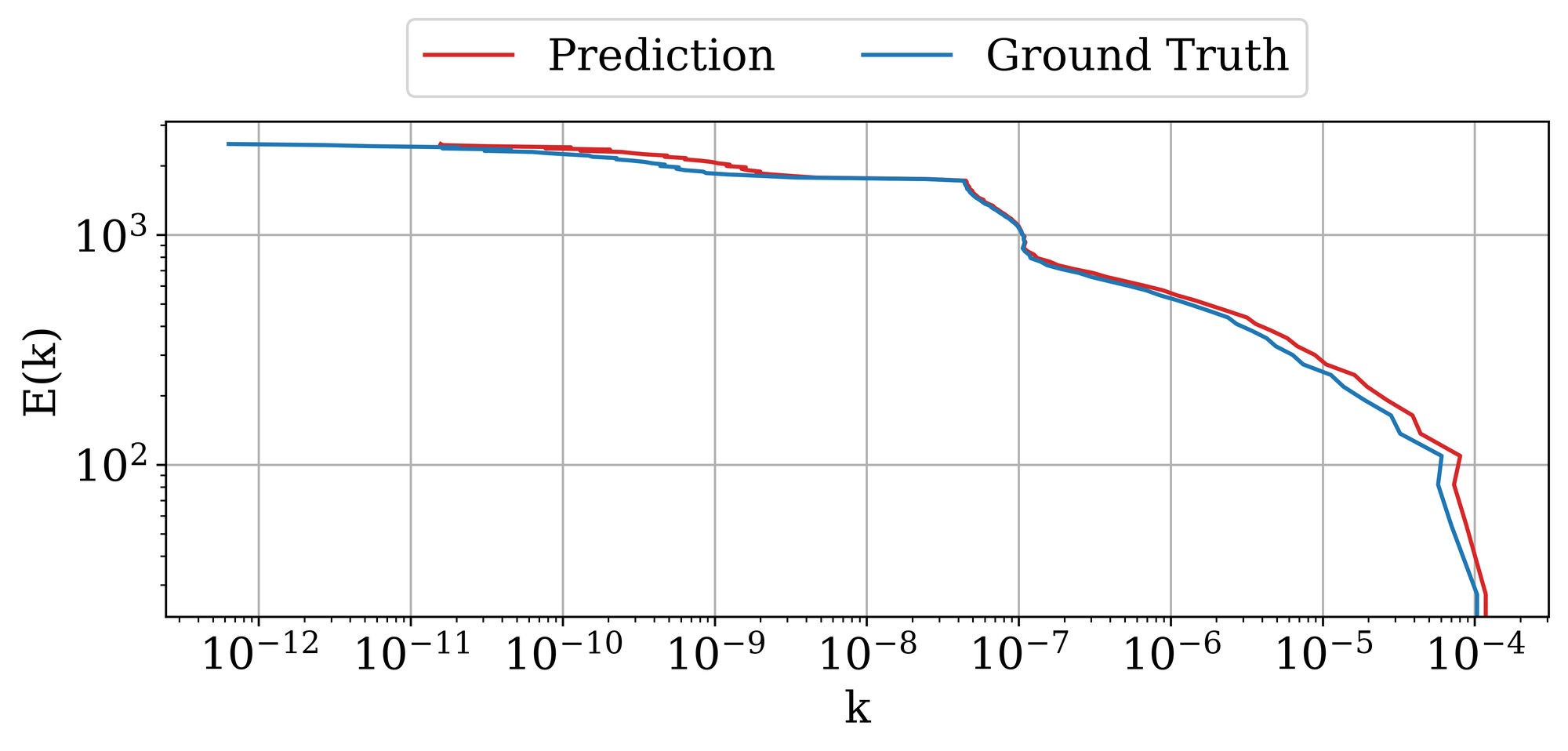}
        \caption{Time-averaged kinetic energy spectrum.}
        \label{fig:Re1100grid_SF_physicsC}
    \end{subfigure}

    \caption{Statistical analysis of reconstructed flow fields using FlowPAINT 16/0 for Reynolds number 1100 and a \emph{grid} probe point constellation.}
    \label{fig:Re1100grid_SF_physics}
\end{figure*}

\begin{figure*}[h] 
    \centering
    
    \begin{subfigure}{0.9\linewidth}
        \centering
        \includegraphics[width=\linewidth]{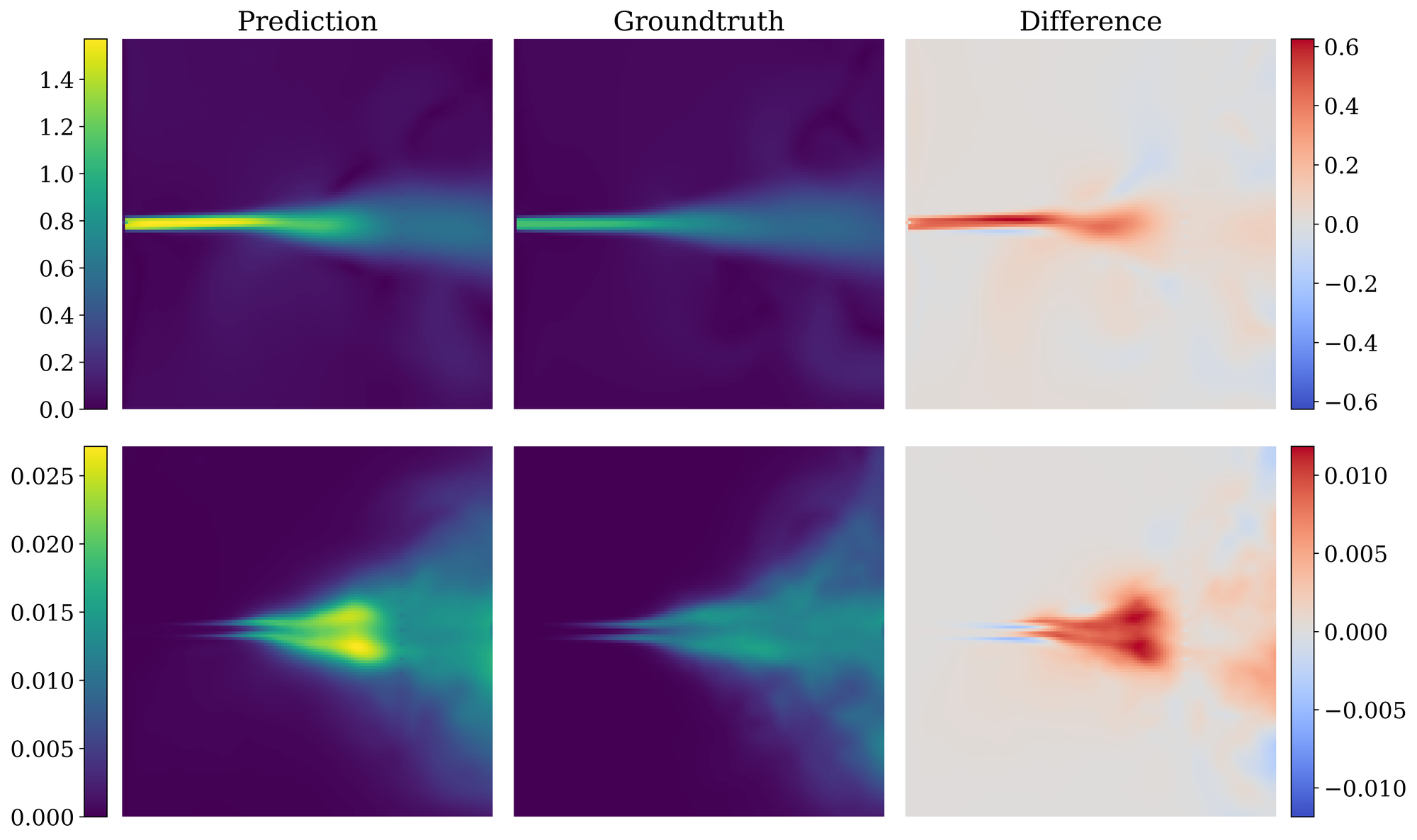}
        \caption{Mean velocity ($\overline{u}$) and variance ($\overline{u'^2}$) of the reconstructed trajectories compared with the ground truth. All values are normalized by the mean inlet velocity.}
        \label{fig:Re1100vertical_SF_physicsA}
    \end{subfigure}
    
    \vspace{0.5cm}
    
    \begin{subfigure}{0.48\linewidth}
        \centering
        \includegraphics[width=\linewidth]{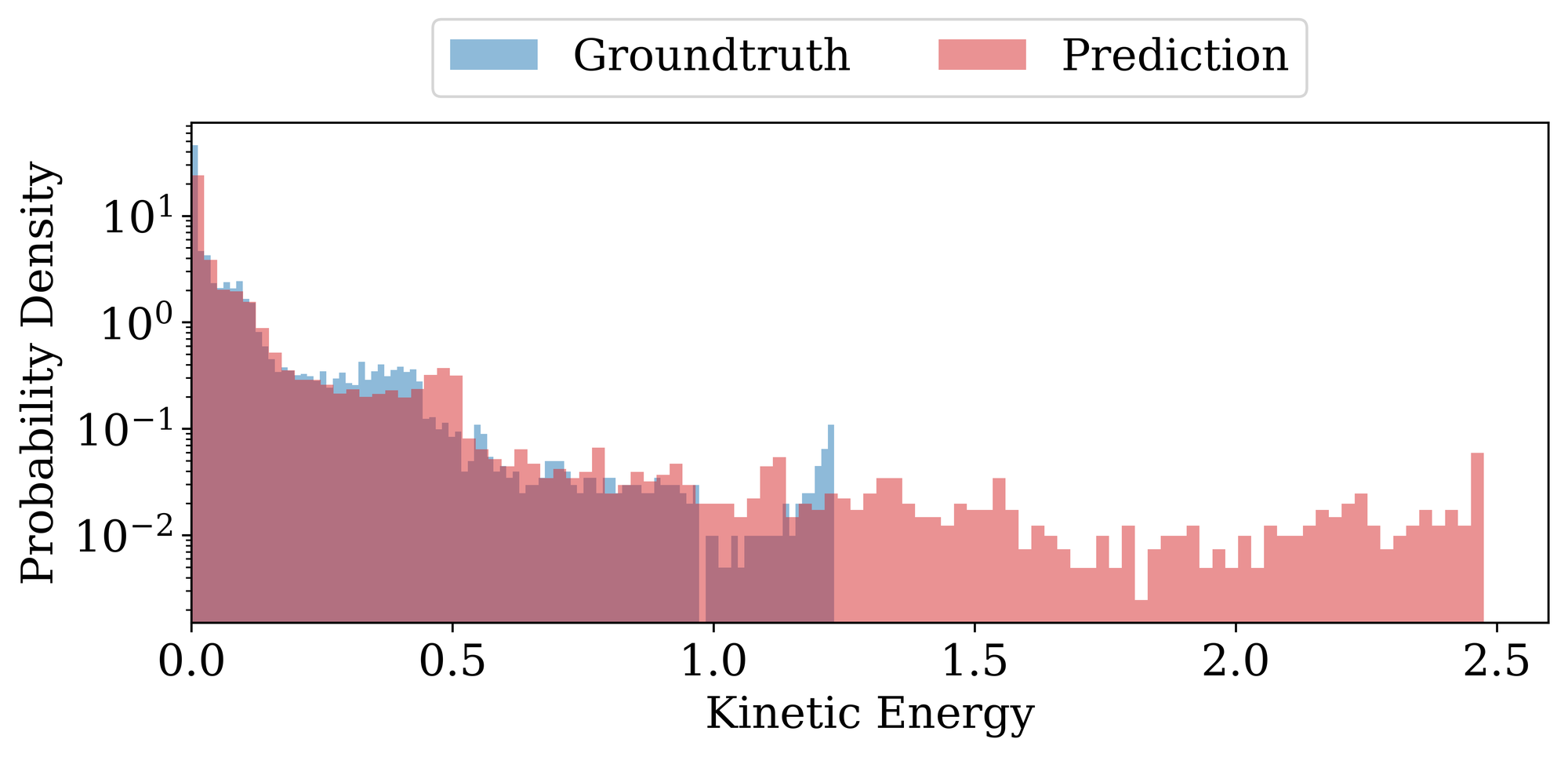}
        \caption{Histogram of time-averaged kinetic energy.}
        \label{fig:Re1100vertical_SF_physicsB}
    \end{subfigure}
    \hfill
    \begin{subfigure}{0.48\linewidth}
        \centering
        \includegraphics[width=\linewidth]{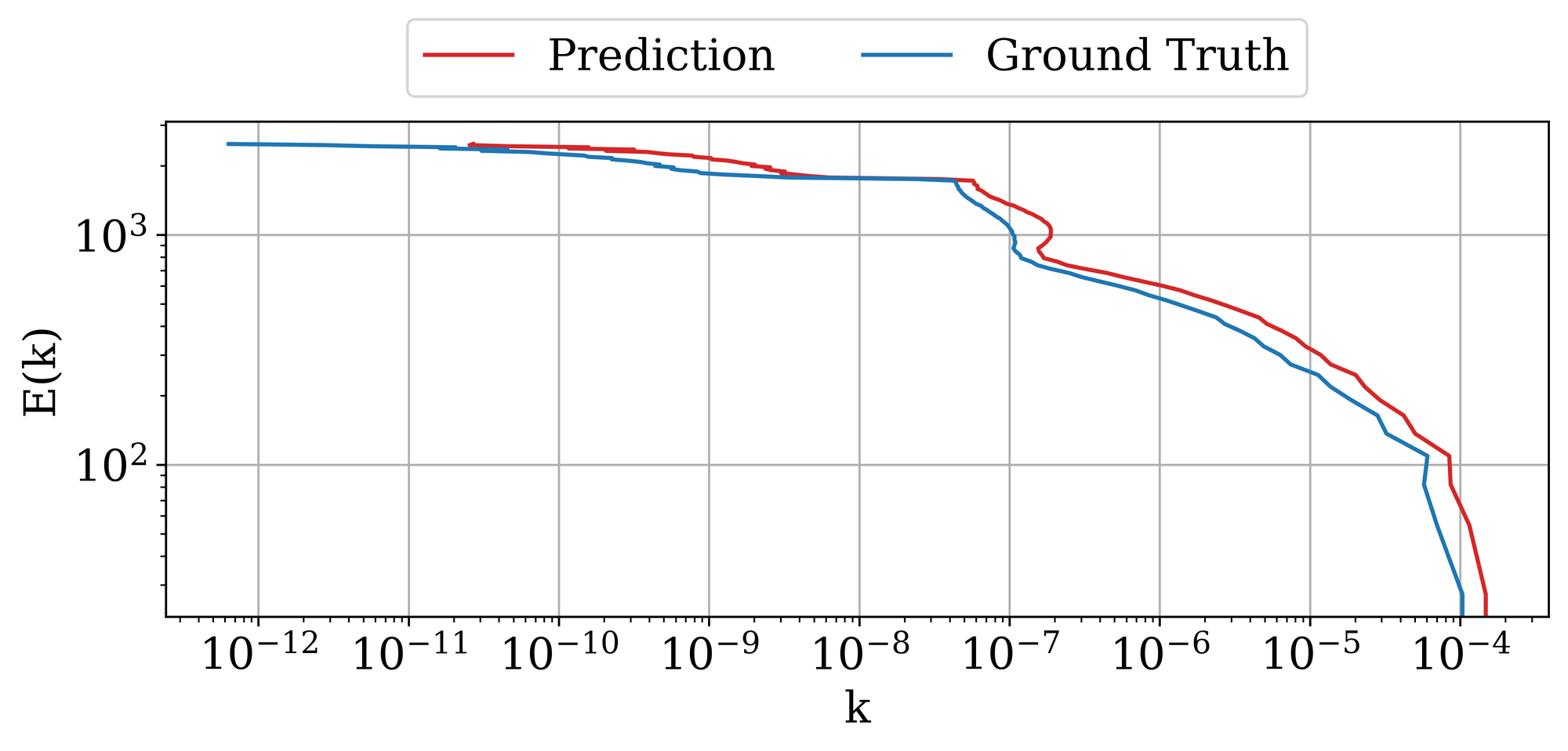}
        \caption{Time-averaged kinetic energy spectrum.}
        \label{fig:Re1100vertical_SF_physicsC}
    \end{subfigure}

    \caption{Statistical analysis of reconstructed flow fields using FlowPAINT 16/0 for Reynolds number 1100 and a \emph{vertical} probe point constellation.}
    \label{fig:Re1100vertical_SF_physics}
\end{figure*}

\begin{figure*}[h!]
    \centering
    \includegraphics[width=0.9\linewidth]{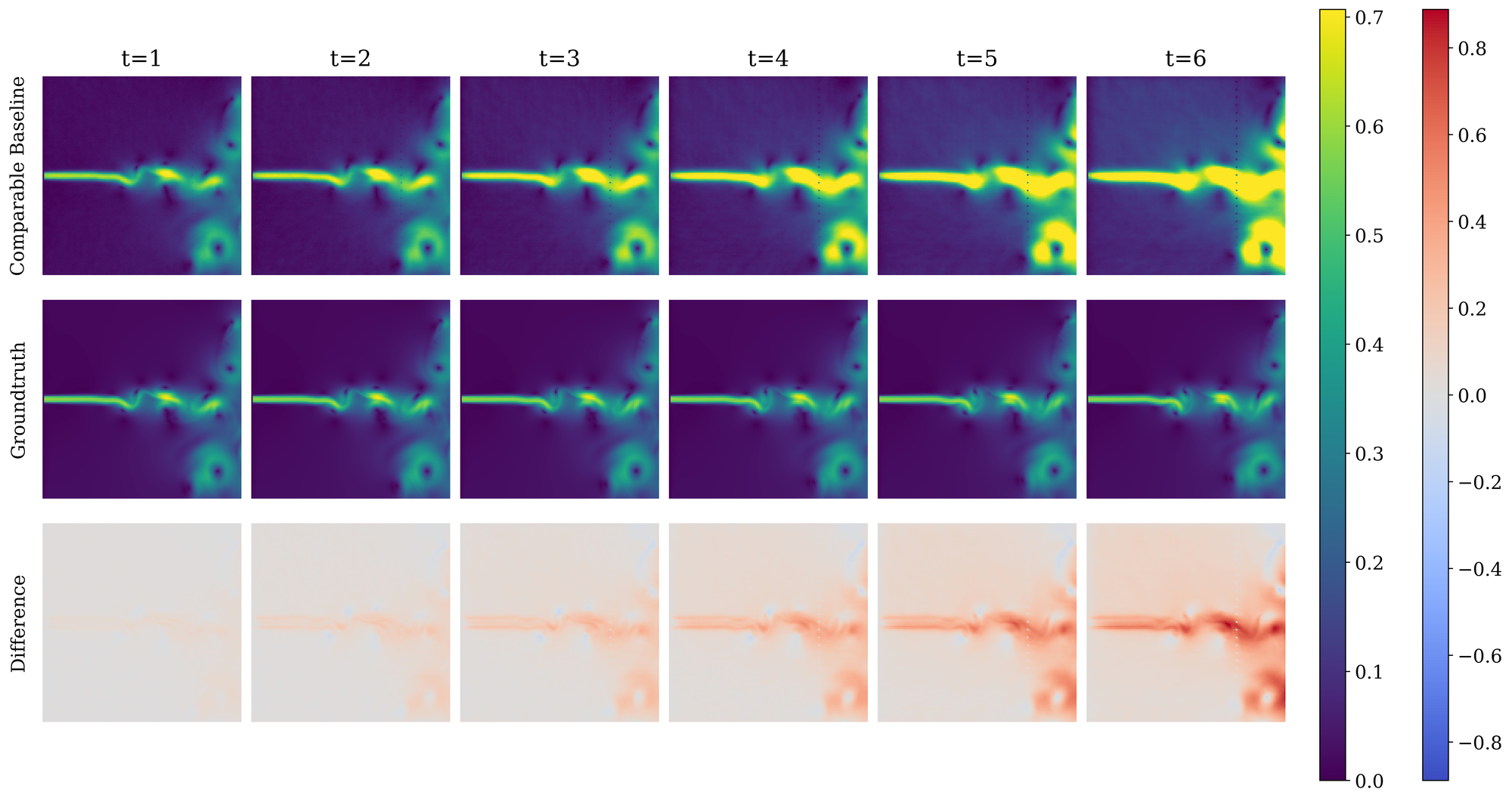}
    \caption{Autoregressive rollout of a baseline model with an architecture comparable to our PAINT implementation. The prediction quality deteriorates almost immediately, motivating the use of a well-established autoregressive baseline to enable a fair comparison between traditional autoregressive methods and our parallel-in-time approach.}
    \label{fig:comp_baseline}
\end{figure*}

We perform ablations to gain further insights on the importance of certain components.
Ablation results are summarized in Table~\ref{tab:results_appendix}, with further evaluations shown in Figures \ref{fig:Re1100grid_SF_physics}, \ref{fig:Re1100vertical_SF_physics}, \ref{fig:Re2100grid_SF_physics}, \ref{fig:Re2100vertical_SF_physics} and \ref{fig:comp_baseline}.

\paragraph{Stability through future predictions.}
Comparing the smoothing-and-filtering variant, FlowPAINT 16/0 with the smoothing-filtering-and-prediction variant, FlowPAINT 16/8 highlights the stabilizing effect of incorporating future-state predictions. While the FlowPAINT 16/0 approach relies exclusively on reconstructing the given measurement series, the inclusion of predictive task seems to enable FlowPAINT 16/8 to produce more stable reconstructions. FlowPAINT 16/0 performs in general worse than FlowPAINT 16/8, though still better than the autoregressive baseline.

\begin{figure}[h!]
    \centering
    \includegraphics[width=0.5\linewidth]{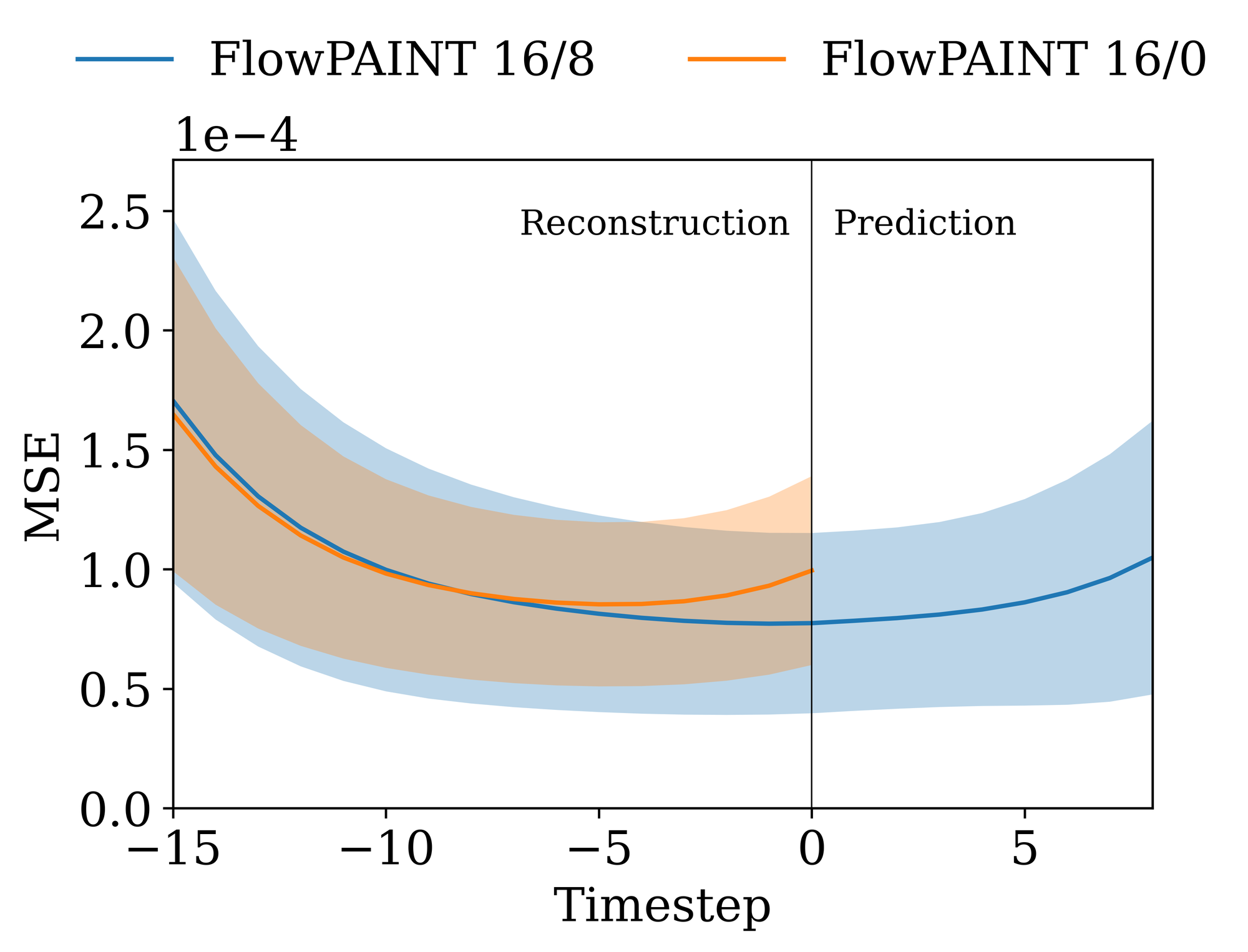}
    \caption{The solid lines indicate the mean MSE over a reconstructed sequence, with the shaded regions indicating ± one standard deviation across samples. While the average MSE for future timesteps remains relatively low, FlowPAINT surprisingly exhibits the highest error for timesteps in the past. This data stems from an evaluation using Reynolds number 2100 and the \textit{grid} probe point constellation.}
    \label{fig:MSE_future}
\end{figure}

\begin{figure}[h!]
    \centering
    \includegraphics[width=0.5\linewidth]{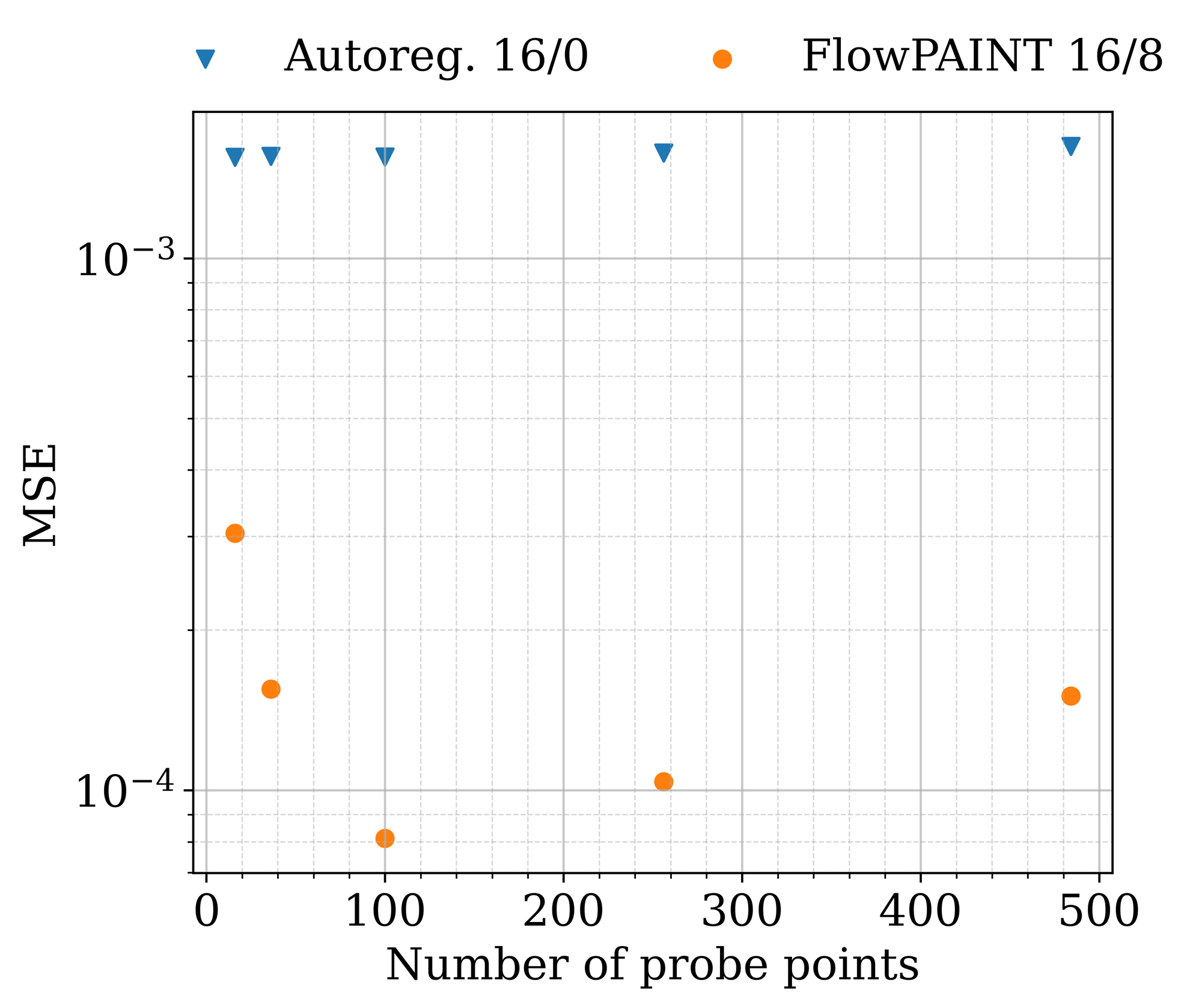}
    \caption{
    MSE over number of probe points provided during inference. Probe points are arranged in a pattern comparable to the \textit{grid} probe constellation. For FlowPAINT 16/8, the MSE decreases as the number of probe points increases, reaching an optimum near 100. Beyond this point, the MSE increases again, as FlowPAINT was only trained with 25 probe points. In contrast, the performance of the autoregressive baseline shows little sensitivity to the number of probe points provided.
    }
    \label{fig:varied_probes}
\end{figure}

\begin{table*}[ht!]
\centering
\caption{Results for physical coherence of FlowPAINT (ours) and the autoregressive UNet baseline (AR UNet).
The MAE, MSE and RMSE are taken between the ground truth and the predicted quantity.
$\overline{\mathbf{u}}$ and $\overline{\mathbf{u}'^2}$ denote the mean and variance over time.
$E(k)$ is the predicted kinetic energy spectra.
$\text{MSE}(u_{[0,t]})$ is the mean squared error between the predicted and the ground-truth trajectory as a whole.
All values are taken along 900 timesteps in the test trajectories. 
The values after $\pm$ are standard deviations from averaging over 10 seeds.
Generally, the higher Reynolds number seems to be more difficult for both models.} 
\label{tab:results_appendix} 

\begin{tabular}{llccl|cccc}
\textbf{Model} & \textbf{Cond.} & \textbf{Batchsize} & \textbf{$Re$} & \textbf{Probes} & ${\text{MAE}}(\overline{\mathbf{u}}) \downarrow$  & ${\text{MAE}}(\overline{\mathbf{u}'^2}) \downarrow$  & $\text{MSE}([\mathbf{\mathbf{u_{[0,t]}}}]) \downarrow$ & $\text{RMSE}(E(k)) \downarrow$ \\
 \hline

& \multicolumn{1}{l}{} &  &  &  & {\fontsize{8}{11}\selectfont $ \times 10^{-3} $} & {\fontsize{8}{11}\selectfont $ \times 10^{-5} $} & {\fontsize{8}{11}\selectfont $ \times 10^{-5} $} & {\fontsize{8}{11}\selectfont $ \times 10^{-6} $} \\
 \hline
\multirow{3}{*}{FlowPAINT} & 16/8 & 144 &  \multirow{3}{*}{2100} & \multirow{3}{*}{grid} & $1.2 \pm 1.6$ & $9.9 \pm 21.7$ & $8.1 \pm 3.6$ & $6.9$ \\
 & 16/0 & 144 &  & & $1.3 \pm 2.6$ & $8.7 \pm 15.1$ & $10.8 \pm 3.9$ & $6.3$ \\
 & 16/0  & 216 &  & & $1.5 \pm 3.2$ & $9.1 \pm 15.5$ & $11.5 \pm 4.9$ & $6.9$ \\
  \hline
\multirow{2}{*}{Autoreg.} & 16/0 & 144 & \multirow{2}{*}{2100} & \multirow{2}{*}{grid} & $8.2 \pm 13.7$ & $93.0 \pm 103.2$ & $155.0 \pm 50.1$ & $66.3$ \\
& 1/0 & 144 & & & $9.5 \pm 14.3$ & $98.7 \pm 110.4$ & $158.3 \pm 46.8$ & $70.9$ \\
 \hline
 \hline
\multirow{3}{*}{FlowPAINT} & 16/8 & 144 & \multirow{3}{*}{2100} & \multirow{3}{*}{vertical} & $2.2 \pm 3.1$ & $30.2 \pm 38.4$ & $49.0 \pm 21.0$ & $20.1$ \\
& 16/0  & 144 & & & $3.0 \pm 3.2$ & $30.0 \pm 41.6$ & $68.2 \pm 25.3$ & $20.5$ \\
& 16/0 &  216 &  & & $2.9 \pm 2.9$ & $25.7 \pm 34.6$ & $70.6 \pm 26.6$ & $18.9$ \\
 \hline
\multirow{2}{*}{Autoreg.} & 16/0 & 144 & \multirow{2}{*}{2100} & \multirow{2}{*}{vertical}& $7.4 \pm 12.5$ & $87.6 \pm 96.1$ & $156.2 \pm 50.9$ & $62.4$ \\
& 1/0 & 144 & & & $9.2 \pm 13.8$ & $97.8 \pm 108.8$ & $157.3 \pm 70.0$ & $71.8$ \\
 \hline
 \hline
\multirow{3}{*}{FlowPAINT} & 16/8 & 144 & \multirow{3}{*}{1100} & \multirow{3}{*}{grid} & $0.9 \pm 1.3$ & $2.2 \pm 5.5$ & $2.0 \pm 1.0$ & $2.0$ \\
& 16/0  & 144 & & & $1.2 \pm 2.8$ & $6.2 \pm 10.7$ & $4.6 \pm 1.6$ & $4.6$ \\
& 16/0  & 216 &  & & $1.0 \pm 2.2$ & $4.9 \pm 8.2$ & $3.8 \pm 1.3$ & $3.4$ \\
 \hline
\multirow{2}{*}{Autoreg.} & 16/0 & 144 & \multirow{2}{*}{1100} & \multirow{2}{*}{grid}& $3.7 \pm 6.3$ & $16.0 \pm 24.1$ & $29.9 \pm 8.8$ & $12.3$ \\
& 1/0  & 144 & & & $3.5 \pm 5.6$ & $17.3 \pm 25.9$ & $28.8 \pm 8.3$ & $12.4$ \\
 \hline
 \hline
\multirow{3}{*}{FlowPAINT} & 16/8 & 144 & \multirow{3}{*}{1100} & \multirow{3}{*}{vertical} & $1.1 \pm 1.6$ & $6.3 \pm 11.1$ & $8.9 \pm 3.4$ & $3.7$ \\
  & 16/0  & 144 & & & $3.0 \pm 5.8$ & $8.2 \pm 15.8$ & $24.2 \pm 6.5$ & $9.9$ \\
  & 16/0  & 216 &  & & $3.4 \pm 6.2$ & $10.2 \pm 17.5$ & $27.2 \pm 6.9$ & $11.7$ \\
   \hline
\multirow{2}{*}{Autoreg.} & 16/0 & 144 & \multirow{2}{*}{1100} & \multirow{2}{*}{vertical} & $3.7 \pm 6.1$ & $15.9 \pm 23.9$ & $29.5 \pm 8.8$ & $12.3$ \\
  & 1/0  & 144 & & & $3.4 \pm 5.5$ & $16.9 \pm 25.5$ & $28.8 \pm 8.1$ & $12.2$
\end{tabular}
\end{table*}

\begin{figure*}[h]
    \centering
    \begin{subfigure}{0.48\linewidth}
        \centering
        \includegraphics[width=\linewidth]{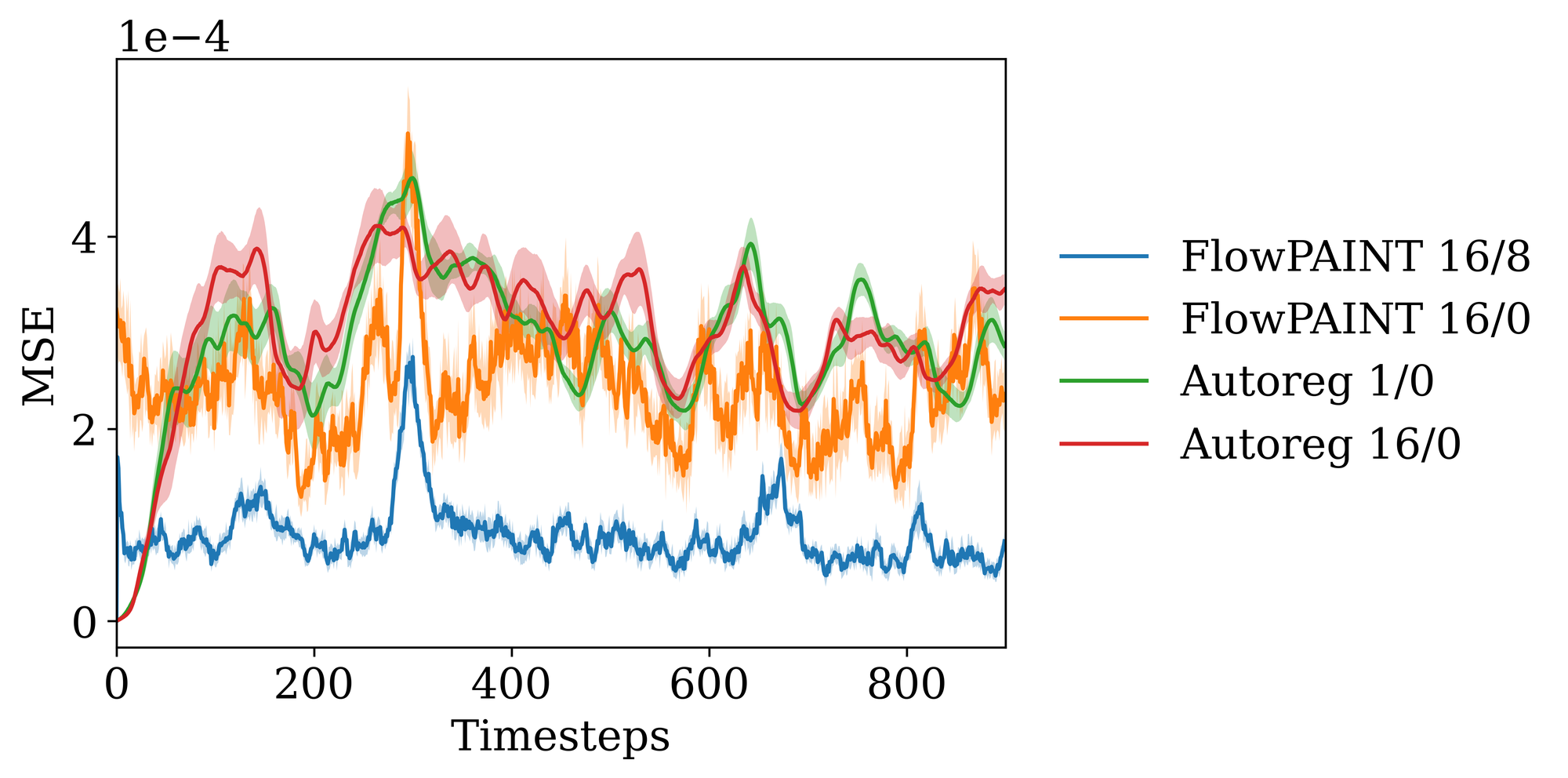}
        \caption{\textit{Vertical} probe constellation.}
    \end{subfigure}
    \hfill
    \begin{subfigure}{0.48\linewidth}
        \centering
        \includegraphics[width=\linewidth]{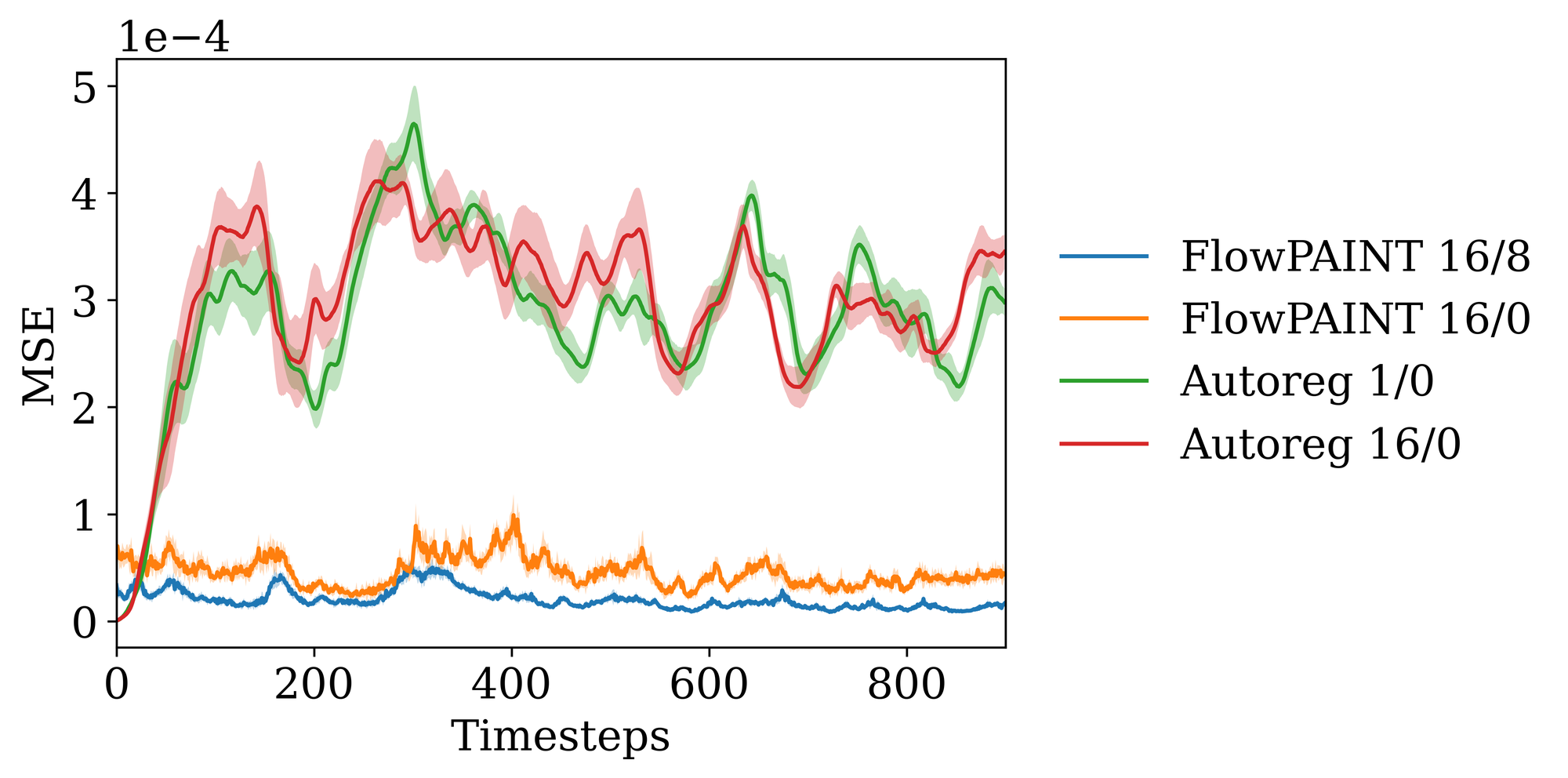}
        \caption{\textit{Grid} probe constellation.}
    \end{subfigure}

    \caption{MSE over time of ground-truth vs. predicted trajectories for the $Re=1100$ test trajectory and different probe point constellations, comparing the autoregressive baselines with FlowPAINT 16/8 and 16/0. The autoregressive model indicates over-reliance on the autoregressive state.}
    \label{fig:Re1100_MSE_over_time}
\end{figure*}

\begin{figure*}[h!]
    \centering
    \includegraphics[width=0.9\linewidth]{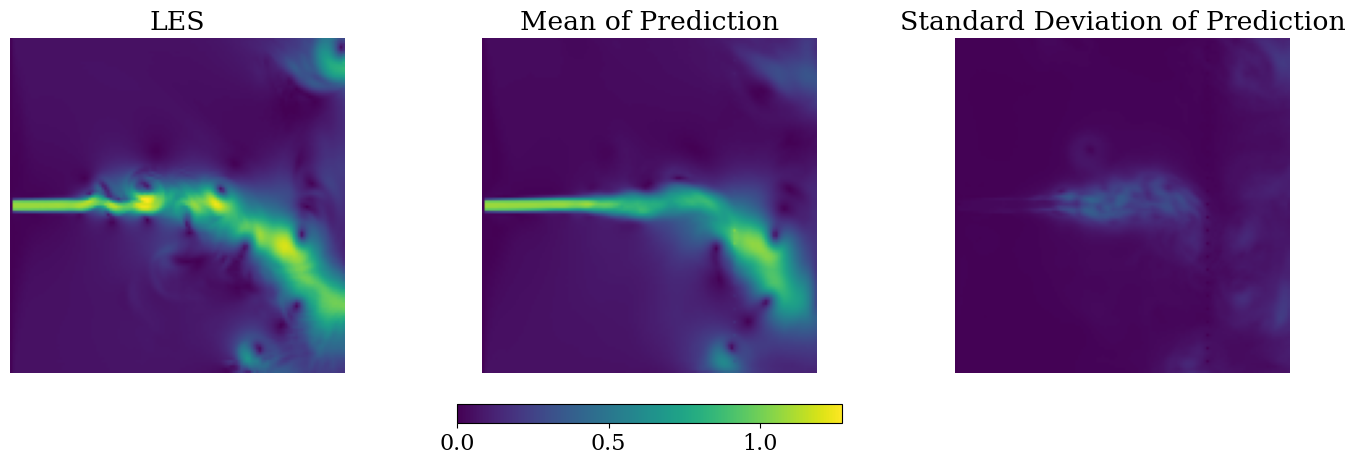}
    \caption{Left: Exemplary groundtruth velocity snapshot, for Reynolds number 2100 and using the \emph{vertical} probe point constellation. Center: Mean of the predictions using FlowPAINT 16/8 over 10 independent seeds. Right: Standard deviation of the predictions using FlowPAINT 16/8 over the same 10 seeds. All values are normalized by the mean inlet velocity.}
    \label{fig:std_of_prediction}
\end{figure*}

\begin{figure*}[h]
    \centering
    
    \begin{subfigure}{0.9\linewidth}
        \centering
        \includegraphics[width=\linewidth]{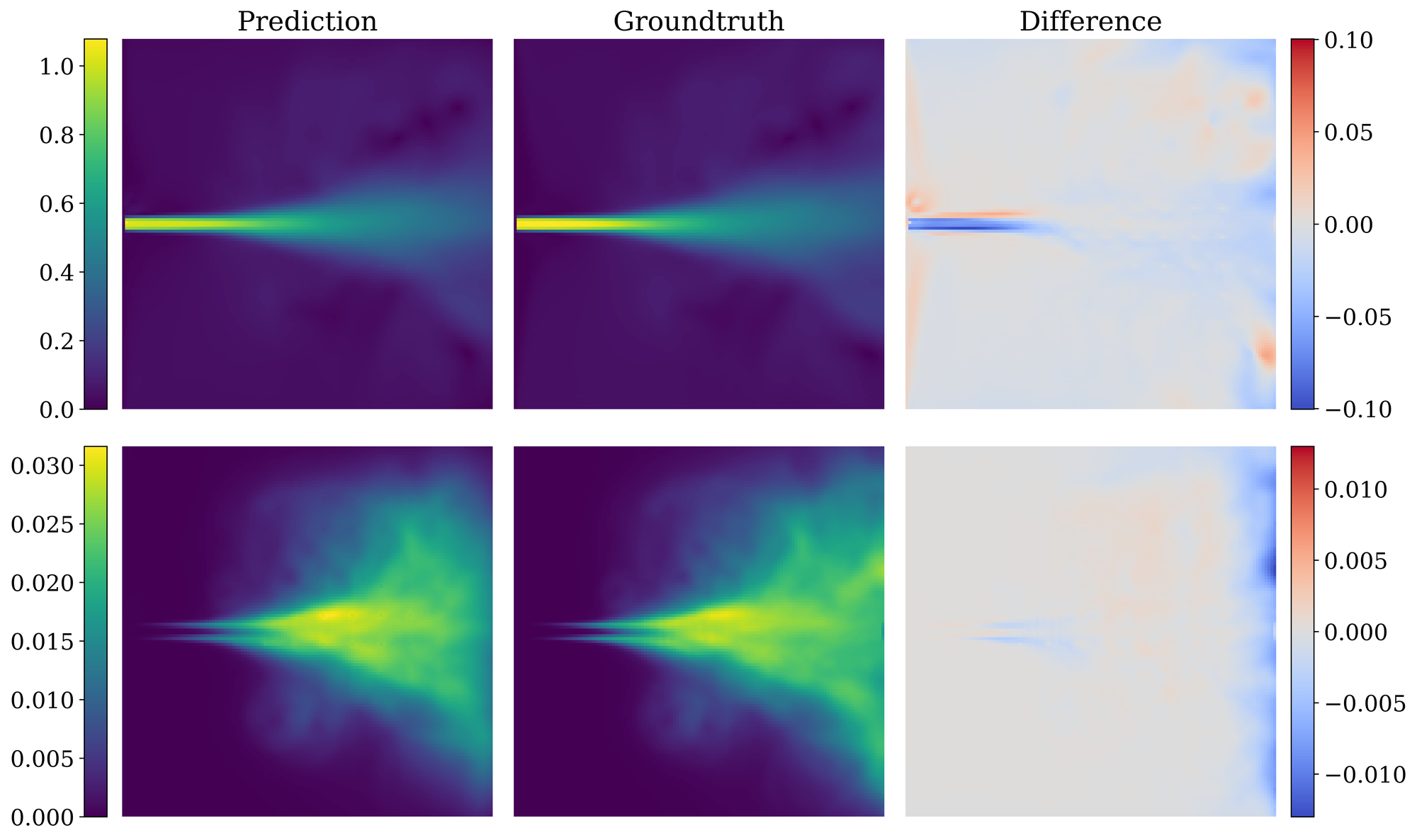}
        \caption{Mean velocity ($\overline{u}$) and variance ($\overline{u'^2}$) of the reconstructed trajectories compared with the ground truth. All values are normalized by the mean inlet velocity.}
        \label{fig:Re2100grid_physicsA}
    \end{subfigure}
    
    \vspace{0.5cm}
    
    \begin{subfigure}{0.48\linewidth}
        \centering
        \includegraphics[width=\linewidth]{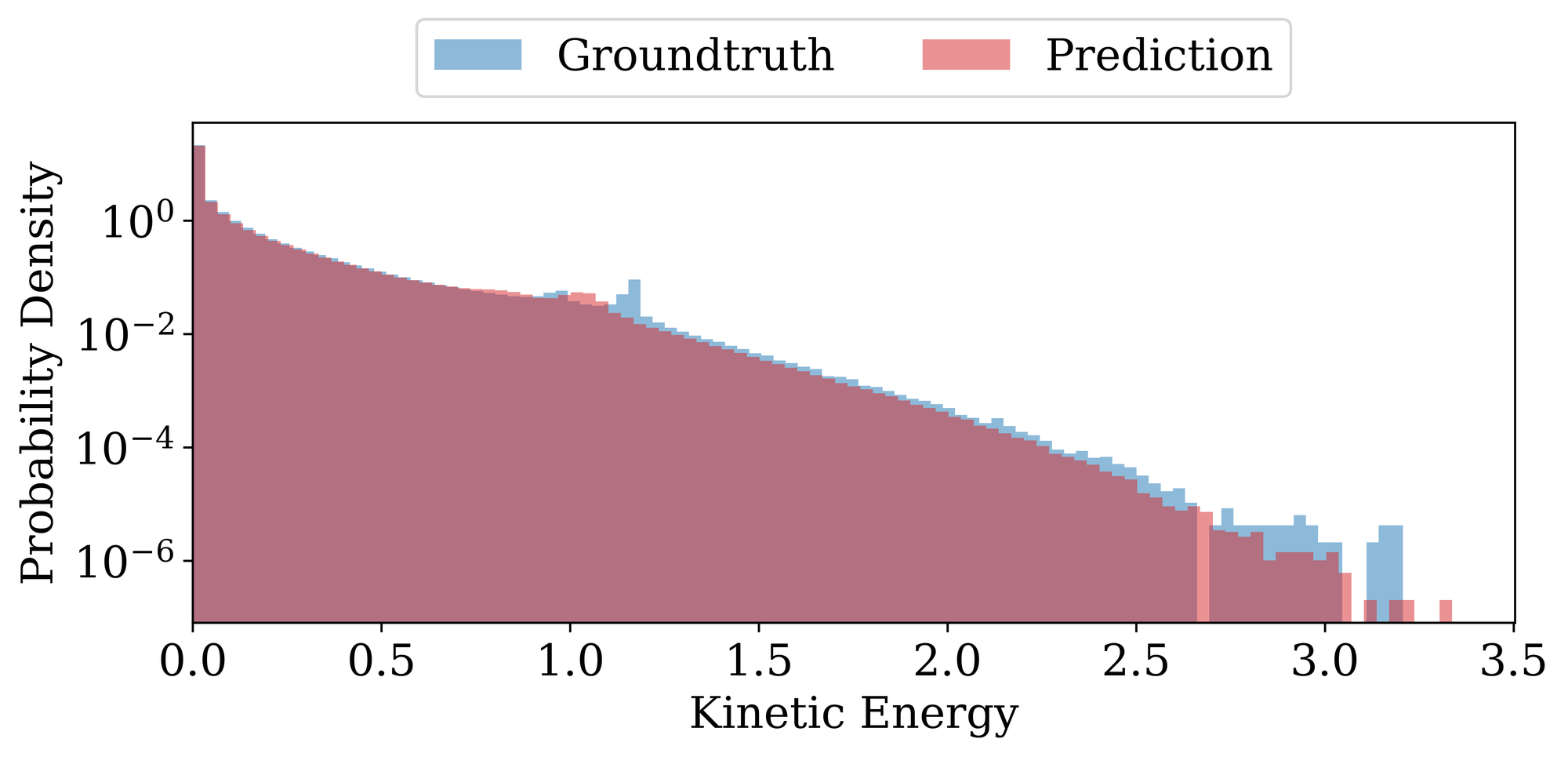}
        \caption{Histogram of time-averaged kinetic energy.}
        \label{fig:Re2100grid_physicsB}
    \end{subfigure}
    \hfill
    \begin{subfigure}{0.48\linewidth}
        \centering
        \includegraphics[width=\linewidth]{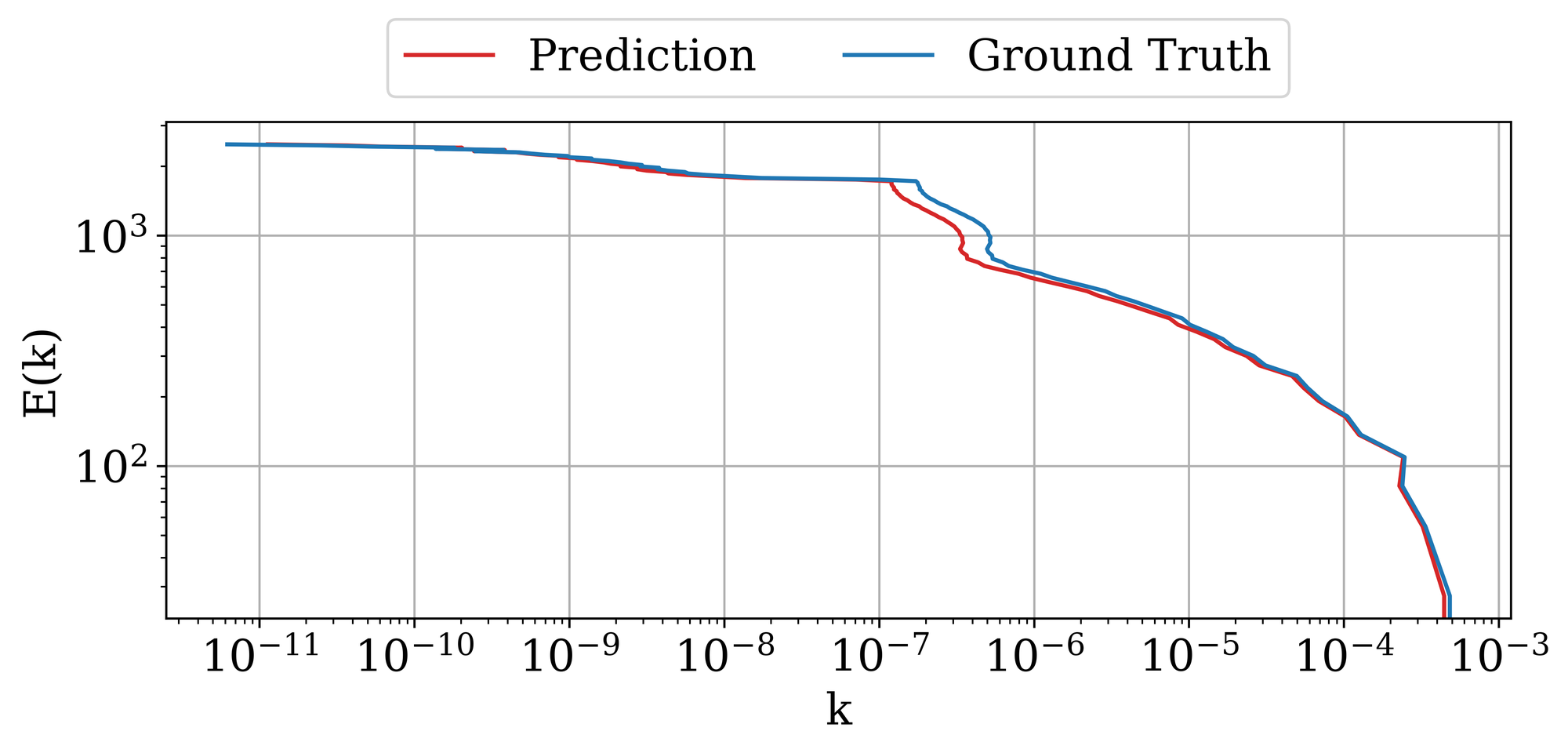}
        \caption{Time-averaged kinetic energy spectrum.}
        \label{fig:Re2100grid_physicsC}
    \end{subfigure}

    \caption{Statistical analysis of reconstructed flow fields using FlowPAINT 16/8 for Reynolds number 2100 and a \emph{grid} probe point constellation.}
    \label{fig:Re2100grid_physics}
\end{figure*}

\begin{figure*}[h] 
    \centering
    
    \begin{subfigure}{0.9\linewidth}
        \centering
        \includegraphics[width=\linewidth]{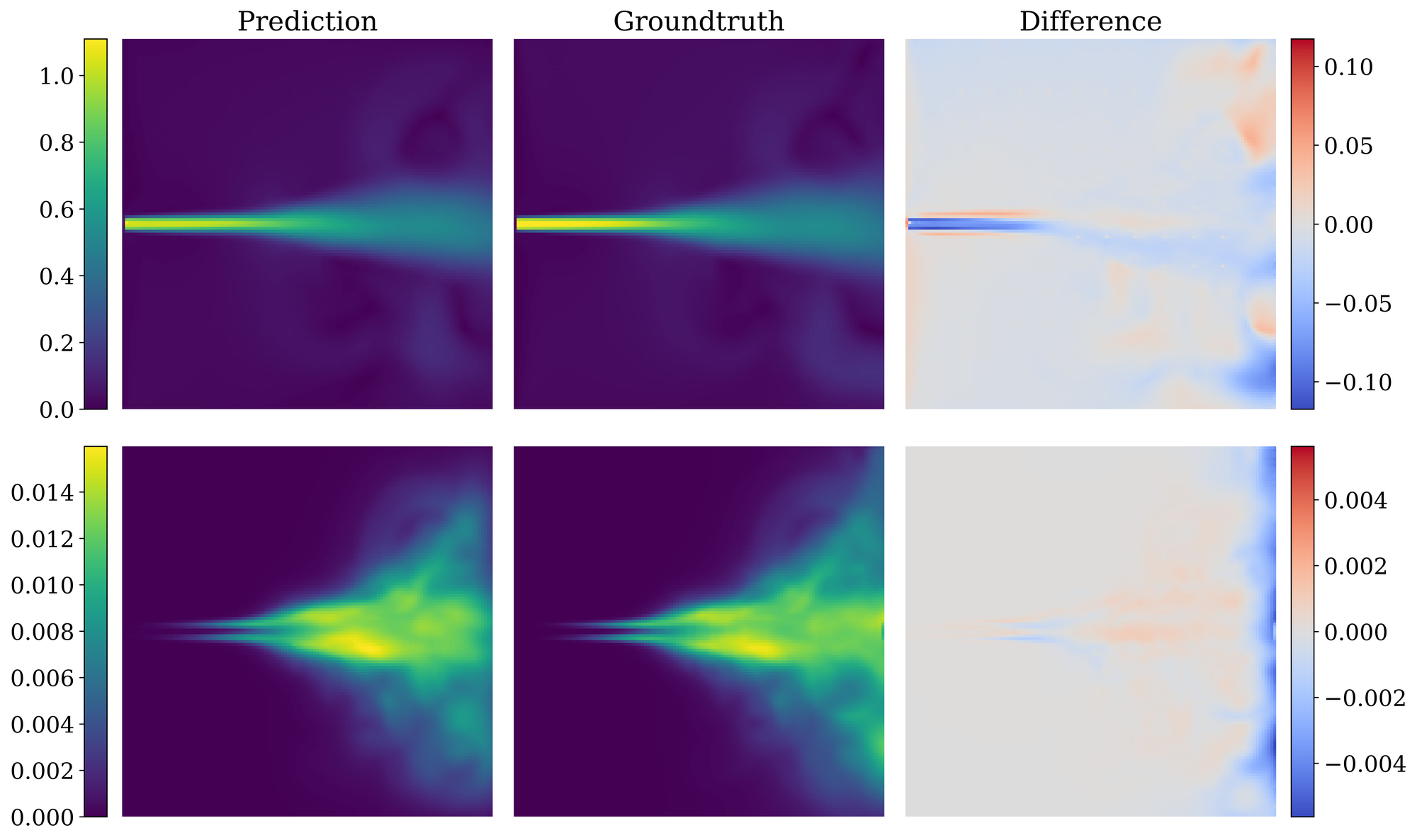}
        \caption{Mean velocity ($\overline{u}$) and variance ($\overline{u'^2}$) of the reconstructed trajectories compared with the ground truth. All values are normalized by the mean inlet velocity.}
        \label{fig:Re1100grid_physicsA}
    \end{subfigure}
    
    \vspace{0.5cm}
    
    \begin{subfigure}{0.48\linewidth}
        \centering
        \includegraphics[width=\linewidth]{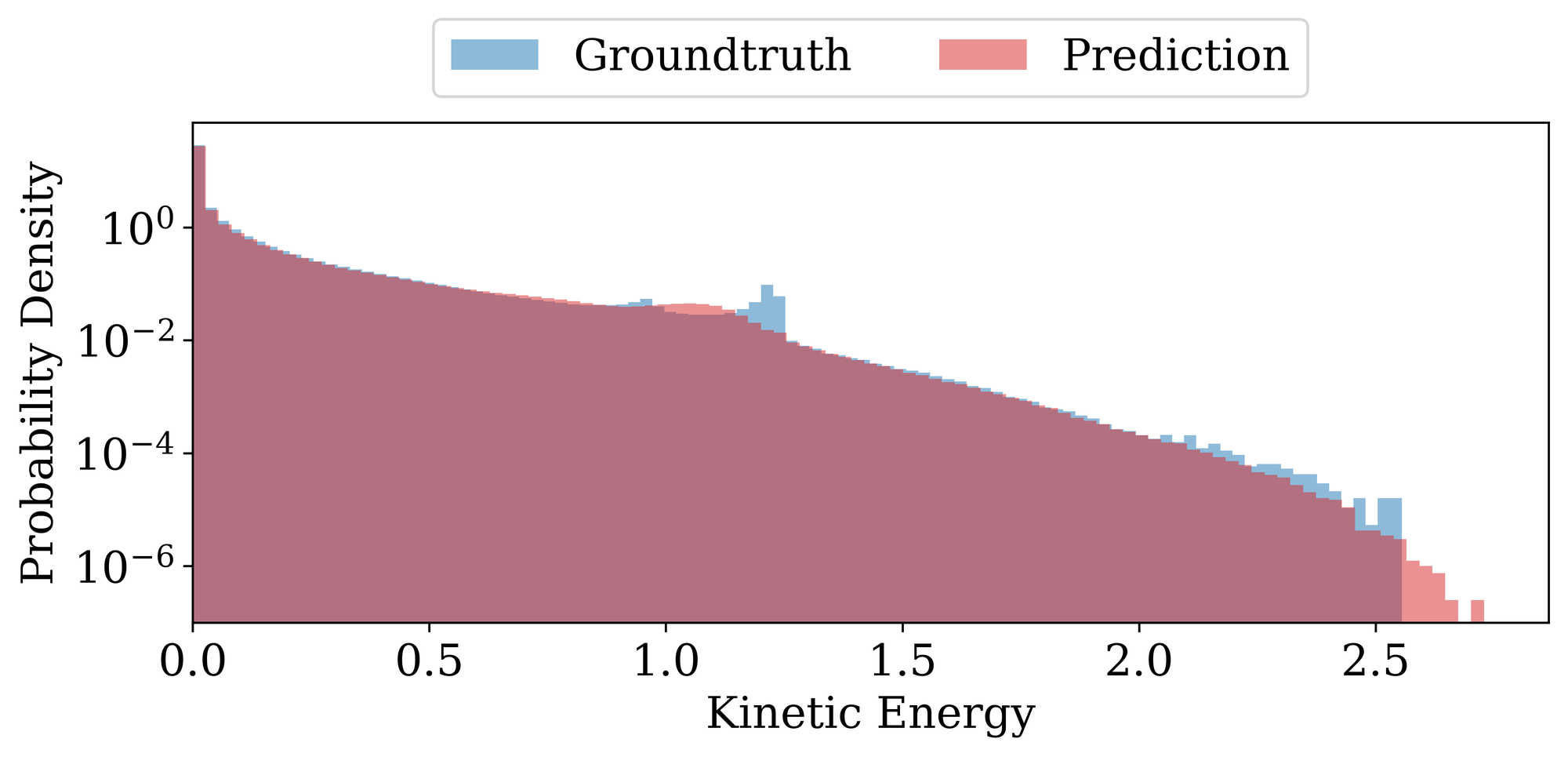}
        \caption{Histogram of time-averaged kinetic energy.}
        \label{fig:Re1100grid_physicsB}
    \end{subfigure}
    \hfill
    \begin{subfigure}{0.48\linewidth}
        \centering
        \includegraphics[width=\linewidth]{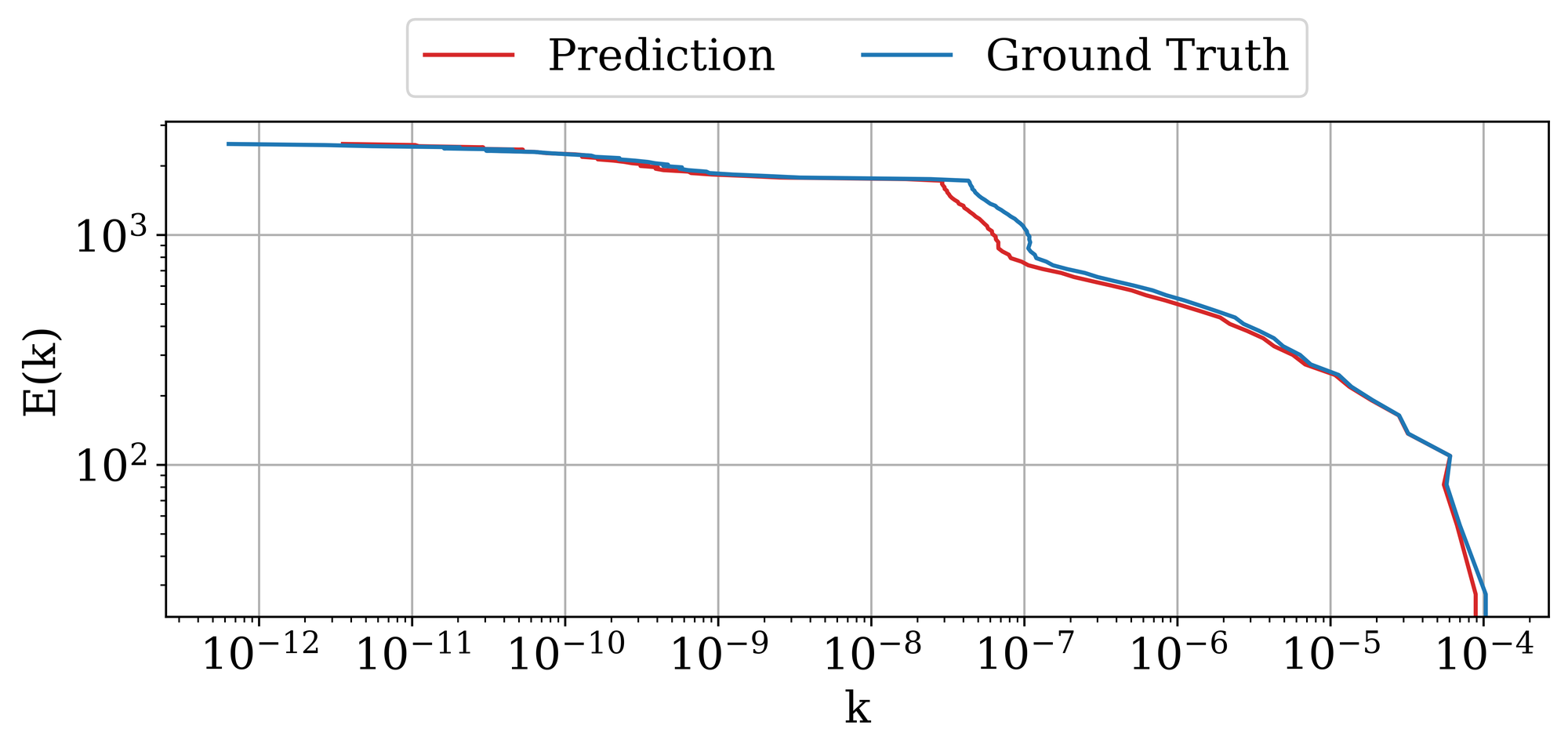}
        \caption{Time-averaged kinetic energy spectrum.}
        \label{fig:Re1100grid_physicsC}
    \end{subfigure}

    \caption{Statistical analysis of reconstructed flow fields using FlowPAINT 16/8 for Reynolds number 1100 and a \emph{grid} probe point constellation.}
    \label{fig:Re1100grid_physics}
\end{figure*}

\begin{figure*}[h] 
    \centering
    
    \begin{subfigure}{0.9\linewidth}
        \centering
        \includegraphics[width=\linewidth]{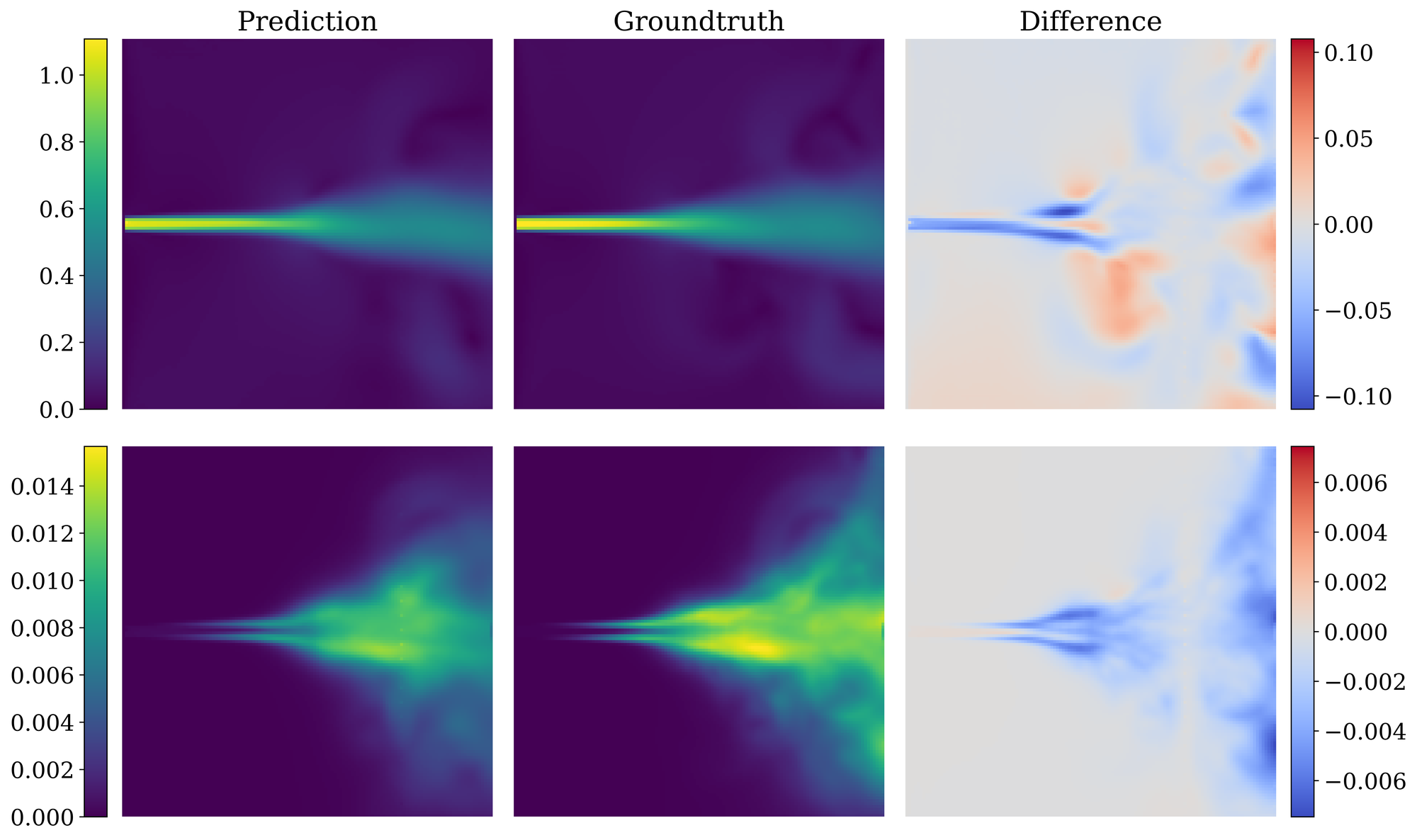}
        \caption{Mean velocity ($\overline{u}$) and variance ($\overline{u'^2}$) of the reconstructed trajectories compared with the ground truth. All values are normalized by the mean inlet velocity.}
        \label{fig:Re1100vertical_physicsA}
    \end{subfigure}
    
    \vspace{0.5cm}
    
    \begin{subfigure}{0.48\linewidth}
        \centering
        \includegraphics[width=\linewidth]{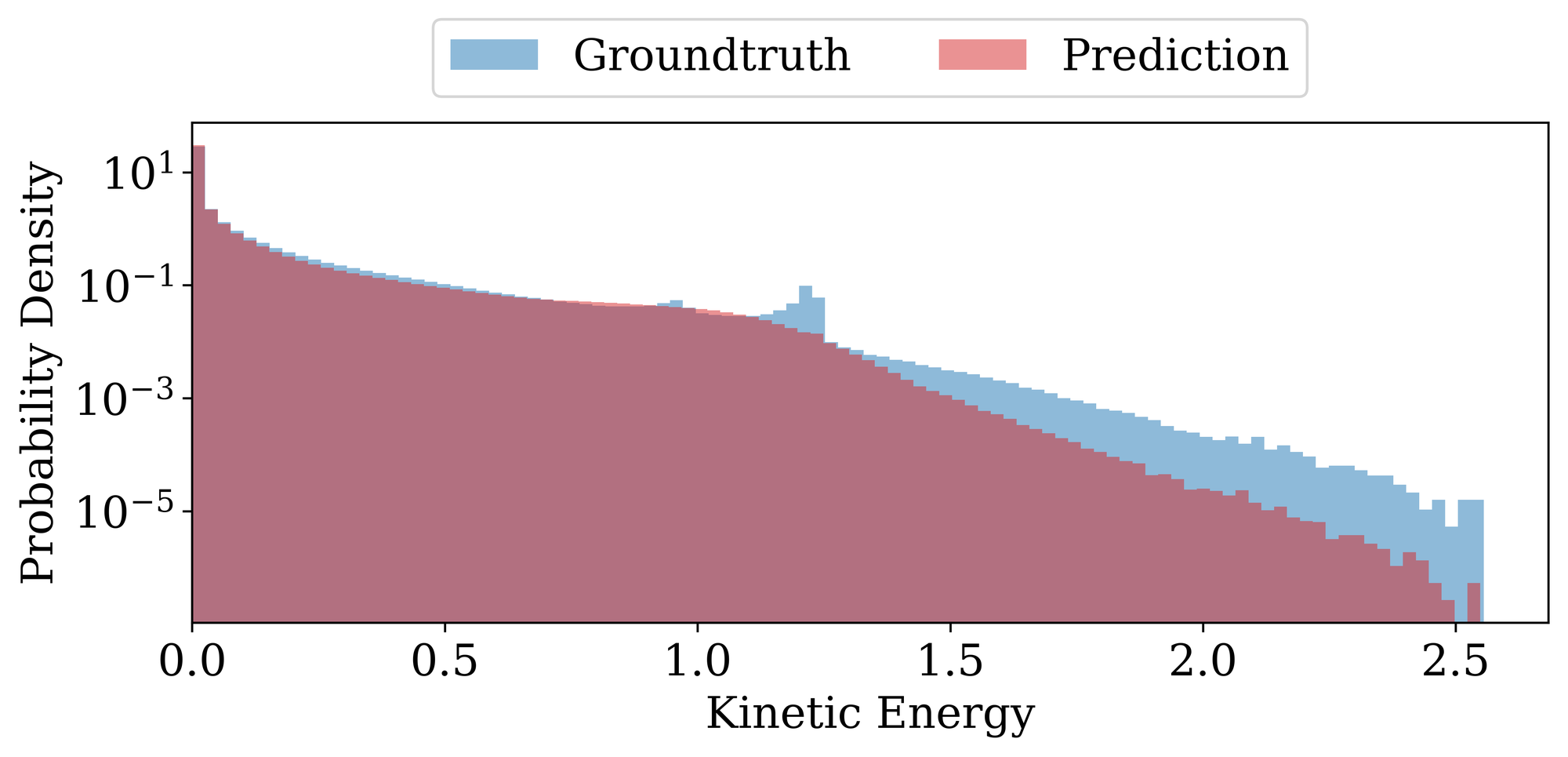}
        \caption{Histogram of time-averaged kinetic energy.}
        \label{fig:Re1100vertical_physicsB}
    \end{subfigure}
    \hfill
    \begin{subfigure}{0.48\linewidth}
        \centering
        \includegraphics[width=\linewidth]{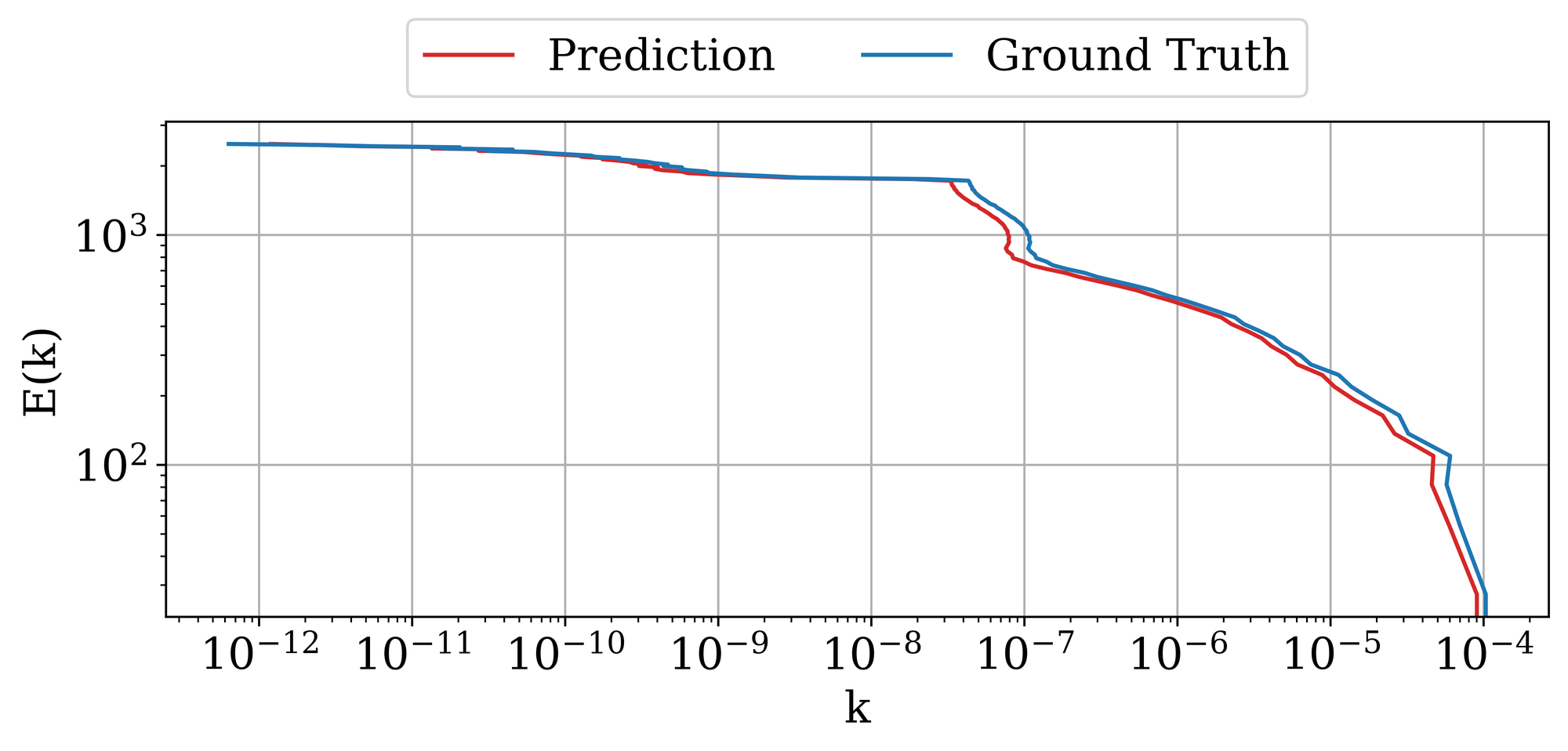}
        \caption{Time-averaged kinetic energy spectrum.}
        \label{fig:Re1100vertical_physicsC}
    \end{subfigure}

    \caption{Statistical analysis of reconstructed flow fields using FlowPAINT 16/8  for Reynolds number 1100 and a \emph{vertical} probe point constellation.}
    \label{fig:Re1100vertical_physics}
\end{figure*}

\begin{figure*}[h]
    \centering
    \begin{subfigure}[t]{0.45\linewidth}  %
        \centering
        \includegraphics[width=\linewidth]{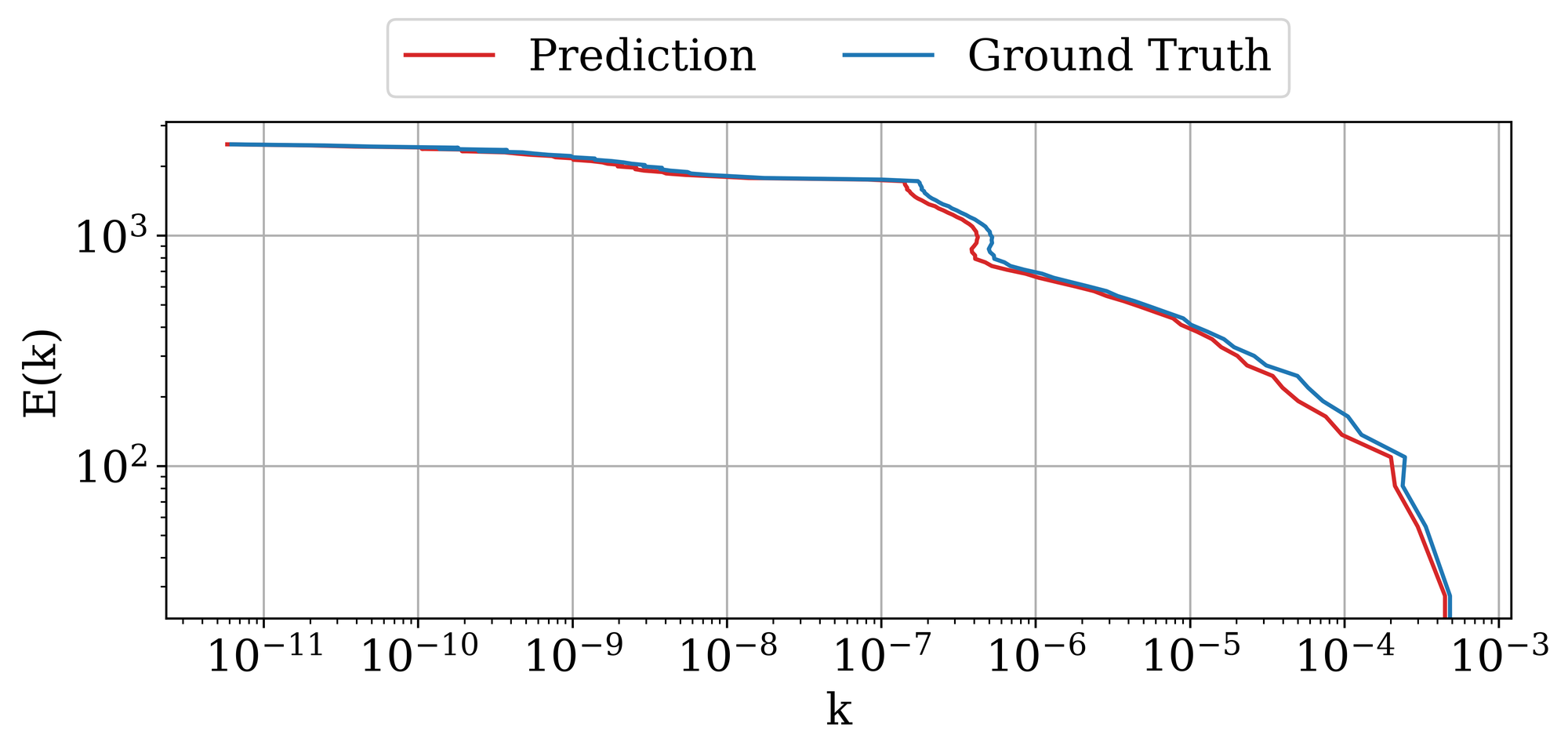}  %
        \caption{Time-averaged kinetic energy spectrum.}
        \label{fig:Re2100vertical_physicsC}
    \end{subfigure}
    \hfill  %
    \begin{subfigure}[t]{0.45\linewidth}
        \centering
        \includegraphics[width=\linewidth]{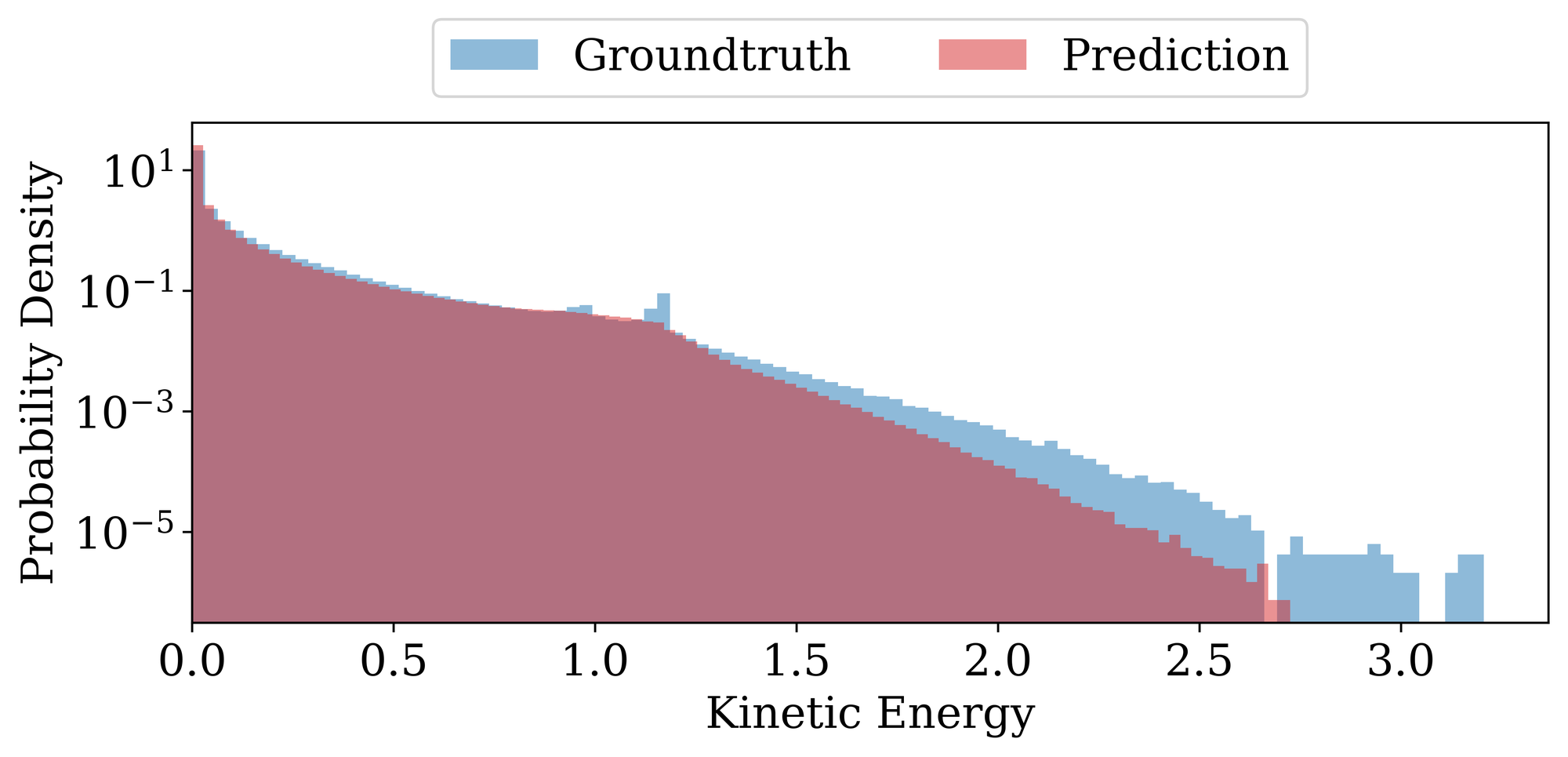}  %
        \caption{Histogram of the kinetic energy.}
        \label{fig:Re2100vertical_physicsB}
    \end{subfigure}
    \caption{Statistical analysis of reconstructed flow fields using FlowPAINT 16/8 for Reynolds number 2100 and a \emph{vertical} probe point constellation.}
    \label{fig:Re2100vertical_physics}
\end{figure*}

\begin{figure*}[h]
    \centering
    \begin{subfigure}{0.75\textwidth}
        \centering
        \includegraphics[width=\linewidth]{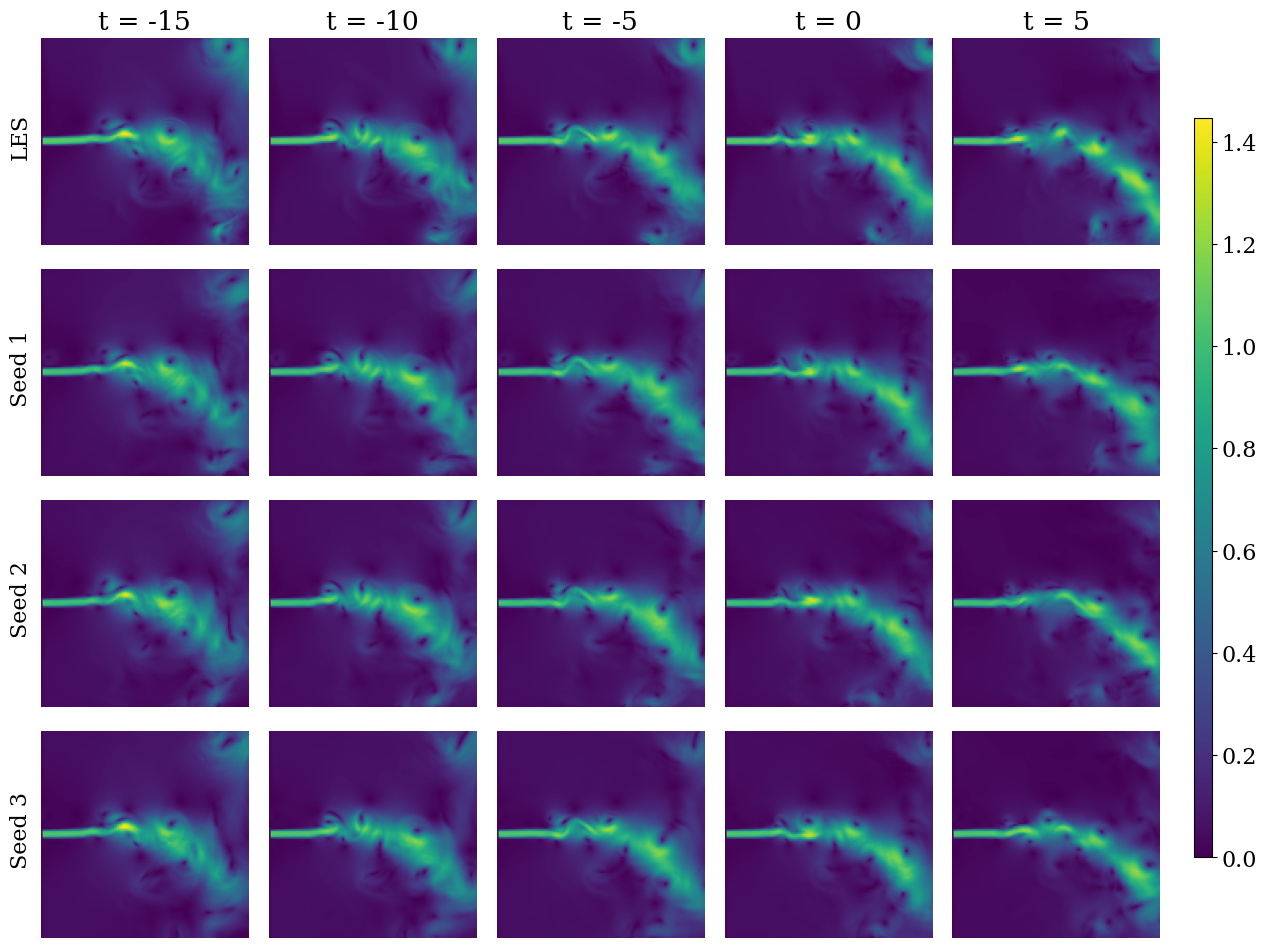}
        \caption{\emph{Grid} probe constellation.}
        \label{fig:sequence_Re2100grid}
    \end{subfigure}

    \vspace{0.5cm}

    \begin{subfigure}{0.75\textwidth}
        \centering
        \includegraphics[width=\linewidth]{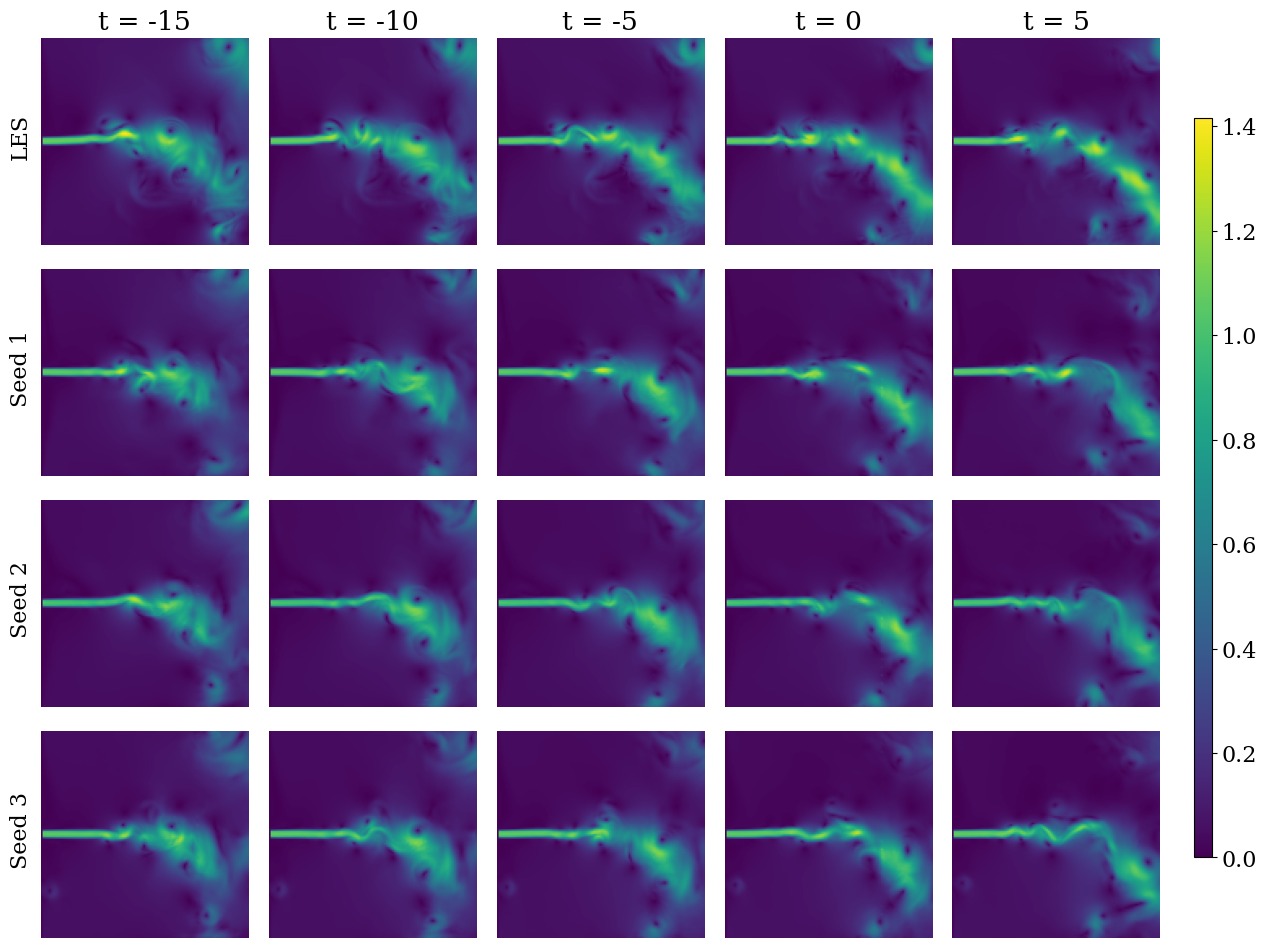}
        \caption{\emph{Vertical} probe constellation.}
        \label{fig:sequence_Re2100vertical}
    \end{subfigure}

    \caption{The top rows show five exemplary LES ground truth snapshots. Time steps with $t \leq 0$ correspond to reconstructions based on measurements, 
    while time steps with $t > 0$ represent predictions with no measurements present. The following three rows display reconstructions and predictions using FlowPAINT 16/8 from three independent random seeds, each showing a connected sequence generated based on the probe information provided. All values are normalized by the mean inlet velocity.}
    \label{fig:sequence_Re2100}
\end{figure*}

\begin{figure*}[h!]
    \centering
    \includegraphics[width=0.9\linewidth]{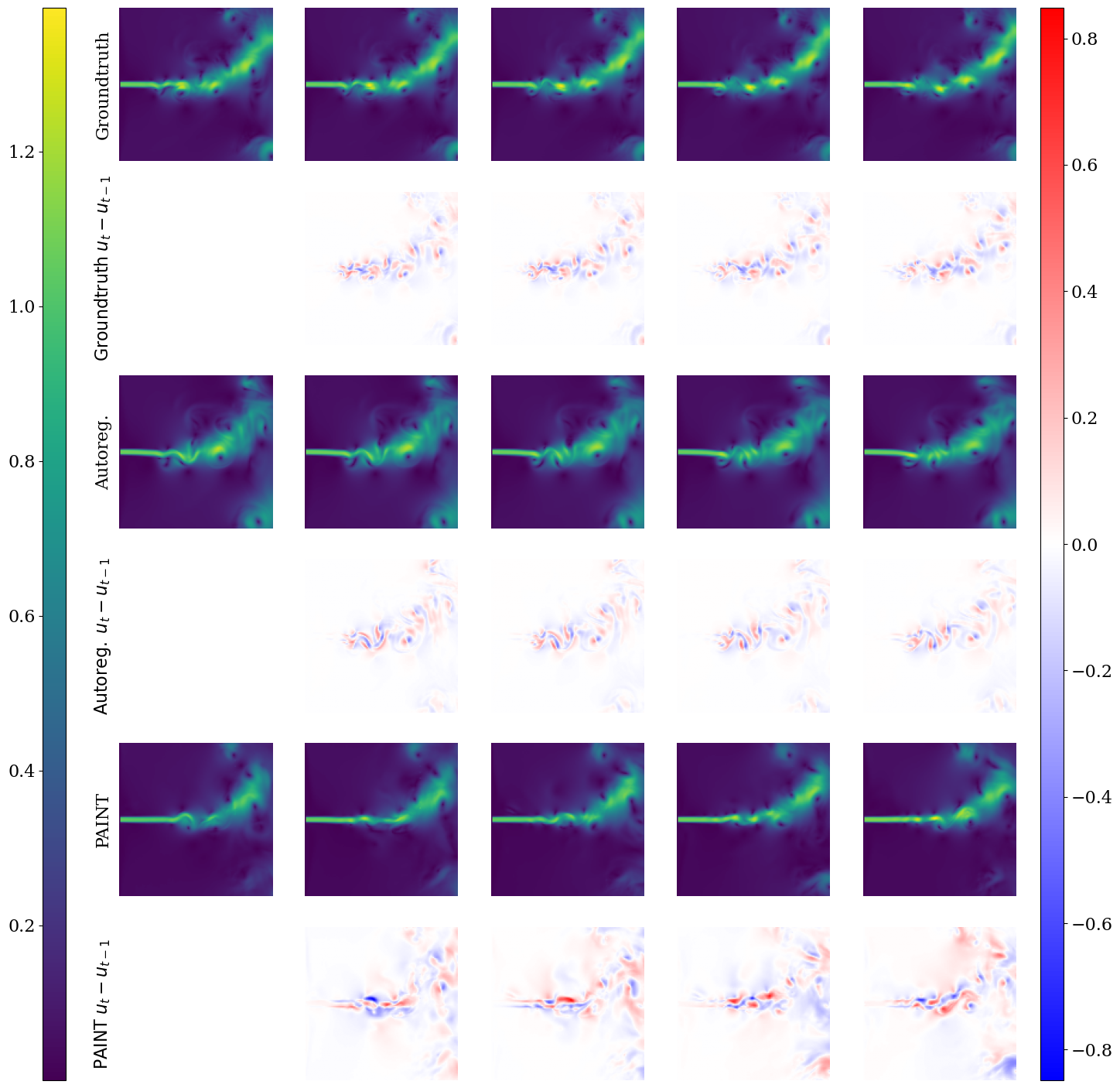}
    \caption{Qualitative illustration of the discontinuities induced by our parallel-in-time approach, using Autoreg. 1/0 and FlowPAINT 16/8 for Reynolds number 2100 and the \textit{vertical} probe point constellation. Shown are consecutive prediction timesteps of a groundtruth trajectory, alongside the autoregressive baseline and the parallel-in-time approach. Due to its inherently temporally discontinuous nature, PAINT exhibits larger frame-to-frame changes. The autoregressive approach, on the other hand, displays frame-to-frame variations comparable to the ground truth, as each prediction is conditioned on previous outputs.}
    \label{fig:discontinuities}
\end{figure*}

\end{document}